\documentclass{article} 
\usepackage{askstream_preprint,times}
\usepackage{fontawesome5}

\usepackage{amsmath,amsfonts,bm}

\def\eqref#1{equation~\ref{#1}}

\def\1{\bm{1}}

\DeclareMathAlphabet{\mathsfit}{\encodingdefault}{\sfdefault}{m}{sl}
\SetMathAlphabet{\mathsfit}{bold}{\encodingdefault}{\sfdefault}{bx}{n}

\usepackage{hyperref}
\usepackage{url}

\usepackage{etoc}
\usepackage{float}

\usepackage{wrapfig}
\usepackage{subcaption}
\usepackage{placeins}
\usepackage{booktabs,xcolor,colortbl,amssymb,graphicx,pifont,tabularx}
\definecolor{mygreen}{HTML}{00B050}
\definecolor{myred}{HTML}{FF0000}
\newcommand{\cmark}{\textcolor{mygreen}{\ding{51}}}
\newcommand{\xmark}{\textcolor{myred}{\ding{55}}}

\usepackage[most]{tcolorbox}
\usepackage{enumitem}

\definecolor{takeawaybg}{RGB}{246,246,252}
\definecolor{takeawaytitle}{RGB}{110,110,110}

\newtcolorbox{takeawaybox}[1]{
    enhanced,
    colback=takeawaybg,
    colframe=black!85,
    boxrule=0.8pt,
    arc=3mm,
    boxsep=0pt,
    left=4mm,
    right=4mm,
    top=2.5mm,
    bottom=2.5mm,
    before skip=6pt,
    after skip=6pt,
    title={#1},
    attach boxed title to top left={
        xshift=6mm,
        yshift*=-\tcboxedtitleheight/2
    },
    boxed title style={
        colback=takeawaytitle,
        colframe=takeawaytitle,
        boxrule=0pt,
        arc=2mm,
        left=3mm,
        right=3mm,
        top=1mm,
        bottom=1mm
    },
    fonttitle=\bfseries\color{white}
}

\usepackage{multirow}
\usepackage{arydshln}
\usepackage{listings}

\tcbuselibrary{listings,breakable}

\definecolor{tableheader}{RGB}{224,224,224}
\definecolor{tablerow}{RGB}{240,240,240}
\definecolor{spiderblue}{RGB}{214,242,246}
\definecolor{spiderline}{RGB}{124,194,204}
\definecolor{gainred}{RGB}{211,145,145}

\newtcblisting{promptbox}[1]{
    enhanced,
    breakable,
    listing only,
    listing engine=listings,
    width=\linewidth,
    colback=black!3,
    colframe=black!75,
    boxrule=0.8pt,
    arc=2mm,
    outer arc=2mm,
    title={#1},
    colbacktitle=black!70,
    coltitle=white,
    fonttitle=\bfseries\small,
    toptitle=1.5mm,
    bottomtitle=1.5mm,
    left=4mm,
    right=4mm,
    top=3mm,
    bottom=3mm,
    before skip=6pt,
    after skip=6pt,
    listing options={basicstyle=\ttfamily\small,breaklines=true,breakatwhitespace=false,columns=fullflexible,keepspaces=true,showstringspaces=false,tabsize=2}
}

\definecolor{examplebg}{RGB}{247,249,252}
\definecolor{exampleborder}{RGB}{190,198,210}
\definecolor{framebg}{RGB}{238,242,247}

\newcommand{\frameitem}[2]{%
    \begin{minipage}[t]{0.132\linewidth}
        \centering
        \fcolorbox{exampleborder}{framebg}{%
            \includegraphics[
                width=0.92\linewidth,
                height=1.65cm,
                keepaspectratio
            ]{#1}%
        }\\[-1pt]
        {\scriptsize #2}
    \end{minipage}%
}

\newtcolorbox{videoexamplebox}{
    enhanced,
    colback=examplebg,
    colframe=exampleborder,
    boxrule=0.7pt,
    arc=3mm,
    outer arc=3mm,
    left=2.5mm,
    right=2.5mm,
    top=2.5mm,
    bottom=2.5mm,
    boxsep=0pt
}
\definecolor{linkblue}{HTML}{216B92}
\usepackage{fontawesome5}

\definecolor{citeblue}{HTML}{0000EE}

\hypersetup{
    colorlinks=true,
    citecolor=citeblue,
    linkcolor=black,
    urlcolor=citeblue
}

\setcitestyle{authoryear,round,semicolon}

\usepackage{tikz}
\usetikzlibrary{calc}
\usepackage{eso-pic}

\definecolor{askpurple}{HTML}{6654D8}
\definecolor{askblue}{HTML}{5B8DEF}
\definecolor{askbg}{HTML}{F6F4FF}
\definecolor{askborder}{HTML}{D9D2F5}
\usepackage{sectsty}

\sectionfont{\color{askpurple}}
\subsectionfont{\color{askpurple!75!white}}
\subsubsectionfont{\color{askpurple!50!white}}

\newcommand{\redref}[2]{%
    \begingroup
    \setlength{\fboxsep}{1.5pt}%
    \setlength{\fboxrule}{0.6pt}%
    \fcolorbox{red}{white}{%
        \hyperref[#2]{\textcolor{black}{#1~\ref*{#2}}}%
    }%
    \endgroup
}

\title{%
\parbox{\textwidth}{%
    \centering

    {\color{askpurple}\rule{\linewidth}{2pt}}\par
    \vspace{0.55em}

    \raisebox{-0.2\height}{%
        \includegraphics[height=1.2em]{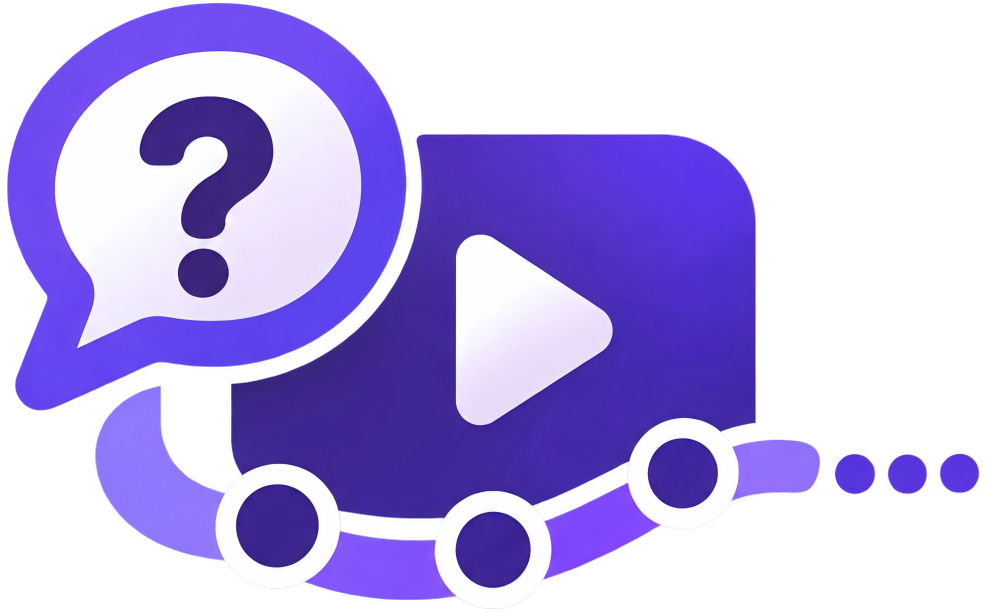}%
    }%
    \hspace{0.15em}%
    AskStream: When and What to Ask Proactively
    under Uncertainty in Streaming Video Understanding

    \par\vspace{0.4em}
    \vspace{-10pt}
    {\color{askpurple}\rule{\linewidth}{0.5pt}}%
}%
}

\author{%
\begin{minipage}[t]{0.99\textwidth}
\centering
\normalfont
{\normalsize\bfseries
Lu Wang\textsuperscript{1,2},
Zhuoran Jin\textsuperscript{1},
Longxiang Wang\textsuperscript{1},
Junhong Chen\textsuperscript{1},
Yupu Hao\textsuperscript{1},\\[2pt]
Yao Xu\textsuperscript{2},
Daojian Zeng\textsuperscript{3},
Yubo Chen\textsuperscript{4},
Kang Liu\textsuperscript{1},
Guang Liu\textsuperscript{2},
Jun Zhao\textsuperscript{1}}\\[0.7em]
{\footnotesize
\textsuperscript{1}The Key Laboratory of Cognition and Decision Intelligence for Complex Systems,\\
Institute of Automation, Chinese Academy of Sciences, Beijing, China\\[2pt]
\textsuperscript{2}Beijing Academy of Artificial Intelligence (BAAI), Beijing, China\\[2pt]
\textsuperscript{3}Hunan Normal University, China\\[2pt]
\textsuperscript{4}Minzu University of China, Beijing, China\\[5pt]
\texttt{wanglu2026@ia.ac.cn}\\[10pt]
{\large
\mbox{%
    \href{www.askstream.top}{%
        \raisebox{-0.2\height}{\includegraphics[height=1.1em]{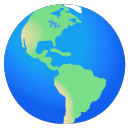}}%
        \hspace{0.25em}\textcolor{linkblue}{\textbf{Website}}%
    }%
    \hspace{1.2em}%
    \href{https://github.com/wanglu-cs/AskStream}{%
        \textcolor{black}{\faGithub}%
        \hspace{0.25em}\textcolor{linkblue}{\textbf{Code}}%
    }%
    \hspace{1.2em}%
    \href{https://huggingface.co/datasets/wanglucs/AskStream}{%
        \raisebox{-0.2\height}{\includegraphics[height=1.1em]{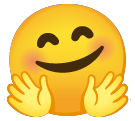}}%
        \hspace{0.25em}\textcolor{linkblue}{\textbf{Data}}%
    }%
}
}
}
\end{minipage}%
}

\begin{document}

\maketitle
\vspace{-15pt}
\begin{tcolorbox}[
    enhanced,
    colback=askbg,
    colframe=askborder,
    boxrule=0.6pt,
    arc=2.5mm,
    left=6mm,
    right=6mm,
    top=3mm,
    bottom=4mm,
    before skip=8pt,
    after skip=10pt
]

{\centering
\bfseries
\color{askpurple}
Abstract\par
}

\vspace{0.4em}

\small
\noindent
Recent advances in multimodal large language models have increased interest in interaction paradigms for video understanding. The evolution from offline to streaming and proactive interaction has gradually shifted models from passive responders toward more proactive agents that can decide when to answer. However, existing proactive paradigms typically assume that the information required to answer a query is complete, leaving models to only \textit{wait} or \textit{answer}. In the real world, however, information is often incomplete due to uncertainty from either the \textbf{video} or the \textbf{user}. In proactive paradigms, such uncertainty can cause the model to wait forever or generate highly hallucinatory responses. We therefore argue that proactive interaction should be redefined with a third action: \textit{ask}, allowing models to actively ask for missing information. To study when and what models should ask before answering, we introduce \textbf{AskStream}, the first multi-turn benchmark spanning video and user uncertainty for streaming video understanding, with four task categories and eight subtasks. It contains 438 videos with 2,714 manually annotated frame-level time windows across three types: wait, ask, and answer. Across both multiple-choice and open-ended settings, even the strongest model attains only 52.18\% on our proposed \textbf{Ask-F1} metric, highlighting substantial limitations in proactive information seeking under uncertainty.

\end{tcolorbox}

\section{Introduction}
\begin{figure}[t]
    \centering
    \includegraphics[width=\linewidth]{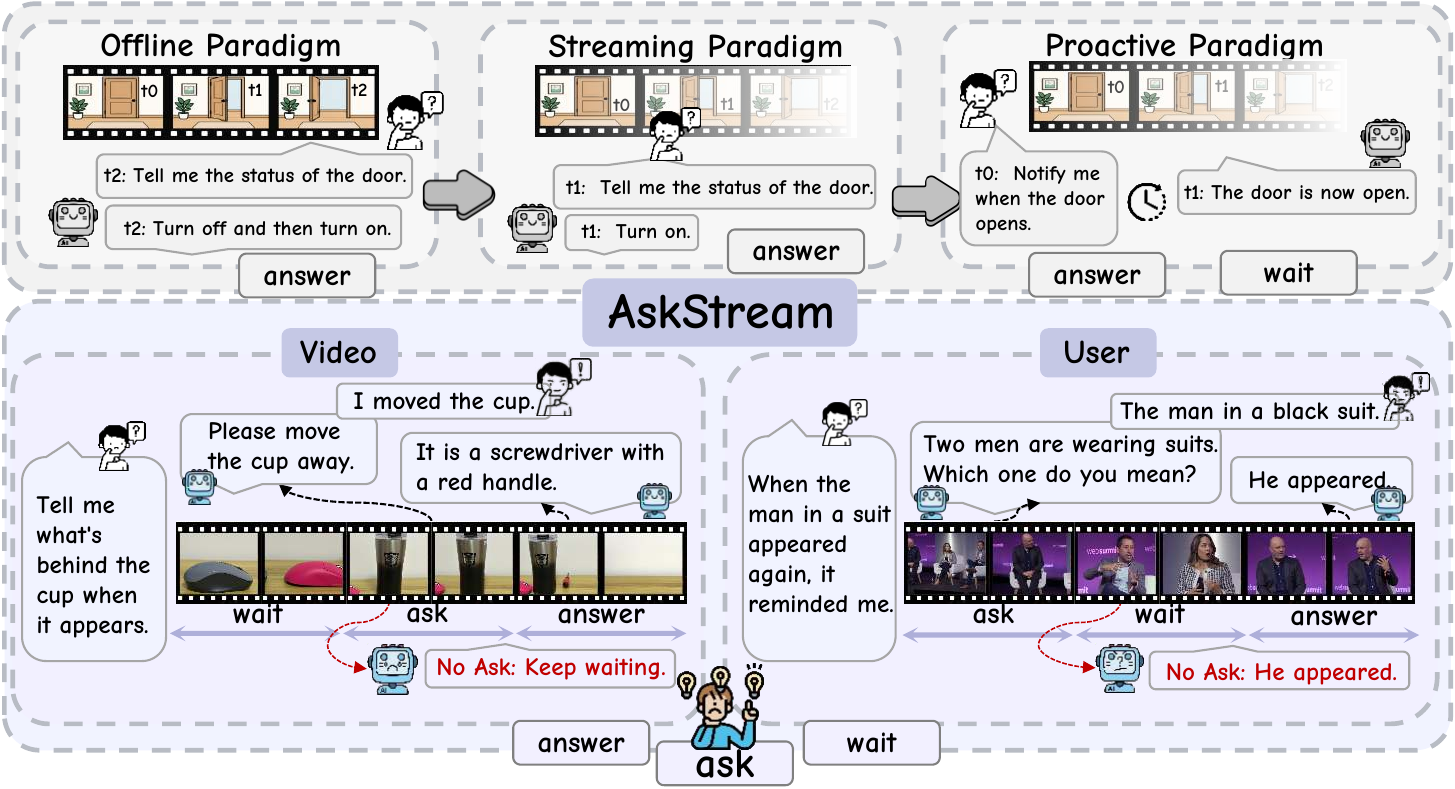}
    \caption{Evolution of interaction paradigms. AskStream extends the conventional action space of \textit{wait} and \textit{answer} by introducing \textit{ask}, enabling models to proactively seek missing information from the video or user before answering.}
    \label{fig1}
\end{figure}

Multimodal large language models (MLLMs) have made significant progress in video understanding and reasoning in recent years~\citep{kimiteam2026kimik3openfrontier,qwenteam2026qwen35omnitechnicalreport,google2026gemini38flash,openai2026gpt56}. Beyond advancements in model architecture and training paradigms, research on \textbf{interaction paradigms} is equally important for advancing models toward more adaptive and generalizable intelligence in real-world environments. As shown at the top of Figure~\ref{fig1}, traditional video interaction primarily follows an \textbf{offline paradigm}, where the model observes the complete video before answering user queries. 
In the real world, videos often arrive continuously as streams. This has driven the emergence of a \textbf{streaming paradigm}, where
models must continuously understand the incoming stream, maintain memory, and interact with users in real time. 
Under both paradigms, the core task remains to determine \textit{how to answer}. However, because the streaming paradigm primarily targets real-time responses, it is less suited to scenarios requiring delayed responses, such as anomaly alerts, procedural guidance, and continuous task monitoring. This motivates the \textbf{proactive paradigm}, where models continuously monitor the video and decide not only \textit{how to answer} but also \textit{when to answer}. Existing benchmarks~\citep{lin2026streamingbench,wang2025proactivevideoqacomprehensivebenchmarkevaluating,azad2026streamreadylearninganswerlong,li2026ipibenchevaluatinginteractiveproactive} progressively extend this paradigm by requiring models to wait for sufficient evidence, supporting multi-turn responses, modeling answer readiness, and adapting response timing under dynamic tasks, with a common focus on increasingly fine-grained response timing.

\begin{figure}[t]
    \centering
    \begin{minipage}[t]{0.49\linewidth}
        \centering
        \includegraphics[width=\linewidth]{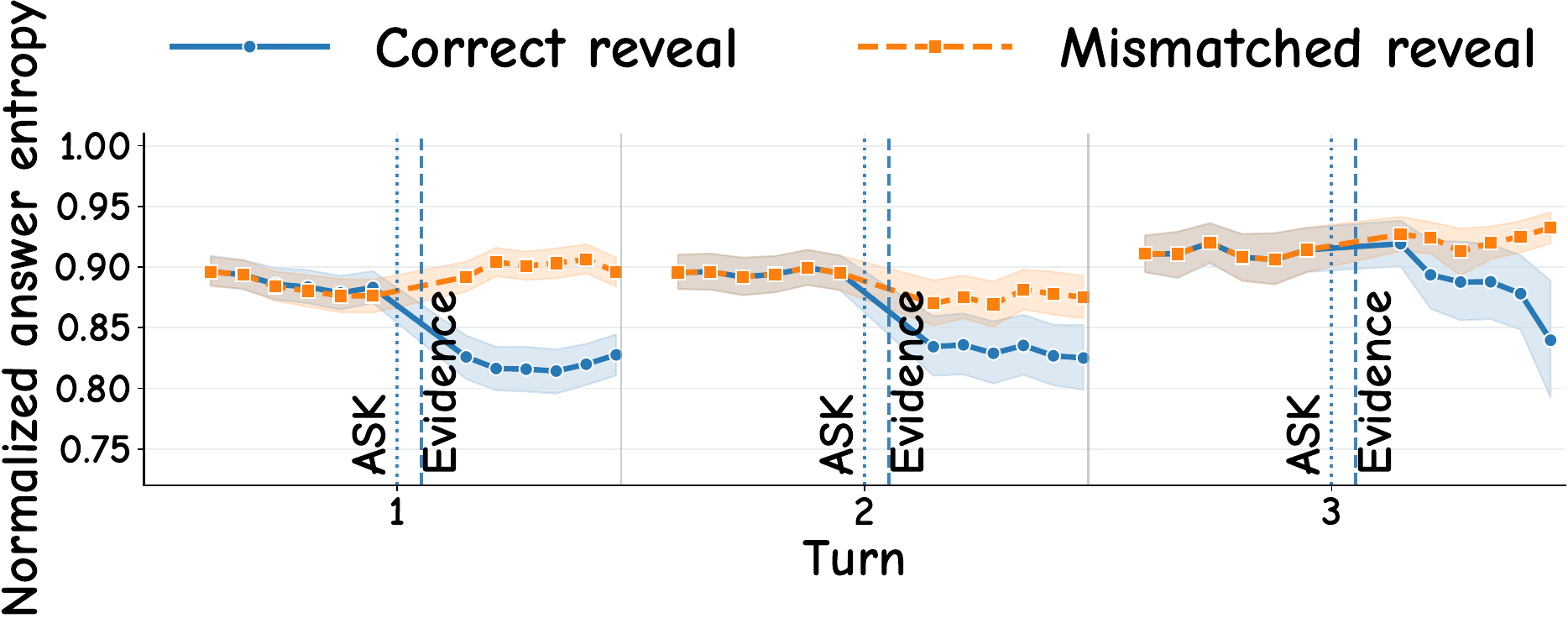}\\[-0.5em]
        {\footnotesize (a)}
    \end{minipage}\hfill
    \begin{minipage}[t]{0.24\linewidth}
        \centering
        \includegraphics[width=\linewidth]{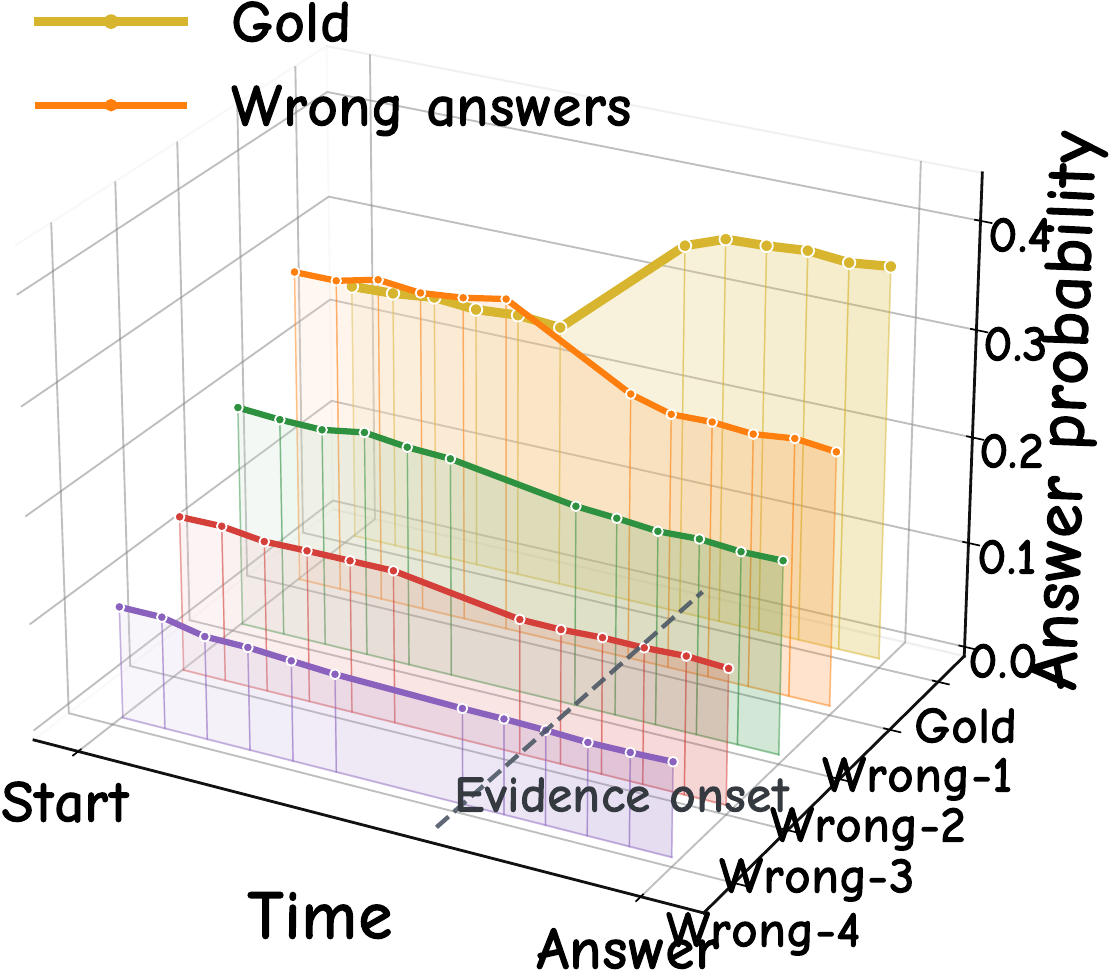}\\[-0.5em]
        {\footnotesize (b)}
    \end{minipage}\hfill
    \begin{minipage}[t]{0.24\linewidth}
        \centering
        \includegraphics[width=\linewidth]{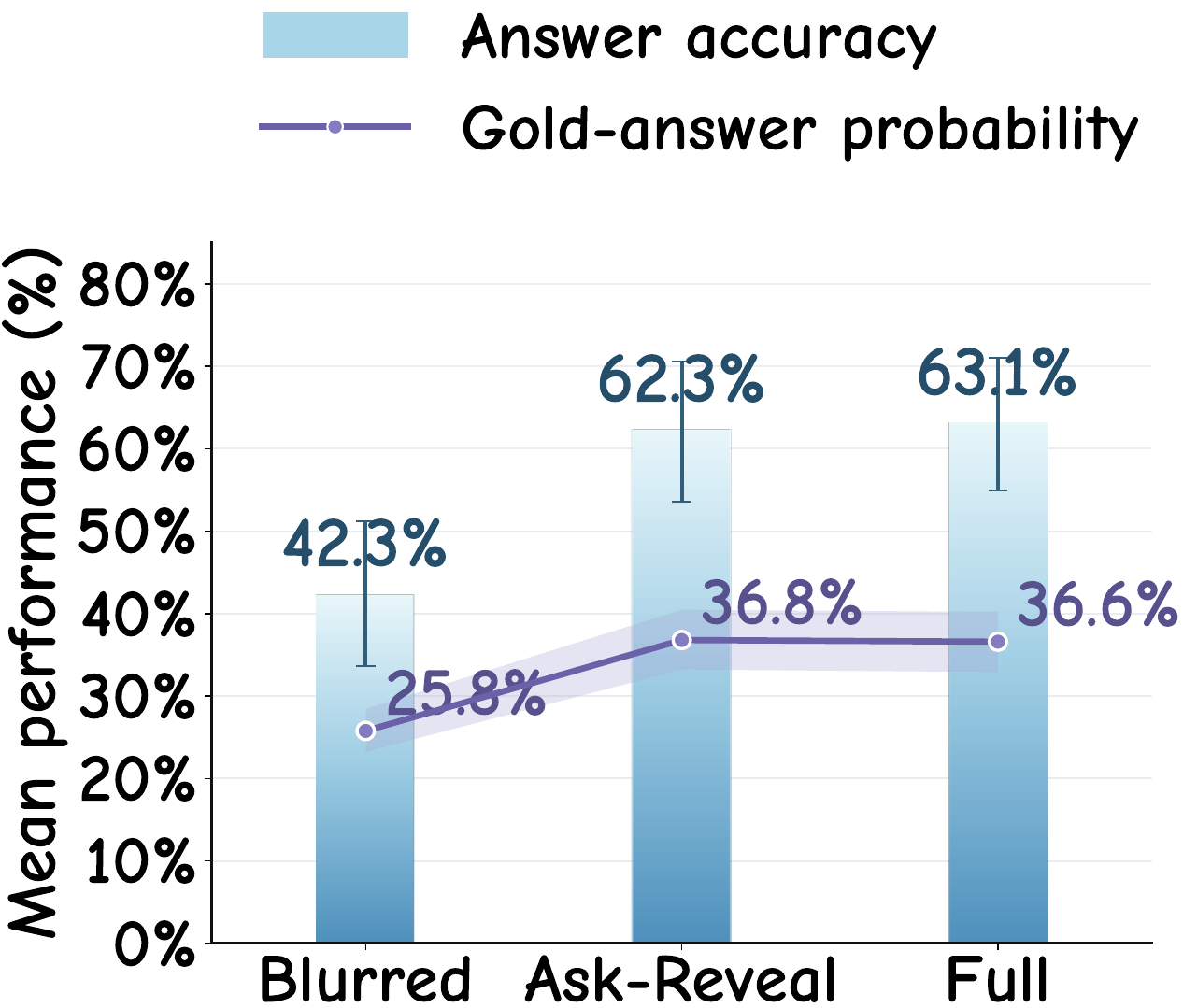}\\[-0.5em]
        {\footnotesize (c)}
    \end{minipage}

    \caption{
    \textbf{Analysis of model uncertainty.}
    (a) Across interaction turns, correct evidence consistently lowers normalized answer entropy, unlike mismatched evidence.
    (b) Relevant evidence shifts probability toward the correct answer.
    (c) Accuracy improves from 42.3\% (blurred) to 62.3\% after asking for key evidence, nearing 63.1\% with fully clear input.
    }
    \label{fig2}

\end{figure}

However, \textbf{the paradigm implicitly assumes that the information for answering is fully contained in the stream and will eventually become available, leaving the model only to \textit{wait} or \textit{answer}.} This assumption overlooks a key challenge: \textit{information in the real world is often incomplete}. Such incompleteness may cause the model to wait forever or produce hallucinatory responses. As shown at the bottom of Figure~\ref{fig1}, the model should proactively ask for missing video information or clarify which person is being referred to before answering. To further quantify the effect of acquiring missing information, we analyze the model's uncertainty.~\footnote{See Appendix~\ref{app:uncertainty_analysis} for experimental details.} As shown in Figure~\ref{fig2}, correct evidence reduces uncertainty and shifts predictions toward the correct answer, raising accuracy from 42.3\% to 62.3\%, while mismatched evidence provides little benefit. These observations show that acquiring key information improves model's performance. We therefore argue that the proactive paradigm should be redefined beyond \textit{wait}\&\textit{answer} to also proactively \textit{ask} missing information when needed.

Based on the above analysis, we introduce \textbf{AskStream}, which is the first multi-turn benchmark to incorporate an \textit{ask} action into the proactive paradigm in streaming video understanding. This paradigm is suitable for applications, such as everyday visual assistance and remote operational guidance.
We categorize information uncertainty into \textbf{video uncertainty} and \textbf{user uncertainty}. Video uncertainty arises from missing visual evidence or missing personalized visual concepts, whereas user uncertainty arises from ambiguous user intent or insufficient knowledge of user preferences. We further divide these four uncertainty categories into eight subtasks, as illustrated in Figure~\ref{fig3}. Compared with prior proactive benchmarks that annotate only timestamps, we provide fine-grained manual annotations of time windows for \textit{wait}, \textit{ask}, and \textit{answer} and introduce \textbf{Ask-F1} to jointly evaluate the timing and content correctness of both asking and answering. To enable more realistic evaluation, we develop \textbf{StreamHarness}, a hierarchical memory framework that retains recent frames while compressing earlier observations and evaluate models under both multiple-choice and open-ended settings. Experimental results reveal a substantial gap between offline and streaming proactive models: the strongest offline model achieves 52.18\% Ask-F1, while the best streaming proactive model reaches only 7.22\%. Early ASK and wrong ASK are the two common error types, while providing oracle information about both \textit{when} and \textit{what} to ask improves Ask-F1 by 43.83\%$\uparrow$, highlighting proactive asking as a major bottleneck in current models.
\noindent\textbf{Our contributions} are as follows:

\begin{itemize}[leftmargin=1.3em, labelsep=0.5em]

\item\textbf{A new interaction paradigm.}
Quantitative and qualitative analyses show that proactively acquiring missing information reduces model uncertainty, motivating a new proactive interaction paradigm with three actions: \textit{wait}, \textit{ask}, and \textit{answer}.

\item\textbf{AskStream benchmark.}
We introduce AskStream, a multi-turn benchmark that jointly evaluates when and what to ask and when to answer across eight subtasks spanning video and user uncertainty, with fine-grained manually annotated action windows, \textbf{Ask-F1} metric and the \textbf{StreamHarness} hybrid framework.

\item\textbf{Experiments.}
Experiments across diverse models under both settings reveal substantial limitations in proactive information seeking, with the best offline and streaming proactive models achieving only 52.18\% and 7.22\% Ask-F1, respectively.
\end{itemize}

\section{Related Work}

\subsection{Offline Paradigm}
Traditional video understanding follows an offline paradigm, in which models answer questions after observing the entire video. Benchmarks such as Video-MME~\citep{DBLP:conf/cvpr/FuDLLRZWZSZCLLZ25}, LVBench~\citep{DBLP:conf/iccv/WangHHCZQDGHXDT25}, LongVideoBench~\citep{DBLP:conf/nips/WuLCL24} and other benchmarks~\citep{zhou2025mlvu,li2024mvbench} focus on visual perception, temporal reasoning, and long-horizon video understanding.

\subsection{Streaming Paradigm}
Streaming video understanding requires models to reason only over video content observed so far. VStream-QA~\citep{zhang2024flash} and StreamingBench~\citep{lin2026streamingbench} established the standards for evaluating early streaming videos. OVO-Bench~\citep{niu2025ovo}, RTV-Bench~\citep{xun2026rtv}, and StreamArena~\citep{zhang2026streamarena}, along with other benchmarks~\citep{lin2026ost,huang2025online,yang2025svbench}, further extend evaluation to dynamic temporal reasoning and long-horizon streaming interaction.
Meanwhile, existing methods mainly improve model performance under the streaming paradigm from three aspects: computational efficiency~\citep{DBLP:conf/cvpr/ChenLWLSGLGMS24,
xu2026streamingvlm,
wang2026accelerating}, GPU memory usage~\citep{chen2026streamkv,
zhang2026hermes,
zhang2024flash}, and long-horizon memory~\citep{wang2026think,
guan2026video,
zhang2026think}.
\subsection{Proactive Paradigm}
Proactive video understanding extends evaluation from \textit{how to answer} to \textit{when to answer}. Early benchmarks~\citep{lin2026streamingbench,niu2025ovo} require models to wait until sufficient evidence is observed, while ProactiveVideoQA~\citep{wang2025proactivevideoqacomprehensivebenchmarkevaluating} and ProReady-QA~\citep{azad2026streamreadylearninganswerlong} further introduce metrics that jointly evaluate answer quality and timing. IPIBench~\citep{li2026ipibenchevaluatinginteractiveproactive} supports dynamic timestamp modification, OmniMMI~\citep{wang2025omnimmi} incorporates audiovisual interaction, and other benchmarks~\citep{lu2026phostream,fu2025vispeak,yan2026proact} cover procedural guidance, mobile interaction, and game commentary. Methodologically, JoyAI-VL-Interaction~\citep{DBLP:journals/corr/abs-2606-14777} and MOSS-VL~\citep{DBLP:journals/corr/abs-2608-15045} improve proactive response capabilities through architectural and training innovations. For proactive information asking, ClearVQA~\citep{jian2025teaching} addresses visual ambiguity through visual feedback, while ProactiveBench~\citep{de2026proactivebench} supports actions such as removing occlusions and changing viewpoints. However, these works mainly focus on static images and do not distinguish different sources of information uncertainty, leaving proactive information asking in streaming video underexplored. \textbf{AskStream} addresses this gap by extending proactive video understanding toward active information asking and jointly modeling video and user uncertainty.

\begin{table}[t]
\centering
\caption{\textbf{Comparison of video understanding benchmarks.}
MC \& OE denote multiple-choice \& open-ended evaluation. Opt., Task, \& Dur. denote average options, task number, and average video duration. TS \& Win. denote timestamps and time windows. FL, MT, \& Pro. denote frame-level annotation, multi-turn interaction, and proactive tasks. \textsuperscript{$\dagger$} denotes values measured by us.}
\label{tab:benchmark_comparison}

\begingroup
\footnotesize
\renewcommand{\arraystretch}{1.04}
\setlength{\tabcolsep}{3.3pt}
\setlength{\aboverulesep}{1.5pt}
\setlength{\belowrulesep}{1.5pt}

\begin{tabularx}{\textwidth}{
l
>{\hsize=1.084\hsize\centering\arraybackslash}X  
>{\hsize=.976\hsize\centering\arraybackslash}X   
>{\hsize=.976\hsize\centering\arraybackslash}X   
>{\hsize=1.084\hsize\centering\arraybackslash}X  
>{\hsize=1.247\hsize\centering\arraybackslash}X  
>{\hsize=.813\hsize\centering\arraybackslash}X   
>{\hsize=.922\hsize\centering\arraybackslash}X   
>{\hsize=.922\hsize\centering\arraybackslash}X   
>{\hsize=.976\hsize\centering\arraybackslash}X   
}
\toprule
\textbf{Benchmark}
& \textbf{Eval.}
& \textbf{Opt.}
& \textbf{Task}
& \textbf{Dur.}
& \textbf{Timing}
& \textbf{FL}
& \textbf{Ask}
& \textbf{MT}
& \textbf{Pro.} \\
\midrule

\multicolumn{10}{l}{\textit{(A) Offline video benchmarks}}\\[-1pt]
Video-MME~\citep{DBLP:conf/cvpr/FuDLLRZWZSZCLLZ25}
& MC & 4 & 12 & 17.0 & -- & \xmark & \xmark & \xmark & \xmark \\

LongVideoBench~\citep{DBLP:conf/nips/WuLCL24}
& MC & 4.73\textsuperscript{$\dagger$} & 17 & 7.9 & -- & \xmark & \xmark & \xmark & \xmark \\

\midrule
\multicolumn{10}{l}{\textit{(B) Streaming video benchmarks}}\\[-1pt]
StreamingBench~\citep{lin2026streamingbench}
& MC & 4 & 18 & 4.1 & TS & \xmark & \xmark & \cmark & \cmark \\

RIVER-Bench~\citep{DBLP:journals/corr/abs-2603-03985}
& ALL & 4 & 3 & 24.1\textsuperscript{$\dagger$}
& TS & \xmark & \xmark & \xmark & \cmark \\

RTV-Bench~\citep{xun2026rtv}
& MC & 3.53\textsuperscript{$\dagger$} & 8 & 18.2 & TS & \xmark & \xmark & \xmark & \xmark \\

StreamArena~\citep{zhang2026streamarena}
& OE & -- & 4 & 88.8 & TS & \xmark & \xmark & \cmark & \cmark \\

\midrule
\multicolumn{10}{l}{\textit{(C) Proactive streaming benchmarks}}\\[-1pt]
OmniMMI~\citep{wang2025omnimmi}
& ALL & -- & 6 & 5.4 & TS & \xmark & \xmark & \cmark & \cmark \\

ProactiveVideoQA~\citep{wang2025proactivevideoqacomprehensivebenchmarkevaluating}
& OE & -- & 4 & 2.1 & Win. & \xmark & \xmark & \cmark & \cmark \\

ProReady-QA~\citep{azad2026streamreadylearninganswerlong}
& OE & -- & 5 & 40.0 & Win. & \xmark & \xmark & \cmark & \cmark \\

IPIBench~\citep{li2026ipibenchevaluatinginteractiveproactive}
& OE & -- & 9 & 1.0\textsuperscript{$\dagger$}
& TS & \xmark & \xmark & \cmark & \cmark \\

\midrule
\rowcolor{gray!13}
\textbf{AskStream (Ours)}
& \textbf{ALL}
& \textbf{9.44}
& \textbf{8}
& \textbf{11.0}
& \textbf{Win.}
& \cmark
& \cmark
& \cmark
& \cmark \\

\bottomrule
\end{tabularx}
\endgroup

\end{table}

\section{AskStream}
\label{AskStream}

\subsection{Task Definition}
\label{Task Definition}

\textbf{Task Modeling.}
We model \textbf{AskStream} as a partially observable Markov decision process (POMDP), where the latent state captures information that is missing from the video stream or the user stream. For each interaction event, the model makes decisions based only on the video observed so far, the user query, and previous interactions. At each decision point, the model chooses one of three actions: \texttt{WAIT}, \texttt{ASK}, or \texttt{ANSWER}.

\textbf{Time Window Annotation.}
We divide a  video stream into multiple events. Event $i$ begins when the current query arrives at $q_i$ and ends when the next query arrives at $q_{i+1}$, defining the event interval as $[q_i,q_{i+1})$. Because the information uncertainty may be recognized at the beginning or as the video unfolds, we define two types of temporal windows:

\begin{itemize}[leftmargin=1.0em, labelsep=0.5em]
\item \textbf{Type I: wait before asking.} In this type of event, the initial observations are insufficient for the model to identify the uncertainty. The model must therefore wait until relevant evidence emerges before asking. Let $a_i$ denote the time at which the model autonomously issues \texttt{ASK},
and let $s_i$ and $e_i$ denote the annotated start and end of the
\texttt{ANSWER} window, respectively. For this type, we annotate only the
\texttt{WAIT} and \texttt{ANSWER} windows:
\begin{equation}
W_i^{\mathrm{wait}} = [q_i, s_i),
\qquad
W_i^{\mathrm{answer}} = [s_i, e_i).
\label{eq1}
\end{equation}
A valid \texttt{ASK} occurs at $a_i \in W_i^{\mathrm{answer}}$.
It ends the model's actual waiting interval $[q_i, a_i)$ and
initiates the answering phase. The model then should answer after
receiving the requested information.

\item \textbf{Type II: ask before waiting.}
In the other type of event, the initial observations are sufficient for the model to identify the uncertainty. The model should issue \texttt{ASK} within the annotated \texttt{ASK} window and
then wait until relevant evidence for answering emerges.
For this type, we annotate only the \texttt{ASK} and \texttt{ANSWER} windows:
\begin{equation}
W_i^{\mathrm{ask}} \subseteq [q_i, s_i),
\qquad
W_i^{\mathrm{answer}} = [s_i, e_i).
\label{eq2}
\end{equation}
For a valid \texttt{ASK} at $a_i \in W_i^{\mathrm{ask}}$, the model's
actual waiting interval is $[a_i, s_i)$, whose start is determined
dynamically by the \texttt{ASK} action rather than manually annotated.
\end{itemize}
\textbf{Action Evaluation.}
Let ${\omega}_{i,k}$ denote the time of the $k$-th WAIT action, and let $a_i$ and ${\rho}_i$ denote the times when the model issues \texttt{ASK} and \texttt{ANSWER} for event $i$, respectively. ${\omega}_{i,k}$ indicates that the model chooses to observe further and ${\rho}_i$ provides the final response. In our main setting, each turn allows at most one \texttt{ASK} and \texttt{ANSWER}. The model is allowed to make one action selection per second. A correct \texttt{ASK} triggers the user to provide the missing information. A turn is successful only if the model issues a correct ask within $W_i^{\mathrm{answer}}$ for Type I and $W_i^{\mathrm{ask}}$ for Type II, and subsequently submits the correct answer within $W_i^{\mathrm{answer}}$.
Otherwise, the turn is marked as failed and the model proceeds
to the next query.

\begin{figure}[t]
    \centering
    \includegraphics[width=\linewidth]{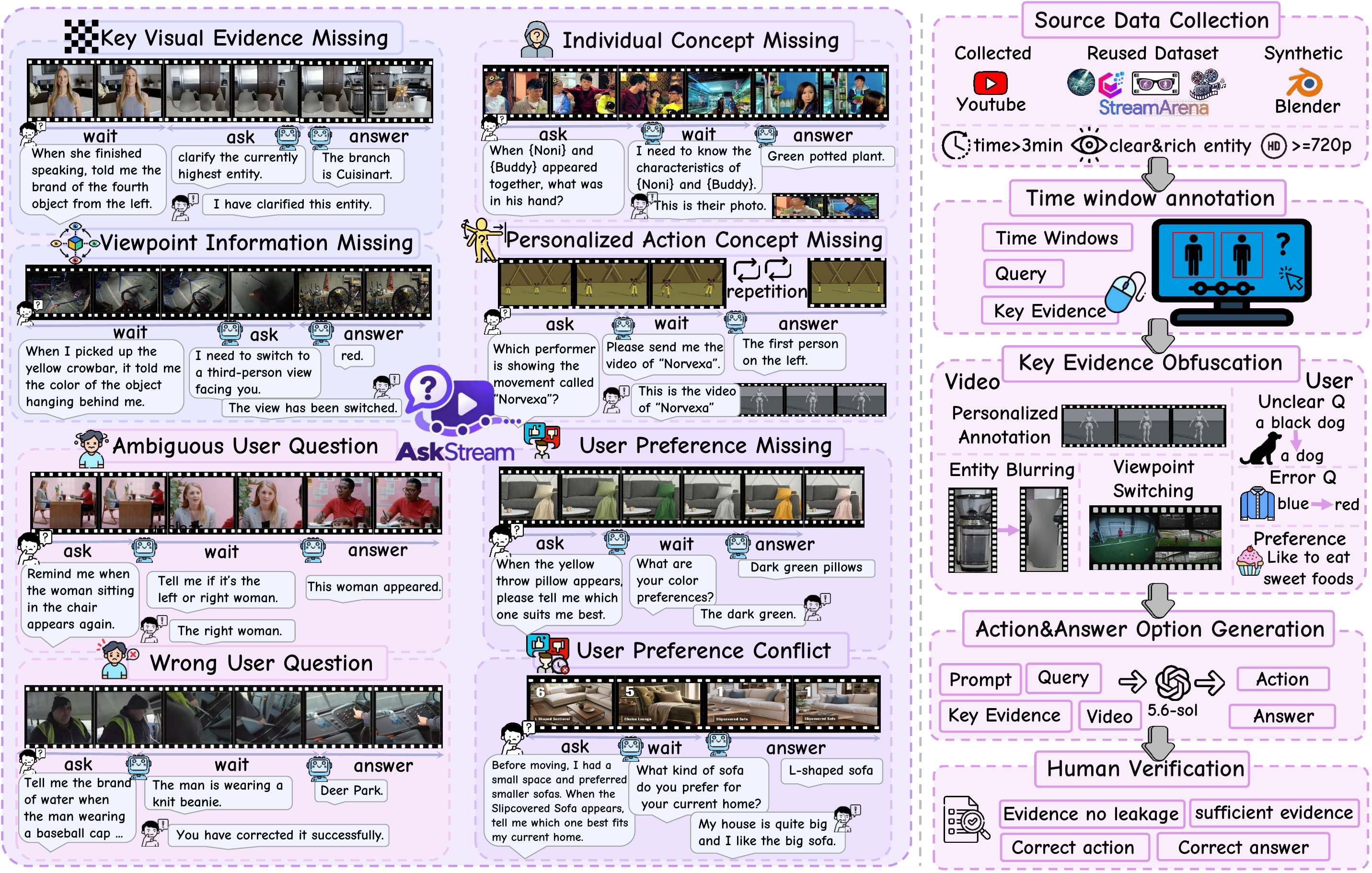}
    \caption{\textbf{AskStream task taxonomy and benchmark construction.}
    \textbf{Left:} eight subtasks spanning video and user uncertainty, with examples of \textit{wait}, \textit{ask}, and \textit{answer} interactions.
    \textbf{Right:} the five-stage construction pipeline.}
    \label{fig3}

\end{figure}

\subsection{ASK-F1}

Inspired by prior timing-aware evaluation~\citep{azad2026streamreadylearninganswerlong,wang2025proactivevideoqacomprehensivebenchmarkevaluating}, 
we introduce \textbf{Ask-F1} to jointly evaluate the temporal and content correctness of both \texttt{ASK} and \texttt{ANSWER}.
For turn $i$, $a_i$ and $\rho_i$ denote the \texttt{ASK} and \texttt{ANSWER} times of the model, 
$W_i^a=[s_i^a,e_i^a)$ and $W_i^\rho=[s_i^\rho,e_i^\rho)$ their valid windows, 
and $\mathbb{I}_i^a$ and $\mathbb{I}_i^\rho$ indicate whether the \texttt{ASK} and \texttt{ANSWER} contents are correct. We further define
$\mathbb{I}_i^{\mathrm{ask}}=\mathbb{I}_i^a\mathbb{I}[a_i\in W_i^a]$,
which equals 1 only when the \texttt{ASK} content is correct and the \texttt{ASK} is issued within the valid \texttt{ASK} window. Accordingly,
$N_{\rm ask}=\sum_i \mathbb{I}_i^{\mathrm{ask}}$
denotes the total number of such correct \texttt{ASK} turns across all turns.
We define
\begin{equation}
\resizebox{\dimexpr\linewidth-3em\relax}{!}{$\displaystyle
T(t;[s,e))
=
\mathbb{I}[s \le t < e]
\left(1-\frac{t-s}{e-s}\right),
\quad
S_{\rm ask}
=
\frac{1}{N}\sum_i
\mathbb{I}_i^a T(a_i;W_i^a),
\quad
S_{\rm ans}
=
\begin{cases}
\displaystyle
\frac{1}{N_{\rm ask}}\sum_i
\mathbb{I}_i^{\rm ask}\mathbb{I}_i^\rho
T(\rho_i;W_i^\rho),
& N_{\rm ask}>0,\\[2pt]
0, & N_{\rm ask}=0.
\end{cases}
$}
\label{eq:ask-scores}
\end{equation}
\begin{equation}
\mathrm{Ask\text{-}F1}
=
\begin{cases}
\displaystyle
\frac{2S_{\rm ask}S_{\rm ans}}
{S_{\rm ask}+S_{\rm ans}},
& S_{\rm ask}+S_{\rm ans}>0,\\[3pt]
0, & S_{\rm ask}=S_{\rm ans}=0.
\end{cases}
\label{eq:ask-f1}
\end{equation}

$S_{\rm ask}$ and $S_{\rm ans}$ represent the timing-aware scores for asking and answering, respectively, while $N$ denotes the total number of turns. Further analysis of the metric design is provided in Appendix~\ref{app:metric}.

\subsection{Task Taxonomy}

The uncertainties in the real world may arise from videos or users. Accordingly, we divide our benchmark into \textbf{video uncertainty} and \textbf{user uncertainty}, which include four tasks and eight subtasks, as illustrated in Figure~\ref{fig3}.

\subsubsection{Video Uncertainty}

Video uncertainty arises at two levels: either the visual evidence required by the question is missing, or the visual evidence is complete but the relevant personalized concepts are absent. These correspond to explicit and implicit video information uncertainty, respectively.

\textbf{Explicit Video Information Uncertainty}: (1) \textit{Key Visual Evidence Missing (KVEM)}:
When the key entity in the initial video is blurred, the model must recognize that the evidence is insufficient and proactively request a video showing the entity clearly. It then answers the question using the clear video provided by the user.
(2) \textit{Viewpoint Information Missing (VIM)}:
When the initial viewpoint lacks evidence needed to answer the question, the model must identify this gap and proactively request an alternative viewpoint that reveals the missing evidence. It then answers after the user switches to the requested view.

\textbf{Implicit Video Information Uncertainty}:
(1) \textit{Individual Concept Missing (ICM)}:
Given a clear video, the model lacks the personalized concept information about a visible entity needed to answer the question. It must recognize this memory gap and proactively request the relevant information from the user before answering.
(2) \textit{Personalized Action Concept Missing (PACM)}:
Although a clear video fully captures a user-specific action, the model lacks the personalized conceptual knowledge needed to answer the question. It must recognize this gap and proactively request the action concept's definition from the user before answering.

\subsubsection{User Uncertainty}

User uncertainty concerns accurately understanding the user’s request or obtaining the current preferences, corresponding to user intent uncertainty and user preference uncertainty, respectively.

\textbf{User Intent Uncertainty}:
(1) \textit{Ambiguous User Question (AUQ)}:
When a question leaves the user's intent unclear, the model must recognize the ambiguity while observing the video, proactively request clarification from the user, and then answer the clarified question. (2) \textit{Wrong User Question (WUQ)}:
When a question contains a false premise about the video, the model must detect its inconsistency with the visual evidence while observing the video, correct the premise, and answer the question.

\textbf{User Preference Uncertainty}:
(1) \textit{User Preference Missing (UPM)}: When answering depends on unknown user preferences, such as style or interests, the model must proactively ask for the relevant preferences and combine the user's feedback with visual evidence to provide a personalized answer. (2) \textit{User Preference Conflict (UPC)}:
Given a video, historical user preferences, and a question requiring updated preferences, the model must recognize that the existing preferences are outdated, proactively ask for the user's current preferences, and answer using the updated information.

\subsubsection{Interaction Timing}

Explicit video requires observing relevant evidence before uncertainty can be identified, following the recognition timing in ~\eqref{eq1}.
Implicit video and user preference uncertainty can be identified from the outset, so the model asks first and then waits for relevant evidence before answering in ~\eqref{eq2}. User intent uncertainty spans both timing patterns, depending on whether the ambiguity can be recognized from the initial query or only after observing relevant visual evidence.

\subsection{Benchmark Construction}

We construct \textbf{AskStream} through the five-stage pipeline in Figure~\ref{fig3}. (1) We collect videos from YouTube, existing datasets and synthesize additional videos using Blender. (2) Using a self-built annotation webpage in Appendix~\ref{Data_Construction}, we then manually annotate user queries, time windows, key entities, and the evidence required to answer each query. Time windows are annotated with frame-level precision to provide timing supervision for \textit{wait}, \textit{ask}, and \textit{answer}. (3) We use Grounded SAM 2~\citep{ren2024grounded} to segment and track entities for selective Gaussian blurring and PySceneDetect to assist video splitting. We also change viewpoints, modify queries, and withhold or alter personalized concepts and preferences. (4) We provide the resulting data to GPT-5.6 Sol~\citep{openai2026gpt56} to generate action and answer options. (5) We manually verify each instance, checking that humans cannot uniquely determine the answer without the missing evidence or clarification but can do so once it is provided, and that each action and answer candidate set contains exactly one correct option. The annotation webpage and task-specific screening, annotation, and verification details are provided in Appendix~\ref{Data_Construction}.

\subsection{Benchmark Statistics}

\begin{figure}[t]
    \centering

    \begin{minipage}[t]{0.50\linewidth}
        \vspace{0pt}
        \centering

        \includegraphics[
            width=\linewidth,
            height=7.6cm,
            keepaspectratio
        ]{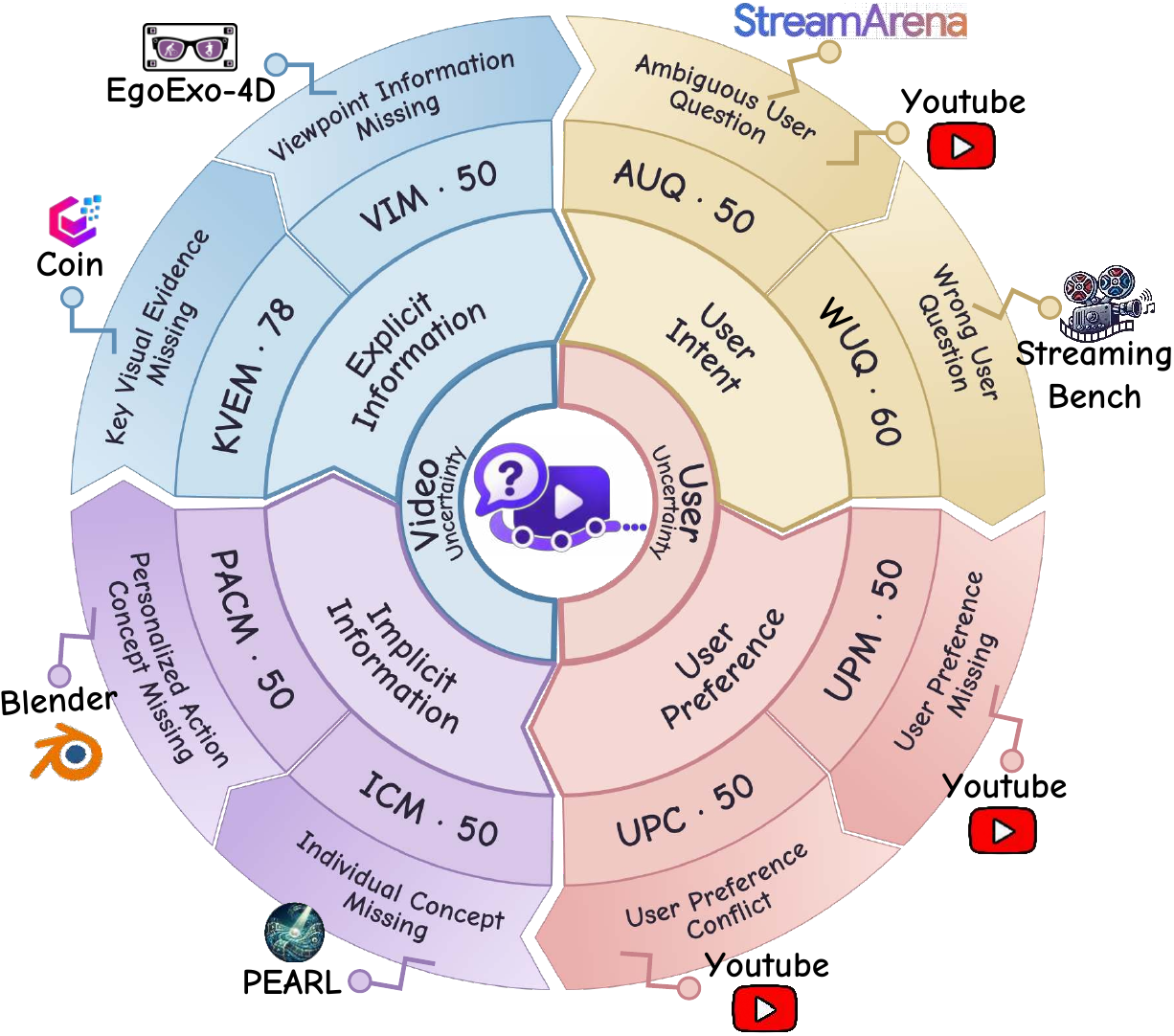}

        \par{\footnotesize (a)}
    \end{minipage}%
    \hfill
    \begin{minipage}[t]{0.48\linewidth}
        \vspace{0pt}
        \centering

        \includegraphics[
            width=\linewidth,
            height=4cm,
            keepaspectratio
        ]{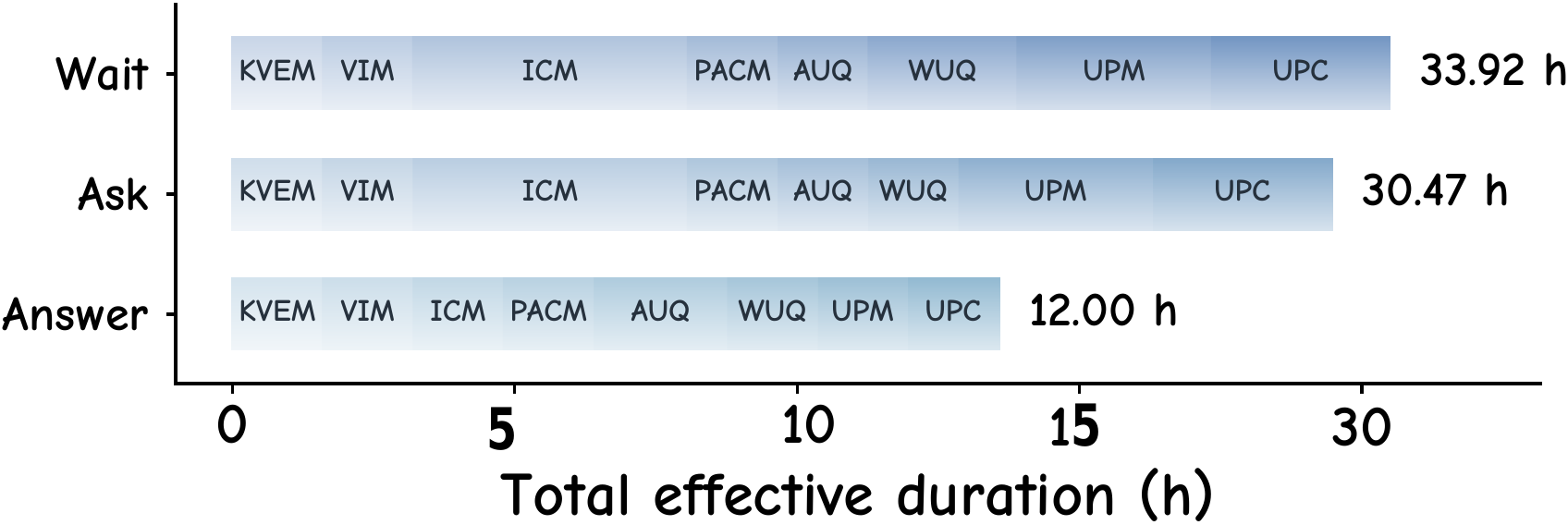}

        \par{\footnotesize (b)}

        \vspace{0.35cm}

        \includegraphics[
            width=\linewidth,
            height=4cm,
            keepaspectratio
        ]{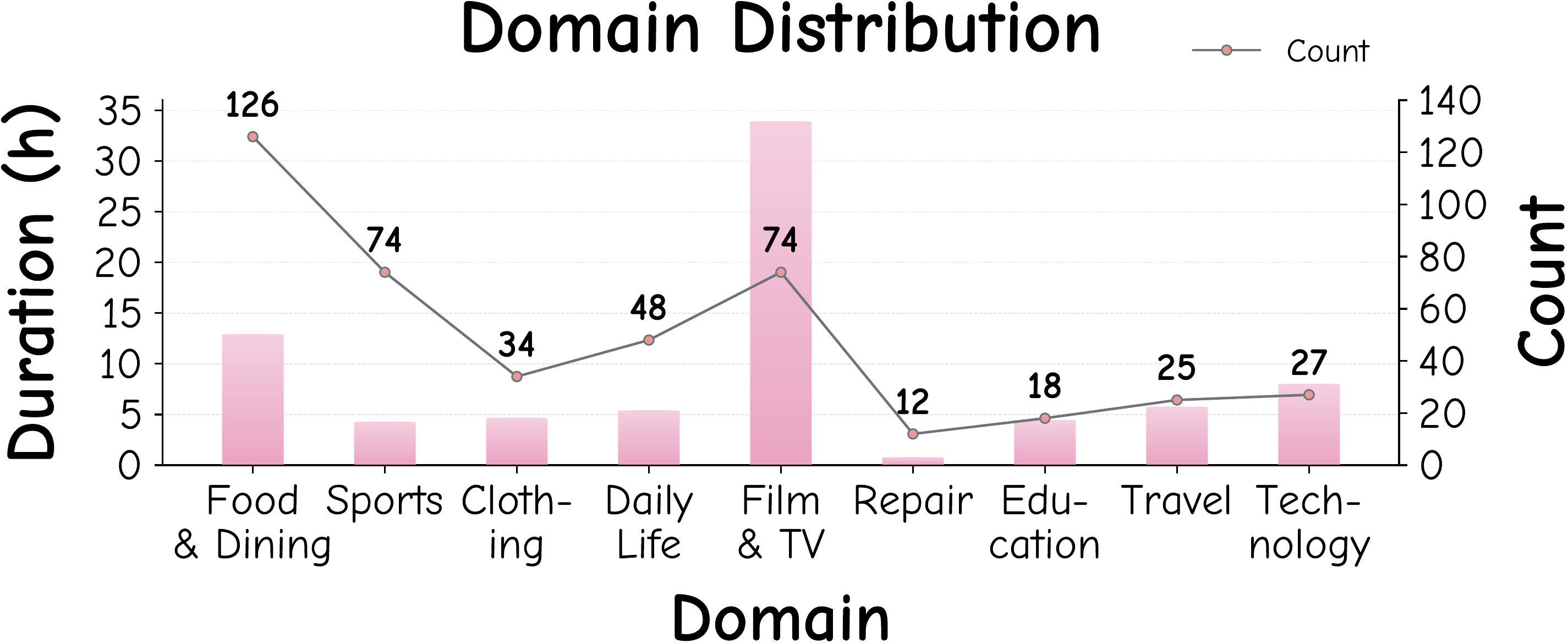}

        \par{\footnotesize (c)}
    \end{minipage}

    \caption{\textbf{Statistics of AskStream.}
    (a) Task taxonomy, data distribution, and sources.
    (b) Total effective durations of \textit{wait}, \textit{ask}, and \textit{answer} across subtasks.
    (c) Video durations and sample counts across domains.
    The benchmark averages 3.2 interaction turns per video and contains 2714 manually annotated timestamps in total.}
    \label{fig4}
\end{figure}

Figure~\ref{fig4} summarizes the main statistics of AskStream, which comprises 438 videos and  1,401 questions across eight subtasks. The videos span nine domains, with durations ranging from 16.72 seconds to 118 minutes, covering diverse streaming scenarios. In addition, the total durations of the three types of temporal windows, \textit{wait}, \textit{ask}, and \textit{answer}, are 33.92, 30.47, and 12.00 hours, respectively.

\section{Experiments}
\label{Experiments}

\subsection{Setup}
\subsubsection{Baseline}
We evaluate three groups of models on \textbf{AskStream}: (A) Offline Closed-source Models, including GPT-5.6-Sol~\citep{openai2026gpt56}, Gemini-3.8 Flash~\citep{google2026gemini38flash}, and Claude Sonnet 5~\citep{anthropic2026claudesonnet5}; (B) Offline Open-Source Models, including Qwen3.8-flash~\citep{qiu2026designqwen38nextarchitectureevaluation}, MiMo-V2.5~\citep{xiaomimimoteam2026fullpipelineinferenceoptimizationmimov25}, Kimi K2.6~\citep{kimiteam2026kimik25visualagentic}, and DeepSeek-V4.1-Flash~\citep{deepseekai2026deepseekv41flashpushinglimitskv}; and (C) Streaming Models with Proactive Capabilities, including VideoLLM-online~\citep{DBLP:conf/cvpr/ChenLWLSGLGMS24}, TimeChat-Online~\citep{DBLP:conf/mm/YaoLW0RLOWLLK0Z25}, MMDuet2~\citep{DBLP:journals/corr/abs-2512-06810}, MiniCPM-o 4.5~\citep{DBLP:journals/corr/abs-2604-27393}, AURA~\citep{DBLP:journals/corr/abs-2604-04184}, JoyAI-VL-Interaction~\citep{DBLP:journals/corr/abs-2606-14777}, and MOSS-VL-Realtime~\citep{DBLP:journals/corr/abs-2608-15045}. We exclude streaming models without proactive capabilities because they are unable to respond to proactive tasks.

\subsubsection{Evaluation protocols}
Existing streaming benchmarks~\citep{lin2026streamingbench,niu2025ovo}
often evaluate offline models through sparse uniform frame sampling, which may
miss the dense temporal information required in streaming scenarios. Sliding
window methods~\citep{shen2026simple} preserve recent frames but discard earlier video context.
We therefore introduce \textbf{StreamHarness}, a hierarchical memory
framework. It retains the most recent 10 frames (1 fps) as short memory,
summarizes every 10 frames into one mid-term memory, consolidates every four
mid-term memories into one long-term memory block, and retains up to eight
long-term blocks. Native streaming models instead use their own memory mechanisms. Across multiple turns, only shared video memory is preserved.
Questions, options, and action histories are discarded after each turn. The model is evaluated at one action opportunity per second, with a tolerance of ±1 second. To more realistically evaluate models' proactive capabilities, we adopt both \textbf{multiple-choice} and \textbf{open-ended} settings. In the open-ended setting, we use Gemini-3.8 Flash as the judge model to assess the correctness of both the model's \texttt{ASK} and \texttt{ANSWER} responses. Please see section~\ref{Task Definition} to check the definition of the correct turn.

\subsection{Main Results}
\label{main_results}
\begin{wrapfigure}{r}{0.42\textwidth}
    \vspace{-8pt}
    \centering

    \includegraphics[
        width=\linewidth
    ]{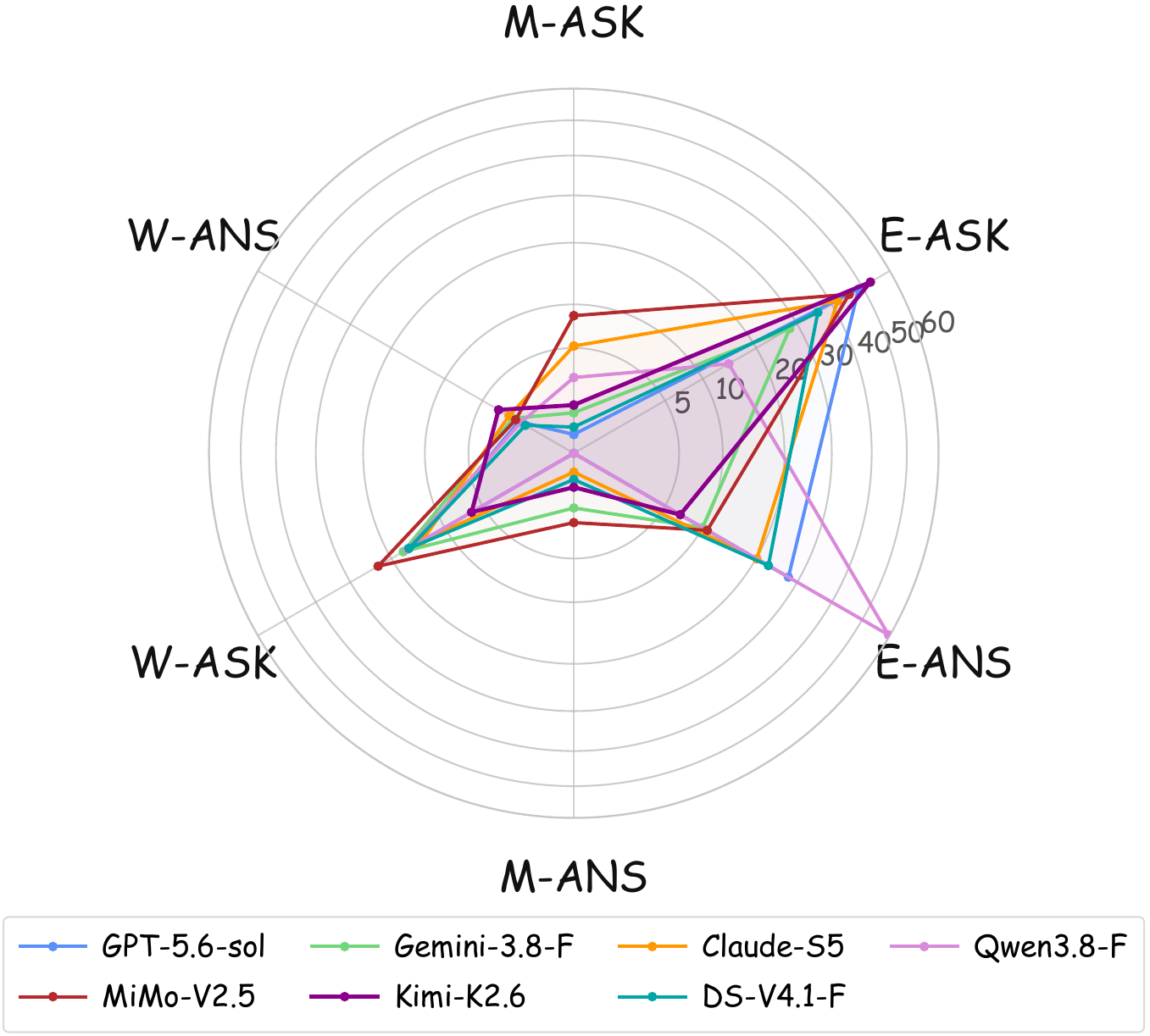}

    \captionsetup{
        font=footnotesize,
        skip=2pt
    }

    \caption{
        \textbf{Error-type distributions} of offline
        models in the multiple-choice setting.
        Each vertex shows the percentage of the
        corresponding error type.
    }
    \label{fig:two_right}
    \vspace{-5pt}
\end{wrapfigure}

Table~\ref{tab:askstream_main_results} presents the main results under the multiple-choice setting, while results for the open-ended setting are provided in Appendix~\ref{app:open_ended_results}. \textbf{(1)} Among offline models, Gemini-3.8 Flash achieves relatively strong performance on most subtasks, reaching 37.17\%, 43.10\%, and 81.36\% on KVEM, WUQ, and UPC, with an overall Ask-F1 score of \textbf{52.18\%}. However, performance differences across different types of uncertainty, and for stronger models, $S_{\mathrm{ask}}$ is significantly lower than $S_{\mathrm{ans}}$, indicating that \texttt{ASK} is the bottleneck. \textbf{(2)} In contrast, streaming proactive models perform significantly worse: they score zero on most subtasks, and most models fail to complete the task. Even AURA, which performs relatively better, achieves a final score of only 7.22\%. This indicates that the current proactive model lacks the ability to \texttt{ASK}. \textbf{(3)} We further analyze different types of model errors. \texttt{E-ASK/E-ANS}, \texttt{W-ASK/W-ANS}, and \texttt{M-ASK/M-ANS} denote early, wrong, and missed \texttt{ASK/ANSWER} actions, respectively. Figure~\ref{fig:two_right} shows that \textbf{early ASK } is the main error types across offline models. This indicates that models often recognize the need for additional information, but tend to either ask too early, suggesting that \textit{when to ask} remain major bottlenecks. Overall, while current MLLMs demonstrate emerging proactive information seeking capabilities, their performance remains constrained by unreliable \texttt{ASK} decisions.

\begin{table}[t]
\centering
\caption{Main results on AskStream in the multiple-choice setting. Scores reported for individual subtasks are Ask-F1 scores. Bold denotes the best performance on each subtask.}
\label{tab:askstream_main_results}

\begingroup

\fontsize{8}{9.6}\selectfont
\setlength{\tabcolsep}{2.4pt}
\renewcommand{\arraystretch}{1.22}

\setlength{\arrayrulewidth}{0.35pt}
\setlength{\aboverulesep}{0pt}
\setlength{\belowrulesep}{0pt}
\setlength{\doublerulesep}{1.0pt}

\definecolor{askheader}{RGB}{224,224,224}
\definecolor{askrow}{RGB}{240,240,240}
\definecolor{askgroup}{RGB}{125,125,125}
\definecolor{askbest}{RGB}{217,245,254}
\definecolor{askbestline}{RGB}{124,194,204}
\definecolor{askgainred}{RGB}{190,75,75}

\arrayrulecolor{black!75}

\newcommand{\askscoregain}[2]{%
  \mbox{%
    \textbf{#1}\hspace{0.15em}%
    \raisebox{0.45ex}{%
      \textcolor{askgainred}{%
        \scalebox{0.70}{$\uparrow$#2}%
      }%
    }%
  }%
}

\newcommand{\asksep}{%
  \arrayrulecolor{black!22}%
  \cdashline{1-13}[0.35pt/1.25pt]%
  \arrayrulecolor{black!75}%
}

\resizebox{\linewidth}{!}{%
\begin{tabular}{l||c|*{4}{c}|*{4}{c}|ccc}

\toprule[0.9pt]

\rowcolor{askheader}
& &
\multicolumn{4}{c|}{\textbf{Video}} &
\multicolumn{4}{c|}{\textbf{User}} &
\multicolumn{3}{c}{} \\

\rowcolor{askheader}
& &
\multicolumn{2}{c}{Explicit} &
\multicolumn{2}{c|}{Implicit} &
\multicolumn{2}{c}{Intent} &
\multicolumn{2}{c|}{Preference} &
\multicolumn{3}{c}{%
  \multirow{-2}{*}{\textbf{Metric}}%
} \\

\rowcolor{askheader}
\multicolumn{1}{c||}{%
  \multirow{-3}{*}{\textbf{Method}}%
} &
\multirow{-3}{*}{\textbf{Backbone}} &
KVEM & VIM & ICM & PACM &
AUQ & WUQ & UPM & UPC &
$S_{\mathrm{ask}}$ &
$S_{\mathrm{ans}}$ &
\textbf{Ask-F1} \\

\hline\hline

\multicolumn{13}{l}{%
  \textcolor{askgroup}{%
    \textit{Offline closed-source models}%
  }%
} \\

GPT-5.6-Sol & -
& 18.10 & 2.93 & 3.05 & 0
& 15.69 & 17.50 & 18.60 & 26.55
& 18.11 & 23.62 & 20.50 \\
\asksep

\rowcolor{askrow}
Claude Sonnet 5 & -
& 21.48 & 13.01 & 7.60 & 9.70
& 13.28 & 29.42 & 7.93 & 29.53
& 20.87 & 24.37 & 22.48 \\

\arrayrulecolor{askbestline}
\hline
\rowcolor{askbest}
\textbf{Gemini-3.8 Flash} & -
& \textbf{37.17}
& 17.46
& \textbf{33.26}
& \textbf{17.73}
& \textbf{31.13}
& \textbf{43.10}
& \textbf{64.03}
& \textbf{81.36}
& \textbf{44.63}
& \textbf{62.80}
& \textbf{52.18} \\
\hline
\arrayrulecolor{black!75}

\multicolumn{13}{l}{%
  \textcolor{askgroup}{%
    \textit{Offline open-source models}%
  }%
} \\

MiMo-V2.5 & -
& 28.49 & 3.73 & 16.16 & 0
& 11.48 & 11.83 & 28.66 & 26.18
& 10.48 & 24.38 & 14.66 \\
\asksep

\rowcolor{askrow}
Kimi K2.6 & -
& 23.41
& \textbf{21.56}
& 9.27
& 0
& 16.74
& 34.40
& 0
& 30.69
& 19.35
& 14.41
& 16.52 \\
\asksep

Qwen-3.8 Flash & -
& 21.03 & 15.81 & 1.29 & 9.94
& 6.28 & 23.20 & 4.64 & 0
& 31.75 & 12.17 & 17.60 \\

\arrayrulecolor{askbestline}
\hline
\rowcolor{askbest}
\textbf{DeepSeek-V4.1-Flash} & -
& 17.68 & 2.33 & 6.27 & 0
& 17.97 & 19.88 & 6.32 & 15.27
& 18.64 & 20.87 & 19.69 \\
\hline
\arrayrulecolor{black!75}

\multicolumn{13}{l}{%
  \textcolor{askgroup}{%
    \textit{Streaming proactive models}%
  }%
} \\

MOSS-VL-Realtime & Qwen3-8B
& 0 & 0 & 0 & 0
& 0 & 0 & 0 & 0
& 0.74 & 0 & 0 \\
\asksep

\rowcolor{askrow}
MMDuet2 & Qwen2.5-VL-3B
& 0 & 0 & 0 & 0
& 0 & 0 & 0 & 0
& 0 & 0 & 0 \\
\asksep

VideoLLM-online & Llama-3-8B
& 0 & 0 & 0 & 0
& 0 & 0 & 0 & 0
& 0 & 0 & 0 \\
\asksep

\rowcolor{askrow}
MiniCPM-o 4.5 & Qwen3-8B
& 6.76 & 0 & 0 & 0
& 0 & 0 & 0 & 0
& 7.17 & 0.06 & 0.12 \\
\asksep

TimeChat-Online & Qwen2.5-VL-7B
& 9.29 & 0 & 0 & 0
& 0 & 4.15 & 0 & 0
& 12.58 & 2.51 & 4.18 \\
\asksep

\rowcolor{askrow}
JoyAI-VL-Interaction & JoyAI-VL-1.0
& 5.73 & 0 & 0 & 0
& 0 & 4.03 & 0 & 0
& 7.54 & 3.64 & 4.91 \\

\arrayrulecolor{askbestline}
\hline
\rowcolor{askbest}
\textbf{AURA} & Qwen3-VL-8B
& 8.31 & 4.35 & 0 & 0
& 5.26 & 2.93 & 0 & 0
& 10.85 & 5.41 & 7.22 \\
\hline
\arrayrulecolor{black!75}

\bottomrule[0.9pt]
\end{tabular}%
}

\endgroup

\end{table}

\WFclear
\subsection{Analysis experiment}

\subsubsection{Where Do Models Fail?}

\begin{figure}[H]
    \centering
    \begin{minipage}[t]{0.31\linewidth}
        \centering
        \includegraphics[width=\linewidth]{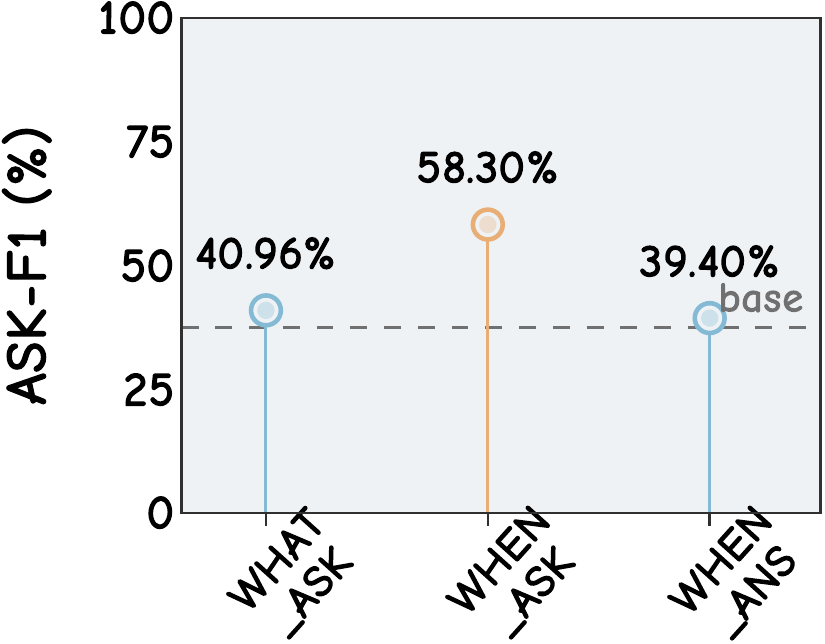}\\[-0.3em]
        {\small (a)}
    \end{minipage}
    \hfill
    \begin{minipage}[t]{0.67\linewidth}
        \centering
        \raisebox{-5mm}{\includegraphics[width=\linewidth]{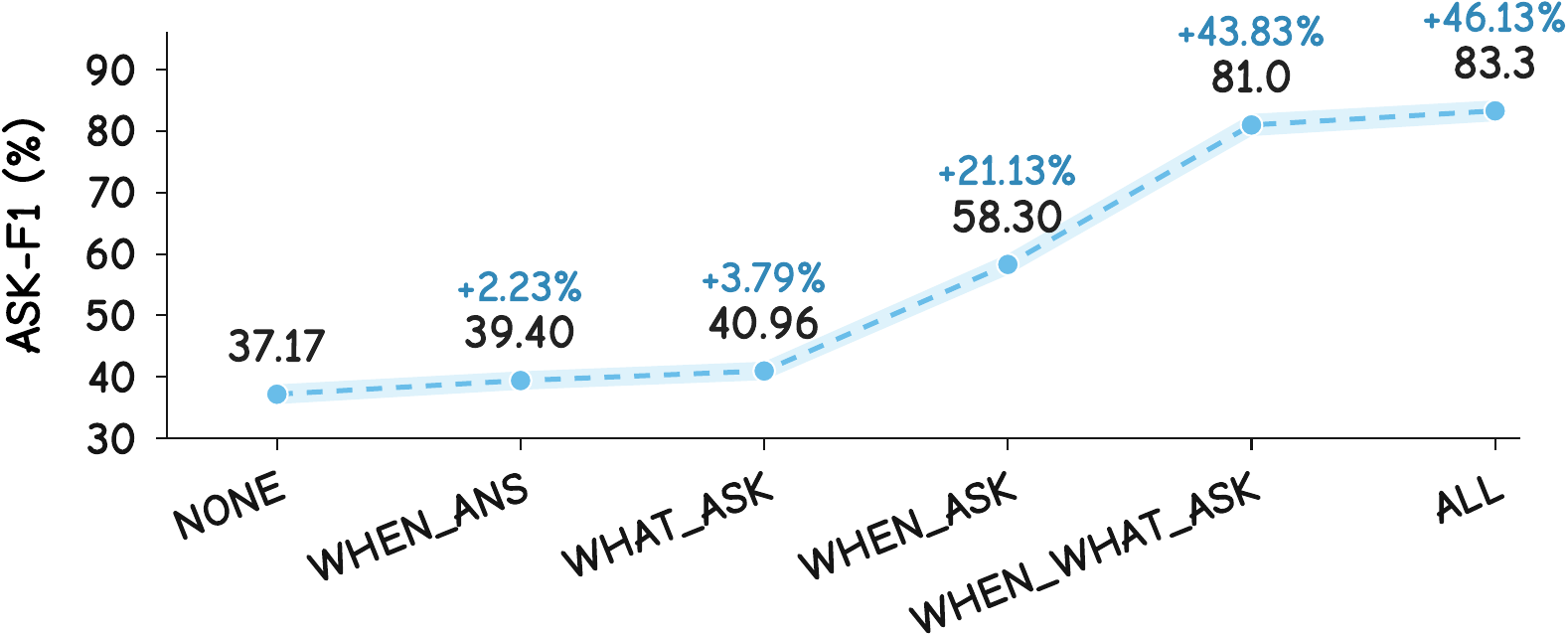}}\\[-0.3em]
        {\small (b)}
    \end{minipage}

    \caption{
    \textbf{Oracle ablation of decision stages.}
    Results on KVEM using Gemini-3.8 Flash in the multiple-choice setting.
    (a) Effect of different oracle signals.
    (b) ASK-F1 under different amounts and combinations of oracle information.
    }
    \label{fig6}

\end{figure}

To understand how decisions at different stages affect the model’s final performance, we design an ablation study that gradually adds oracle information. We remove uncertainty about when to \texttt{ASK}, what to \texttt{ASK}, and when to \texttt{ANSWER}, as well as both \texttt{ASK} factors and all three factors, and measure how much performance can be recovered. Figure~\ref{fig6}(a) shows that providing oracle information about when to \texttt{ASK} leads to the largest performance improvement, raising Ask-F1 from 37.17\% to 58.30\% (+21.13 points), further supporting the finding in Section~\ref{main_results} that \textit{when to ask} is the main bottleneck. Figure~\ref{fig6}(b) compares different amounts and combinations of oracle information, showing that performance generally improves as more decision-stage uncertainty is removed.


\subsubsection{The Impact of Contextual Memory}
In the main setting, we do not keep interaction history across turns so that each turn can be evaluated independently. We further study whether keeping historical context helps with different types of uncertainty. As shown in Figure~\ref{fig7}(a), contextual memory has little effect on \textbf{video uncertainty}, because the model mainly relies on the visual evidence in the current turn. In contrast, it greatly improves performance on \textbf{user uncertainty}, with gains of 4.40 and 5.38 points on AUQ and UPM, respectively, because user intent and preferences often continue across turns, and previous interactions provide useful information for understanding the current request. These results show that the benefit of contextual memory strongly depends on the source of uncertainty.

\subsubsection{The Effect of Ask Budget}
\label{ask_budget}
\begin{figure}[t]
    \centering
    \begin{minipage}[t]{0.35\linewidth}
        \centering
        \includegraphics[width=\linewidth]{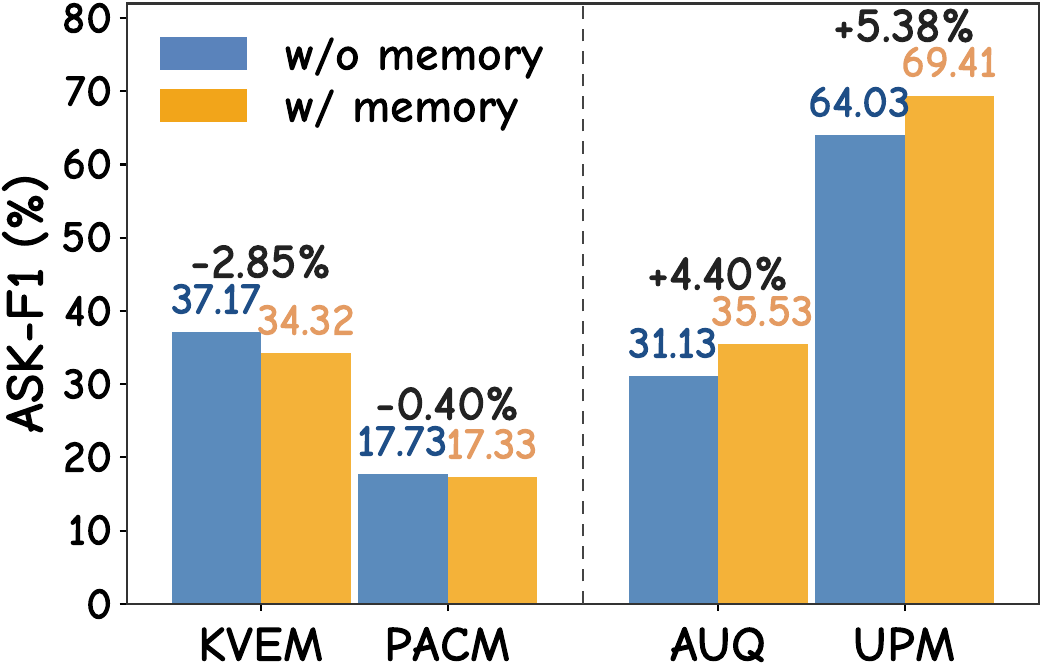}\\[-0.5em]
        {\footnotesize (a)}
    \end{minipage}\hfill
    \begin{minipage}[t]{0.64\linewidth}
        \centering
        \includegraphics[width=\linewidth]{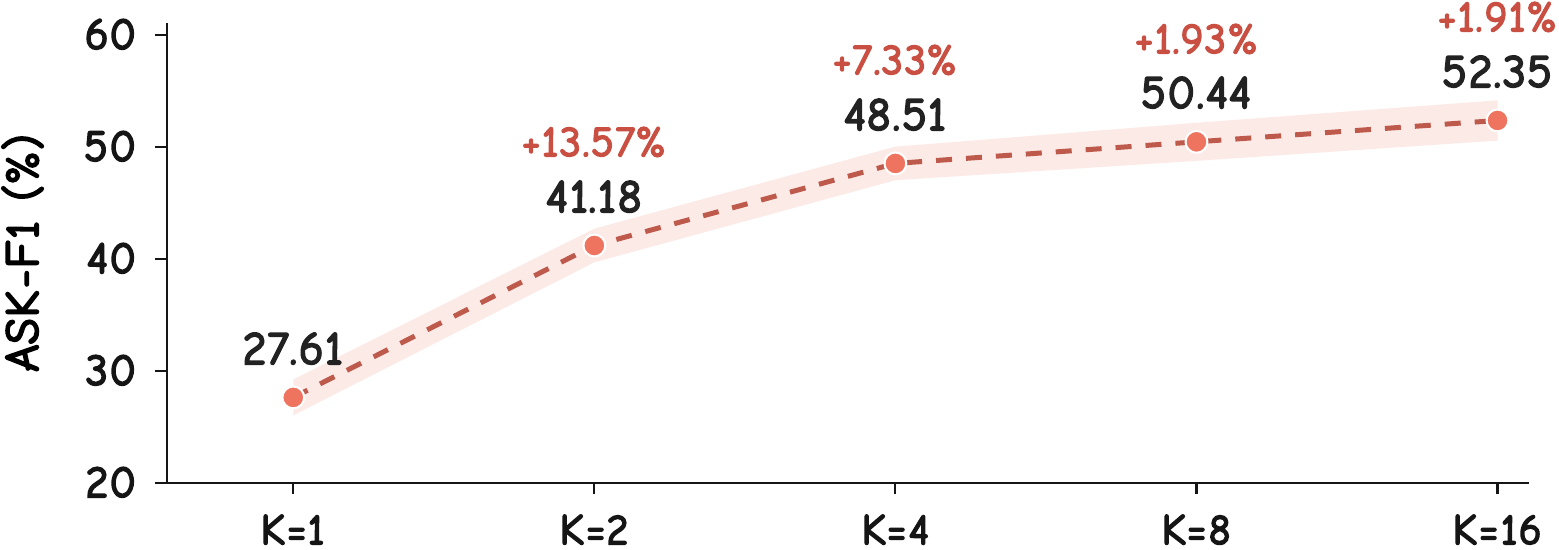}\\[-0.5em]
        {\footnotesize (b)}
    \end{minipage}\hfill

    \caption{
    \textbf{Analysis Experiments.} 
    (a) Impact of contextual memory. We use Gemini-3.8 Flash under the multiple-choice setting on four subtasks, KVEM, PACM, AUQ, and UPM, to analyze the effect of cross-turn contextual memory.
    (b) Impact of ASK budget. We evaluate Gemini-3.8 Flash on KVEM under the open-ended setting to analyze the effect of different ASK budgets.
    }
    \label{fig7}

\end{figure}

In real-world interactions, a model may struggle to obtain sufficient information through a single ask. Therefore, we further investigate whether increasing the ASK budget can improve the model's proactive information seeking capability. As shown in Figure~\ref{fig7}(b), performance improves as the ASK budget increases from 1 to 16 but gradually saturates at around 52.35\%. This suggests that more ASK opportunities are helpful, but the improvement is limited.

\section{Conclusion}
We introduce \textbf{AskStream}, a benchmark for proactive information seeking under uncertainty in streaming video understanding. By extending the proactive paradigm with an explicit \textit{ask} action, AskStream evaluates when and what models should ask before answering across eight subtasks covering video and user uncertainty. We further introduce frame-level action windows, Ask-F1, and StreamHarness to support timing-aware evaluation under both settings. Experiments show that current MLLMs remain limited in proactive information seeking, with the main bottleneck lying in determining \textit{when and what to ask}.


\clearpage

\bibliography{askstream_references}
\bibliographystyle{askstream_bibstyle}

\appendix
\clearpage

\refstepcounter{section}
\addcontentsline{toc}{section}{Appendix}

\begin{center}
    {\Large\bfseries
    AskStream: When and What to Ask Proactively\\
    under Uncertainty in Streaming Video Understanding\par}

    \vspace{0.45em}

    {\large\bfseries Supplementary Material\par}
\end{center}

\subsection{Limitations}
AskStream still simplifies several aspects of real-world proactive interaction. First, the current setting allows at most one \textit{ASK} action per turn, while practical scenarios may require repeated clarification, especially when multiple entities are simultaneously ambiguous or when one ASK reveals further missing information. Second, our benchmark focuses on streaming video understanding and does not consider embodied agents that must jointly coordinate perception, information seeking, and physical actions, such as vision-language-action (VLA) models. Third, each interaction is initiated by an explicit user query, whereas more autonomous agents should be able to proactively detect information insufficiency even without being directly asked and decide whether clarification is needed. Finally, although AskStream introduces frame-level temporal windows, real-world uncertainty may evolve over longer and more complex horizons, with overlapping, discontinuous, or dynamically changing opportunities for ASK and ANSWER. Extending the benchmark along these directions would enable more realistic evaluation of proactive information seeking.

\subsection{Metric Design}
\label{app:metric}

In this section, we further discuss the design of \textbf{Ask-F1}
and analyze its sensitivity to different temporal weighting functions.

\paragraph{Why harmonic aggregation?}

AskStream evaluates two coupled stages of proactive information seeking:
(1) whether a model can correctly and timely acquire missing information
through an \texttt{ASK} action, and
(2) whether it can subsequently provide a correct and timely
\texttt{ANSWER} after the requested information becomes available.
We measure these two capabilities using $S_{\mathrm{ask}}$ and
$S_{\mathrm{ans}}$, respectively, and aggregate them using

\begin{equation}
\mathrm{Ask\text{-}F1}
=
\frac{
2S_{\mathrm{ask}}S_{\mathrm{ans}}
}{
S_{\mathrm{ask}} + S_{\mathrm{ans}}
}.
\label{eq:appendix_ask_f1}
\end{equation}

We use the term \textit{F1} to denote an \textbf{F1-style harmonic
aggregation}, rather than the conventional classification F1 defined
from precision and recall.
The harmonic mean is suitable for our setting because successful proactive
interaction requires both information acquisition and subsequent
answering to perform well.
A model should not obtain a high overall score if it answers accurately
only after the missing information has been provided but rarely identifies
when and what to ask.
Conversely, issuing appropriate ASK actions is insufficient if the model
cannot subsequently answer the user's query.
Compared with an arithmetic mean, the harmonic mean more strongly
penalizes imbalance between these two stages.

\paragraph{Why is $S_{\mathrm{ans}}$ conditional on a correct ASK?}

We define $S_{\mathrm{ans}}$ over turns in which the model has
successfully issued a correct ASK.
This conditioning is intentional because it separates failures in
\textit{information acquisition} from failures in
\textit{reasoning after information acquisition}.
Without this conditioning, an incorrect or missed ASK prevents the model
from obtaining the required information and therefore necessarily affects
the subsequent ANSWER.
Counting such a turn again as an ANSWER failure would conflate the two
decision stages and make it difficult to identify the actual source of
failure. Accordingly, $S_{\mathrm{ask}}$ measures how reliably the model acquires
the required information, while $S_{\mathrm{ans}}$ measures whether the
model can correctly use that information after successful acquisition.
When $N_{\mathrm{ask}}=0$, we define $S_{\mathrm{ans}}=0$ and consequently
$\mathrm{Ask\text{-}F1}=0$.
Importantly, $S_{\mathrm{ans}}$ is interpreted jointly with
$S_{\mathrm{ask}}$: even if the conditional answer score is high,
a small $S_{\mathrm{ask}}$ directly limits the final harmonic score.

\paragraph{Temporal weighting.}

Besides content correctness, proactive interaction is inherently
time-sensitive.
For an action with valid temporal window $W=[s,e)$, our default metric
uses the linear temporal weighting function

\begin{equation}
T_{\mathrm{lin}}(t;[s,e))
=
\mathbb{I}[s \leq t < e]
\left(
1-\frac{t-s}{e-s}
\right).
\label{eq:linear_temporal_metric}
\end{equation}

The valid window first determines whether an action is temporally
acceptable, while the decay term further distinguishes different
latencies within that window.
Once the beginning of a valid window is reached, sufficient information
has become available for the corresponding action.
Therefore, unnecessarily delaying an otherwise correct action provides
no additional utility in a streaming interaction. We choose linear decay because it is bounded, parameter-free, and easy
to interpret.
An action at the beginning of the valid window receives maximum temporal
credit, while the reward decreases smoothly toward the end of the window.
However, linear decay is not the only reasonable temporal utility
function.
We therefore examine whether the conclusions of AskStream remain stable
under alternative temporal weighting schemes.

\paragraph{Alternative temporal weighting functions.}

We examine whether Ask-F1 is sensitive to the particular temporal
weighting function used within a valid action window.
Let
\(
u=(t-s)/(e-s)\in[0,1)
\)
denote the normalized position of an action within the window.
In addition to the default linear weighting, we consider two nonlinear
variants. The first is a logarithmic weighting,

\begin{equation}
T_{\mathrm{log}}(t;[s,e))
=
\mathbb{I}[s \leq t < e]
\frac{\log(2-u)}{\log 2},
\label{eq:log_temporal_metric}
\end{equation}

which decays more slowly than the linear function and therefore assigns
relatively more credit to later but still valid actions.
The second is an exponential weighting,

\begin{equation}
T_{\mathrm{exp}}(t;[s,e))
=
\mathbb{I}[s \leq t < e]
\exp(-2u),
\label{eq:exp_temporal_metric}
\end{equation}

which places a stronger nonlinear preference on earlier actions.
We use a fixed decay coefficient of $2$ as a simple stress-test setting
rather than tuning it to individual models. In particular, the reward
decreases to $e^{-2}\approx0.135$ near the end of the valid window,
providing a substantially stronger timing penalty while still retaining
nonzero credit for late but valid actions.
Table~\ref{tab:metric_temporal_robustness} reports the results
for the seven API-evaluated models in the multiple-choice setting.
Although the absolute scores vary with the weighting function, the model
ordering remains unchanged across all three variants, indicating that
the comparative results are not sensitive to the precise form of the
within-window temporal decay.

\begin{table}[H]
\centering
\caption{
\textbf{Sensitivity of Ask-F1 to alternative temporal weighting functions.}
Results are reported for the seven models under the
multiple-choice setting. Linear is the default metric, while logarithmic
and exponential replace its within-window temporal weighting function.
Spearman $\rho$ w.r.t. Linear measures the rank correlation of model rankings relative to the default Linear metric, where $\rho=1.00$ indicates identical rankings.
}
\label{tab:metric_temporal_robustness}

\begingroup
\fontsize{8}{9.6}\selectfont
\setlength{\tabcolsep}{5.0pt}
\renewcommand{\arraystretch}{1.22}

\begin{tabular}{l|ccc}
\toprule
\textbf{Method}
&
\textbf{Logarithmic}
&
\textbf{Linear}
&
\textbf{Exponential}
\\
\midrule

\multicolumn{4}{l}{\textit{Offline closed-source}} \\

GPT-5.6-Sol
& 22.04
& 20.50
& 16.65 \\

Claude Sonnet 5
& 24.16
& 22.48
& 18.27 \\

\textbf{Gemini-3.8 Flash}
& \textbf{56.11}
& \textbf{52.18}
& \textbf{42.34} \\

\midrule
\multicolumn{4}{l}{\textit{Offline open-source}} \\

MiMo-V2.5
& 15.81
& 14.66
& 11.82 \\

Kimi K2.6
& 17.70
& 16.52
& 13.51 \\

Qwen3.8-Flash
& 18.78
& 17.60
& 14.52 \\

DeepSeek-V4.1-Flash
& 21.16
& 19.69
& 16.01 \\

\midrule
\textbf{Spearman $\rho$ w.r.t. Linear}
& \textbf{1.00}
& \textbf{1.00}
& \textbf{1.00} \\

\bottomrule
\end{tabular}

\endgroup
\end{table}

\paragraph{Multiple-ASK protocol.}

The main evaluation allows at most one ASK action per turn.
To study whether additional opportunities for information seeking improve
performance, we further consider a multiple-ASK protocol with an ASK
budget
\(B \in \{1,2,4,8,16\}\).
Under a budget of $B$, the model may issue up to $B$ ASK actions before
producing its final ANSWER.
Each ASK is evaluated at the time it is issued using the same temporal
and content criteria as in the main setting.
A valid ASK triggers the corresponding user or visual feedback, after
which the model continues the interaction with the newly acquired
information.
An invalid or redundant ASK consumes one unit of the ASK budget but does
not immediately terminate the turn, allowing the model to recover through
subsequent interactions.
Once the ASK budget is exhausted, the model may continue to WAIT or
produce its final ANSWER, while the ANSWER evaluation remains unchanged.
This protocol isolates the effect of repeated information-seeking
opportunities without modifying the underlying task or answer criterion.

\subsection{Uncertainty Analysis}
\label{app:uncertainty_analysis}
\paragraph{Experimental configuration.}
We evaluate Qwen3.5 9B on 130 five choice events from 78 videos to examine whether obtaining the information requested by an \textit{ask} action changes the model's belief over possible answers. At each checkpoint, the model receives ten chronological frames sampled at 1~Hz from the causal history ending at that checkpoint, with six checkpoints before asking and six after the reveal. The correct evidence is revealed one second after asking. The \textbf{Mismatched Reveal} condition follows the same temporal alignment but reveals an irrelevant entity from another event in the same video, while \textbf{Blurred} masks the task relevant entity and \textbf{Full} uses the original clear video. Let $x$ denote the ten frame input and let $\{a_1,\ldots,a_5\}$ denote the five complete answer texts. For $a_j=(y_{j,1},\ldots,y_{j,L_j})$, we compute its mean token log probability as
\[
s_j(x)=\frac{1}{L_j}\sum_{t=1}^{L_j}
\log p_\theta(y_{j,t}\mid x,y_{j,<t}).
\]
We then normalize the five scores into an answer distribution
\[
q_j(x)=\frac{\exp(s_j)}
{\sum_{k=1}^{5}\exp(s_k)}.
\]
Model uncertainty is measured using normalized entropy
\[
H(q)=\frac{1}{\log 5}
\sum_{j=1}^{5}q_j\log\frac{1}{q_j}.
\]

\paragraph{Entropy across interaction turns.}
Figure~\ref{fig2}(a) shows the evolution of normalized answer entropy across three interaction turns. Before the requested evidence appears, the model remains relatively uncertain. Once the correct evidence becomes available, normalized entropy consistently decreases across all three turns. In contrast, the \textbf{Mismatched Reveal} condition remains close to its previous uncertainty level. This comparison shows that the reduction in uncertainty is associated with the relevance of the revealed evidence rather than with additional visual observations or elapsed time alone.

\paragraph{Answer probability evolution.}
Figure~\ref{fig2}(b) visualizes how the answer distribution changes as visual evidence accumulates over time. Increasing support for the gold entity raises the probability assigned to the gold answer while reducing the probabilities assigned to the four incorrect alternatives. When the revealed evidence is irrelevant to the current question, the same separation does not emerge. The probability evolution therefore shows that relevant evidence specifically shifts the model's belief toward the correct answer.

\paragraph{Final answer quality.}
Figure~\ref{fig2}(c) summarizes the final answer performance. With blurred input, the model achieves $42.3\%$ accuracy and assigns an average probability of $25.8\%$ to the gold answer. After asking for and receiving the key evidence, accuracy increases to $62.3\%$ and the mean gold answer probability rises to $36.8\%$. These results approach the fully visible condition, which achieves $63.1\%$ accuracy and a mean gold answer probability of $36.6\%$. Together, the three panels show that obtaining the requested evidence reduces model uncertainty and improves answer quality by shifting belief toward the correct answer.

\subsection{Open-Ended Evaluation Results}
\label{app:open_ended_results}
Table~\ref{tab:askstream_main_results_open} reports the results under the open-ended setting. Overall, performance drops substantially compared with the multiple-choice setting, indicating that generating appropriate ASK and ANSWER responses is considerably more challenging than selecting from predefined candidates. Gemini-3.8 Flash achieves the strongest overall performance with an Ask-F1 of \textbf{29.19\%}, while its $S_{\mathrm{ask}}$ (23.34\%) remains notably lower than $S_{\mathrm{ans}}$ (38.97\%), again suggesting that proactive information acquisition is a major bottleneck. Open-source offline models generally achieve lower scores, and streaming proactive models struggle even more severely: most obtain zero on nearly all subtasks, with AURA reaching only 4.01\% Ask-F1. These results further demonstrate that current models have limited ability to reliably formulate and time proactive information-seeking actions in unconstrained open-ended interactions.





\subsection{Validation of the Open-Ended Judge}
\label{app:judge_validation}

To validate the reliability of Gemini-3.8 Flash as the judge in the open-ended evaluation, we randomly sample 300 turns and ask two human annotators to independently assess the correctness of both ASK and ANSWER responses, with disagreements resolved by an additional annotator. We compare inter-annotator agreement and the agreement between the Gemini judge and the human consensus, and separately report results on \textbf{Gemini generated} and \textbf{non Gemini generated} responses to examine potential self-preference. As shown in Table~\ref{tab:judge_validation}, the Gemini judge exhibits high agreement with human judgments for both ASK and ANSWER. Moreover, its agreement on Gemini generated responses is comparable to that on non Gemini responses, providing no clear evidence of self-preference within the sampled data.

\begin{table}[H]
\centering
\caption{
Validation of Gemini-3.8 Flash as the open-ended judge on 300 sampled turns.
\textbf{Human\&Human} reports the agreement between two independent human annotators and serves as a reference for annotation consistency.
\textbf{ASK Agreement} and \textbf{ANSWER Agreement} denote the percentage of identical correctness judgments for ASK and ANSWER responses, respectively.
\textbf{Overall Agreement} is computed over both ASK and ANSWER judgments.
For the Gemini Judge vs.\ Human Consensus rows, human consensus is used as the reference judgment. Results are further separated into Gemini-3.8 Flash and non Gemini outputs to examine potential self-preference.
}
\label{tab:judge_validation}

\small
\setlength{\tabcolsep}{5pt}
\renewcommand{\arraystretch}{1.08}

\begin{tabularx}{\linewidth}{
@{}
>{\raggedright\arraybackslash}X
c c c c
@{}
}
\toprule
\textbf{Comparison}
& \textbf{ASK}
& \textbf{ANSWER}
& \textbf{Overall} \\
& \textbf{Agreement (\%)}
& \textbf{Agreement (\%)}
& \textbf{Agreement (\%)}
& \\
\midrule

Human\&Human
& 95.33
& 97.00
& 96.17
 \\

\midrule
\multicolumn{5}{@{}l}{\textit{Gemini Judge vs.\ Human Consensus}} \\

All outputs
& \textbf{92.67}
& \textbf{94.33}
& \textbf{93.50} \\

Gemini-3.8 Flash outputs
& 93.06
& 94.44
& 93.75 \\

Non Gemini outputs
& 92.31
& 94.23
& 93.27 \\

\bottomrule
\end{tabularx}

\end{table}

\subsection{StreamHarness Ablation}
\label{sec:streamharness_ablation}

We compare StreamHarness with three simpler context strategies using Gemini-3.8 Flash under the multiple-choice setting. The sliding-window baseline keeps only the most recent 10 frames and removes the mid- and long-term memory, while the other two baselines uniformly sample 64 or 128 frames from the observed video. For uniform sampling, we ensure that each valid temporal window in the observed video contains at least one sampled frame, while the temporal window boundaries are not provided to the model. As shown in Table~\ref{tab:streamharness_ablation}, removing the hierarchical memory reduces Ask-F1 from 52.18\% to 41.70\%. Uniform sampling leads to a larger performance drop, reaching 26.72\% with 128 frames and 16.99\% with 64 frames. These results show that preserving dense recent observations together with compressed historical context is important for proactive ASK and ANSWER decisions in streaming video.

\begin{table}[H]
\centering
\small
\caption{Ablation of StreamHarness using Gemini-3.8 Flash under the multiple-choice setting. Uniform sampling ensures that each valid temporal window contains at least one sampled frame.}
\label{tab:streamharness_ablation}
\begin{tabular}{lccc}
\toprule
Context Strategy
& $S_{\mathrm{ask}}$
& $S_{\mathrm{ans}}$
& Ask-F1 \\
\midrule
StreamHarness
& \textbf{44.63}
& \textbf{62.80}
& \textbf{52.18} \\

Sliding Window
& 35.92
& 49.71
& 41.70 \\

Uniform Sampling (128)
& 22.41
& 33.08
& 26.72 \\

Uniform Sampling (64)
& 13.86
& 21.94
& 16.99 \\
\bottomrule
\end{tabular}
\end{table}

\begin{table}[t]
\centering
\caption{Main results on AskStream in the Open-ended setting.}
\label{tab:askstream_main_results_open}

\begingroup

\fontsize{8}{9.6}\selectfont
\setlength{\tabcolsep}{2.4pt}
\renewcommand{\arraystretch}{1.22}

\setlength{\arrayrulewidth}{0.35pt}
\setlength{\aboverulesep}{0pt}
\setlength{\belowrulesep}{0pt}
\setlength{\doublerulesep}{1.0pt}

\definecolor{askheader}{RGB}{224,224,224}
\definecolor{askrow}{RGB}{240,240,240}
\definecolor{askgroup}{RGB}{125,125,125}
\definecolor{askbest}{RGB}{217,245,254}
\definecolor{askbestline}{RGB}{124,194,204}
\definecolor{askgainred}{RGB}{190,75,75}

\arrayrulecolor{black!75}

\newcommand{\askscoregain}[2]{%
  \mbox{%
    \textbf{#1}\hspace{0.15em}%
    \raisebox{0.45ex}{%
      \textcolor{askgainred}{%
        \scalebox{0.70}{$\uparrow$#2}%
      }%
    }%
  }%
}

\newcommand{\asksep}{%
  \arrayrulecolor{black!22}%
  \cdashline{1-13}[0.35pt/1.25pt]%
  \arrayrulecolor{black!75}%
}

\resizebox{\linewidth}{!}{%
\begin{tabular}{l||c|*{4}{c}|*{4}{c}|ccc}

\toprule[0.9pt]

\rowcolor{askheader}
& &
\multicolumn{4}{c|}{\textbf{Video}} &
\multicolumn{4}{c|}{\textbf{User}} &
\multicolumn{3}{c}{} \\

\rowcolor{askheader}
& &
\multicolumn{2}{c}{Explicit} &
\multicolumn{2}{c|}{Implicit} &
\multicolumn{2}{c}{Intent} &
\multicolumn{2}{c|}{Preference} &
\multicolumn{3}{c}{%
  \multirow{-2}{*}{\textbf{Metric}}%
} \\

\rowcolor{askheader}
\multicolumn{1}{c||}{%
  \multirow{-3}{*}{\textbf{Method}}%
} &
\multirow{-3}{*}{\textbf{Backbone}} &
KVEM & VIM & ICM & PACM &
AUQ & WUQ & UPM & UPC &
$S_{\mathrm{ask}}$ &
$S_{\mathrm{ans}}$ &
\textbf{Ask-F1} \\

\hline\hline

\multicolumn{13}{l}{%
  \textcolor{askgroup}{%
    \textit{Offline closed-source models}%
  }%
} \\

Claude Sonnet 5 & -
& 13.59
& 17.62
& 0
& \textbf{4.78}
& 12.69
& 12.28
& 0
& \textbf{19.73}
& 15.88
& 10.99
& 12.99 \\
\asksep

\rowcolor{askrow}
GPT-5.6-Sol & -
& 6.30 & 3.90 & 2.80 & 0
& 16.20 & 9.96 & 7.63 & 16.52
& 12.05 & 18.80 & 14.69 \\

\arrayrulecolor{askbestline}
\hline
\rowcolor{askbest}
\textbf{Gemini 3.8 Flash} & -
& \textbf{27.61}
& \textbf{29.40}
& \textbf{35.79}
& 0
& 35.87
& \textbf{32.12}
& 14.41
& 18.43
& \textbf{23.34}
& \textbf{38.97}
& \textbf{29.19} \\
\hline
\arrayrulecolor{black!75}

\multicolumn{13}{l}{%
  \textcolor{askgroup}{%
    \textit{Offline open-source models}%
  }%
} \\

MiMo-V2.5 & -
& 10.15 & 9.29 & 2.57 & 0
& 7.86 & 8.90 & 6.63 & 0
& 4.73 & 25.72 & 7.99 \\
\asksep

\rowcolor{askrow}
Kimi K2.6 & -
& 15.54
& 0
& 7.82
& 0
& \textbf{41.13}
& 15.93
& \textbf{19.31}
& 0
& 6.12
& 17.77
& 9.10 \\
\asksep

DeepSeek-V4.1-Flash & -
& 6.28 & 5.74 & 0 & 0
& 26.54 & 7.50 & 10.16 & 11.49
& 10.09 & 17.94 & 12.92 \\

\arrayrulecolor{askbestline}
\hline
\rowcolor{askbest}
\textbf{Qwen3.8-flash} & -
& 6.03 & 10.13 & 1.29 & 0
& 7.30 & 5.52 & 0 & 0
& 12.54 & 16.03 & 14.07 \\
\hline
\arrayrulecolor{black!75}

\multicolumn{13}{l}{%
  \textcolor{askgroup}{%
    \textit{Streaming proactive models}%
  }%
} \\

MOSS-VL-Realtime & Qwen3-8B
& 0 & 0 & 0 & 0
& 0 & 0 & 0 & 0
& 0 & 0 & 0 \\
\asksep

\rowcolor{askrow}
MiniCPM-o 4.5 & Qwen3-8B
& 0 & 0 & 0 & 0
& 0 & 0 & 0 & 0
& 0 & 0 & 0 \\
\asksep

MMDuet2 & Qwen2.5-VL-3B
& 0 & 0 & 0 & 0
& 0 & 0 & 0 & 0
& 0 & 0 & 0 \\
\asksep

\rowcolor{askrow}
TimeChat-Online & Qwen2.5-VL-7B
& 0 & 0 & 0 & 0
& 0 & 0 & 0 & 0
& 0 & 0 & 0 \\
\asksep

VideoLLM-online & Llama-3-8B
& 0 & 0 & 0 & 0
& 0 & 0 & 0 & 0
& 0 & 0 & 0 \\
\asksep

\rowcolor{askrow}
JoyAI-VL-Interaction & JoyAI-VL-1.0
& 0 & 0 & 2.89 & 0
& 0 & 0 & 0 & 0
& 0.43 & 7.75 & 0.81 \\

\arrayrulecolor{askbestline}
\hline
\rowcolor{askbest}
\textbf{AURA} & Qwen3-VL-8B
& 4.19 & 5.76 & 0 & 0
& 6.50 & 0 & 0 & 0
& 2.77 & 7.25 & 4.01 \\
\hline
\arrayrulecolor{black!75}

\bottomrule[0.9pt]
\end{tabular}%
}

\endgroup
\end{table}

\subsection{Data Construction}
\label{Data_Construction}
\subsubsection{Annotation and Verification Interface.}
\begin{figure}[t]
    \centering
    \includegraphics[width=\linewidth]{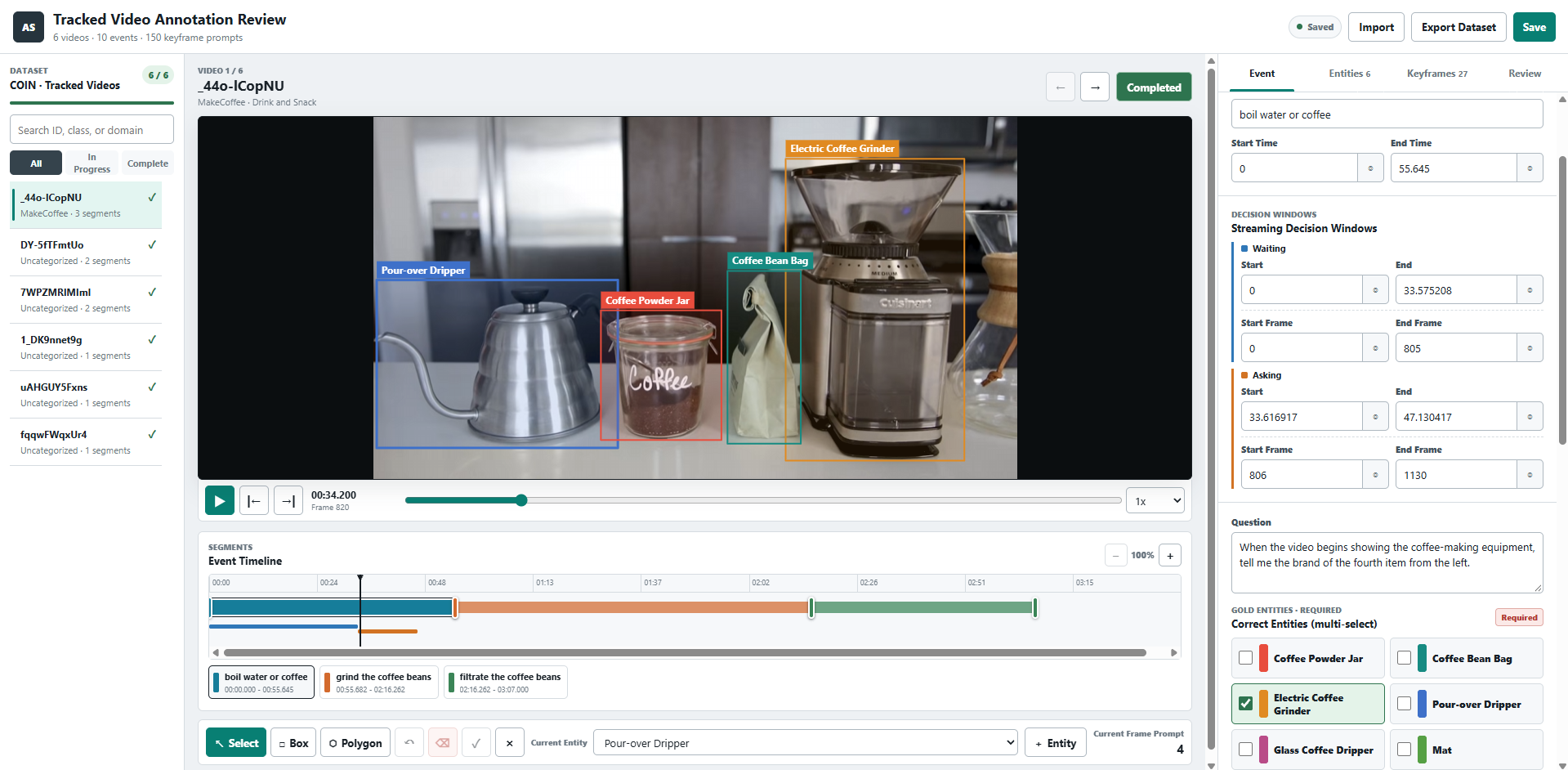}
    \caption{
    \textbf{Web interface for KVEM annotation.}
    Annotators review each event together with its tracked entities and keyframes, refine entity bounding boxes when necessary, select the gold entity, and annotate the temporal decision windows for streaming interaction. The interface also supports event-level query annotation and visualization of the corresponding timeline, facilitating consistent construction of KVEM instances.
    }
    \label{web_page_1}
\end{figure}

\begin{figure}[t]
    \centering
    \includegraphics[width=\linewidth]{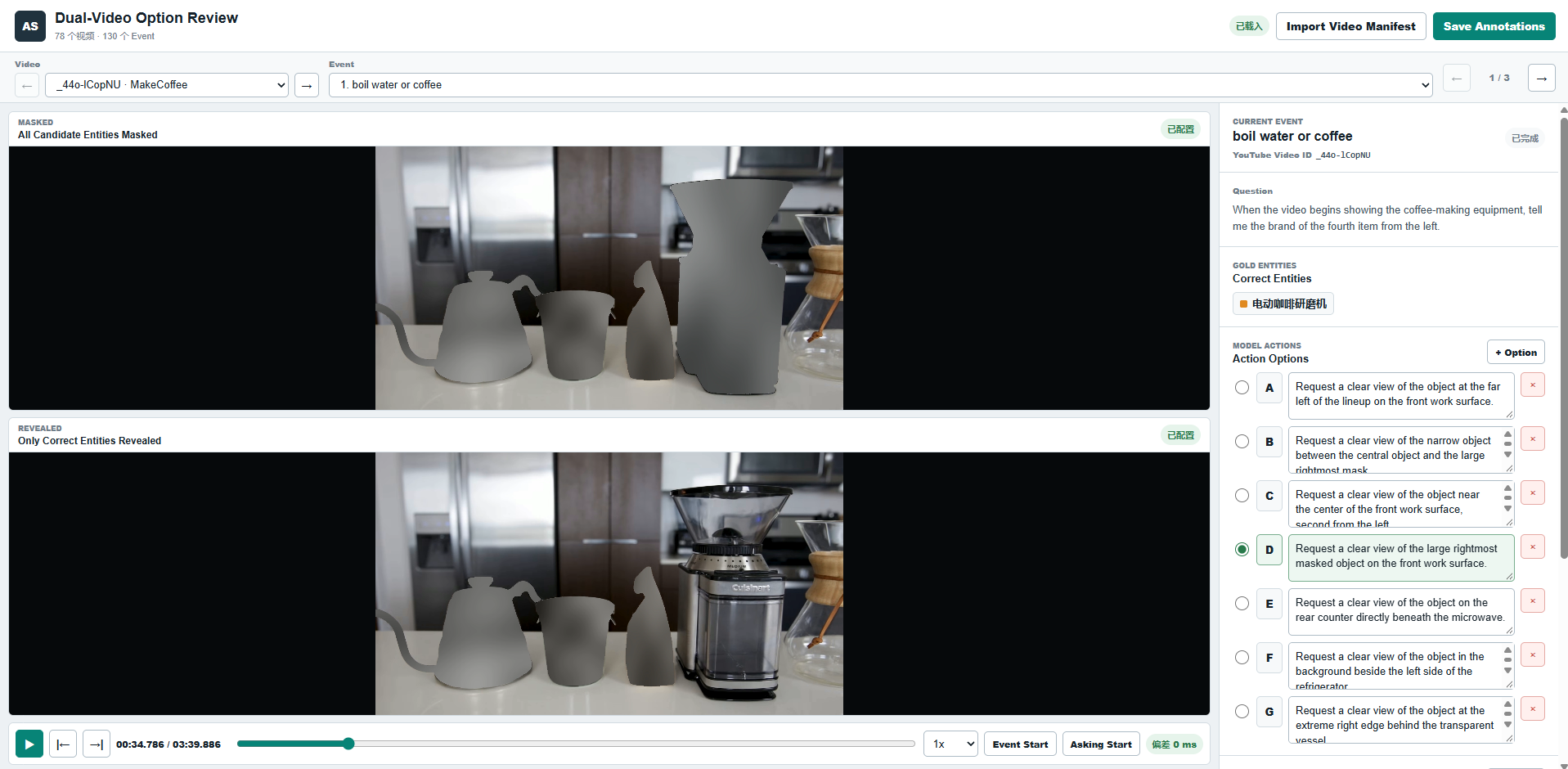}
    \caption{
    \textbf{Web interface for KVEM human verification.}
    The fully masked video and the revealed video, in which only the gold entity is visible, are presented side by side for comparison. Annotators jointly inspect the query, gold entity, and ASK candidates to verify that the masked video contains no information leakage, that revealing the requested gold entity is sufficient to uniquely answer the query, and that the correct and incorrect ASK options correspond to appropriate and inappropriate information requests, respectively.
    }
    \label{web_page_2}
\end{figure}

To facilitate efficient and consistent data construction, we develop task-specific web interfaces for all eight subtasks, covering both annotation and human verification. We use KVEM as a representative example. As shown in Figure~\ref{web_page_1}, the annotation interface allows annotators to review event-level video segments, inspect and refine tracked entities at keyframes, select the gold entity, and annotate the temporal decision windows associated with WAIT, ASK, and ANSWER actions. The interface also supports the annotation of proactive queries and other event-level metadata required for constructing each instance. After data construction, we conduct an additional human verification stage using the interface shown in Figure~\ref{web_page_2}. For each instance, annotators jointly inspect the fully masked video and the corresponding revealed video in which only the gold entity is visible, together with the query and candidate ASK options. This dual-video design enables annotators to verify that the masked version does not leak sufficient information to uniquely answer the question, while the revealed version provides the necessary evidence to identify the correct answer. The verification interface further allows annotators to check whether the correct ASK requests the intended missing visual evidence and whether the incorrect ASK candidates correspond to irrelevant or insufficient evidence.

\subsubsection{Task-specific Construction Details}
\textbf{KVEM.} For the KVEM task, we first select videos from the COIN~\citep{tang2019coinlargescaledatasetcomprehensive} dataset that contain diverse entities and whose key visual evidence can be reliably masked. We determine the number of interaction turns according to the step segmentation provided by the source dataset, and identify multiple key entities in each turn, which are subsequently used to construct correct and incorrect ASK candidates. To ensure reliable tracking, we manually annotate bounding boxes for the key entities in each turn and re-annotate the corresponding entities whenever a scene transition occurs, thereby reducing tracking drift or failure. Based on these annotations, we use Grounded SAM 2 to segment and track the entities across frames, and apply Gaussian blur to mask them. For each instance, we generate two video variants: one in which all key entities are masked, and another in which only the gold entity remains unmasked. The latter is returned as the visual evidence provided after a correct ASK. We then manually annotate the temporal interval during which the gold entity appears and disappears, and use these annotations to define the corresponding temporal windows. Based on the annotated windows, we manually construct proactive queries, typically using formulations such as ``when ...tell me ...'', as well as the correct and incorrect ANSWER candidates. Finally, each instance undergoes manual verification using both video variants: the fully masked version is inspected to ensure that no information leakage allows the question to be uniquely answered without the requested evidence, while the version with only the gold entity visible is used to verify that the revealed evidence is sufficient to uniquely determine the correct answer.

\begin{figure}[t]
    \centering
    \includegraphics[width=\linewidth]{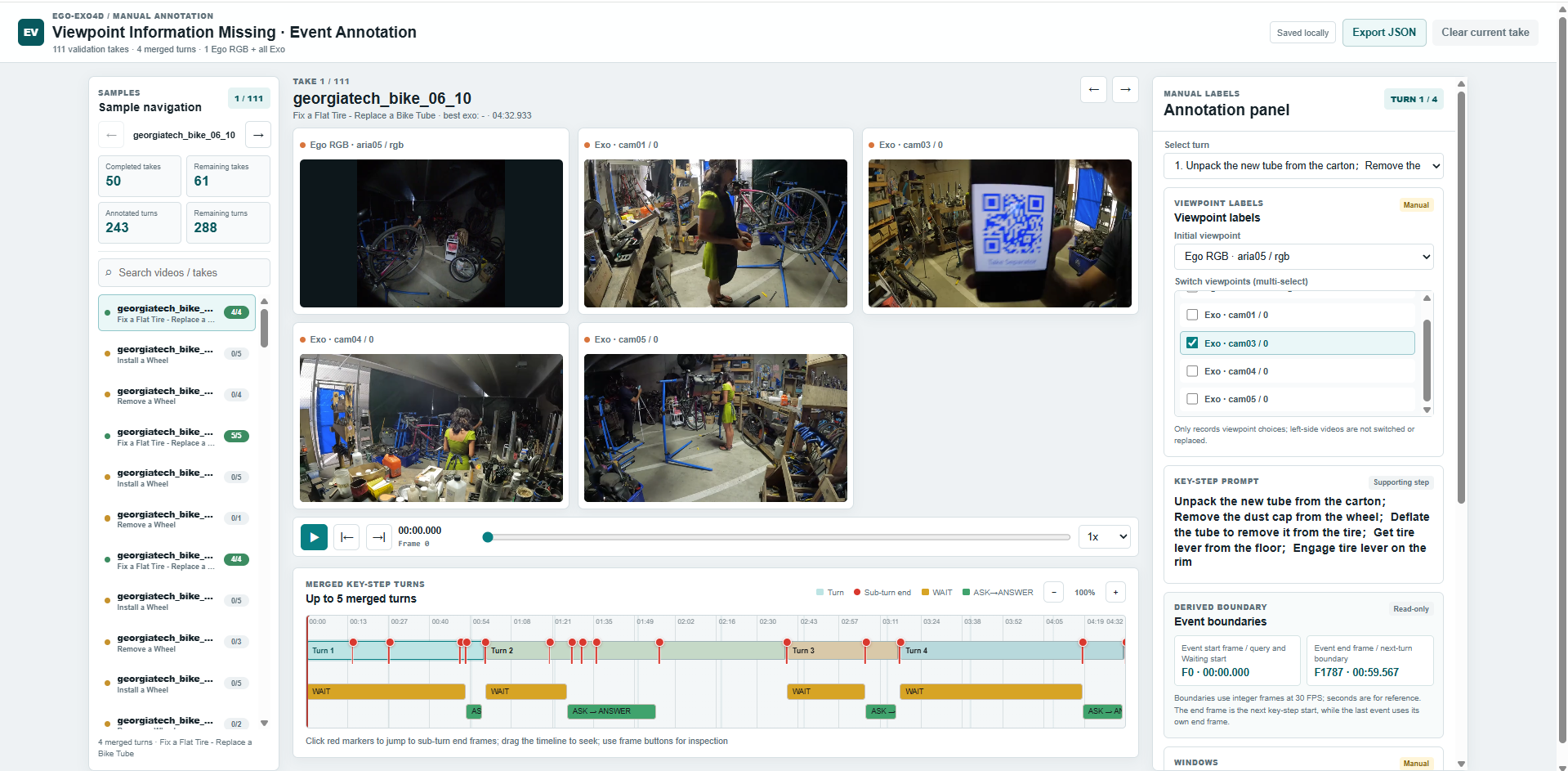}
    \caption{
    \textbf{Multi-view annotation interface for the VIM task.}
    The interface presents temporally synchronized egocentric and multiple exocentric video streams from Ego-Exo4D, together with the key-step-based turn structure and temporal decision windows. Annotators jointly inspect all viewpoints, designate the initial viewpoint, identify alternative viewpoints that reveal the missing visual evidence, and construct viewpoint-dependent queries and ASK candidates.
    }
    \label{fig:vim_annotation_interface}
\end{figure}

\textbf{VIM.} For the VIM task, we build upon the Ego-Exo4D~\citep{grauman2024egoexo4dunderstandingskilledhuman} dataset, which provides temporally synchronized egocentric and multiple exocentric video streams of the same activity. We reuse the key-step annotations provided by the source dataset to define the start and end boundaries of interaction turns. For each turn, annotators jointly inspect all available viewpoints using the multi-view annotation interface shown in Figure~\ref{fig:vim_annotation_interface}, and construct a query that cannot be answered from the designated initial viewpoint but can be uniquely answered from one or more alternative viewpoints. We then manually annotate the corresponding temporal decision windows according to the onset and disappearance of the query-relevant visual evidence. The ASK candidates are formulated as precise textual descriptions of different available viewpoints, such that a correct ASK requests a viewpoint that exposes the missing evidence required to answer the query, whereas incorrect ASK candidates request viewpoints that remain insufficient for resolving the information gap. After a correct ASK, the video stream is switched to the requested viewpoint, allowing the model to observe the previously missing evidence before producing its answer. Finally, each instance undergoes manual verification. Annotators first inspect the designated initial viewpoint together with the query to confirm that the answer cannot be uniquely inferred and that no relevant information is leaked. They then inspect the candidate alternative viewpoints to verify that the viewpoint associated with the correct ASK reveals sufficient evidence to uniquely determine the answer, while the viewpoints associated with incorrect ASK candidates remain insufficient or irrelevant. This procedure ensures that successful answering requires an appropriate viewpoint-switching request rather than relying on information already available from the initial stream.

\textbf{ICM.} For the ICM task, we build upon PEARL~\citep{zheng2026pearlpersonalizedstreamingvideo}, a streaming-video dataset with personalized concepts. We remove the original personalized-concept annotations while preserving the multi-turn interaction structure provided by the dataset and manually annotate the queries, answers, and temporal decision windows for each turn. The ASK candidates are designed to evaluate whether the model can correctly identify and request clarification for all personalized entity concepts referenced in an ambiguous query. A correct ASK must accurately describe and localize all relevant entities whose personalized concepts are required to disambiguate the user's intent, whereas incorrect ASK candidates either contain incorrect entity descriptions or provide only partial clarification that is insufficient to resolve the ambiguity. After a correct ASK, we return images of the corresponding entities as additional evidence, allowing the model to ground the previously missing personalized concepts before answering. Finally, each instance undergoes manual verification. Annotators first inspect the original interaction without the additional entity evidence to confirm that the query cannot be uniquely resolved from the available video and dialogue context alone. They then verify that the correct ASK covers all personalized entity concepts necessary for disambiguation, while each incorrect ASK is either factually inconsistent with the referenced entities or incomplete and therefore insufficient to uniquely determine the user's intended referents. After the corresponding entity images are revealed, annotators further confirm that the ambiguity is resolved and that the query admits a unique and unambiguous correct answer.

\textbf{PACM.} For the PACM task, since personalized actions in the real world can be highly diverse and may overlap with action concepts already encoded in a model's prior knowledge, we use Blender to synthesize personalized action videos with newly defined action concepts. Specifically, we adopt existing human character models and manually design 25 atomic actions, which are then compositionally combined into diverse personalized actions and assigned novel names to minimize the influence of model priors. We synthesize two types of videos: (1) multiple characters repeatedly perform different personalized actions in parallel, and (2) multiple characters perform personalized actions sequentially. For each target personalized action, we additionally construct several hard negative actions by modifying one of its constituent atomic actions while keeping the remaining components unchanged. For the first video type, the query follows the form ``after observing the first complete execution of all characters, tell me XXX,'' and the valid ANSWER window starts once all characters have completed their first full execution and extends to the end of the turn. For the second video type, the query follows the form ``tell me when you can determine which character is performing the correct personalized action.'' In this case, the ANSWER window starts when the character performing the target action has completed the action and extends to the end of the turn. ASK candidates are constructed according to the discriminative characteristics of the personalized actions: a correct ASK requests an action characteristic that can uniquely distinguish the target from the other candidates, whereas incorrect ASK candidates request characteristics that are shared by at least two characters and therefore cannot uniquely identify the target. To provide the missing personalized action concept after a correct ASK, we additionally render the corresponding action using a textureless white humanoid character and return this action video as the requested visual demonstration, thereby avoiding additional appearance-related cues. Finally, the correct and incorrect ANSWER candidates indicate which character performs the target personalized action. Finally, each instance undergoes manual verification to ensure both the validity of the ASK candidates and the answerability of the task. Annotators first verify that the correct ASK requests an action characteristic that uniquely distinguishes the target personalized action from all other candidates, while each incorrect ASK remains ambiguous and is compatible with at least two candidate actions. They then inspect the returned white-humanoid demonstration to confirm that it provides sufficient information to identify the target action without introducing additional appearance or identity cues. Finally, annotators verify that, after the requested action concept is revealed, the correct character can be uniquely identified and the corresponding ANSWER is unambiguous.

\textbf{AUQ.} For the AUQ task, we construct the data using videos from StreamArena~\citep{zhang2026streamarena}, supplemented with additional videos collected from YouTube. We first use PySceneDetect to assist manual segmentation of each video into interaction turns, with all detected boundaries subsequently reviewed and adjusted by annotators. For each turn, we manually construct a video-grounded ambiguous query whose intended referent or interpretation cannot be uniquely determined from the query alone. We annotate the start of the valid ASK window as the earliest time at which the ambiguity becomes identifiable from the observed video context. The ANSWER window begins when the visual event or entity required to answer the clarified query first becomes available and ends when the corresponding evidence disappears from the stream. We construct ASK candidates to cover different possible interpretations of the ambiguity, where the correct ASK explicitly requests the clarification necessary to resolve the uncertainty, while incorrect ASK candidates either target an irrelevant source of ambiguity or fail to sufficiently disambiguate the query. After a correct ASK, the model receives the corresponding user clarification, which resolves the ambiguity and enables subsequent answering. Finally, each instance undergoes manual verification in two stages. Annotators are first presented with the original ambiguous query and the observed video context to confirm that the intended answer cannot be uniquely determined without clarification. They are then provided with the clarified query and verify that the ambiguity is fully resolved and that the available visual evidence supports a unique and unambiguous correct answer within the annotated ANSWER window.

\textbf{WUQ.} For the WUQ task, we construct the data using videos from StreamingBench~\citep{lin2026streamingbench} and additional videos collected from YouTube. Similar to AUQ, we first use PySceneDetect to assist the manual segmentation of each video into interaction turns, with all automatically detected boundaries subsequently reviewed and adjusted by annotators. For each turn, we first annotate a standard proactive query associated with a visual trigger and determine the corresponding ANSWER window based on the temporal extent during which the required visual evidence is available. We then manually select an appropriate entity whose appearance makes it possible to detect an inconsistency in the user's query and annotate the onset time of this entity as the beginning of the valid ASK window. Based on the selected entity, we construct a corrupted version of the query by injecting an incorrect factual statement about the entity, such that the query contains a false premise that becomes identifiable once the relevant visual evidence appears. The model is expected to recognize and explicitly correct the erroneous premise rather than directly answering the corrupted query. After the model correctly identifies the inconsistency and provides the appropriate correction, the user returns a confirmation acknowledging the correction, after which the interaction proceeds using the aligned information. Finally, each instance undergoes manual verification. Annotators first inspect the corrupted query together with the video to confirm that the injected statement is indeed inconsistent with the observed evidence and that the inconsistency becomes detectable no earlier than the annotated ASK window. They then verify that the proposed correction accurately resolves the false premise and that, after the correction is confirmed, the original information needed can be uniquely and unambiguously answered within the annotated ANSWER window.

\textbf{UPM and UPC.} For the UPM and UPC tasks, we collect preference-oriented videos from YouTube, covering domains such as clothing, furniture, and other everyday scenarios in which human preferences naturally influence the desired answer. We manually segment each video into interaction turns according to scene transitions. For each turn, annotators construct a proactive query whose answer depends on the user's preference, and identify multiple candidate entities appearing in the relevant frames These entities are used to construct the ANSWER candidates. We then randomly select one candidate as the ground-truth preferred entity and manually annotate the corresponding ASK and ANSWER windows according to the temporal availability of the query-relevant evidence.
For \textbf{UPM}, the constructed instance is used to generate a set of user-preference descriptions along different preference dimensions. The correct preference is designed to uniquely identify the ground-truth entity among the available candidates, whereas the incorrect preference options are intentionally insufficient to uniquely determine that entity. After a correct ASK, the user provides the missing preference information, allowing the model to combine the revealed preference with the observed visual candidates and produce the final answer. During human verification, annotators first confirm that the query cannot be uniquely answered without additional preference information. They then verify that the preference associated with the correct ASK uniquely selects the ground-truth entity, while each incorrect preference remains compatible with multiple candidates and therefore cannot uniquely determine the answer.
For \textbf{UPC}, we additionally construct scenarios in which a previously known user preference has become outdated. To make preference changes objective and interpretable rather than arbitrary, we select contexts in which preferences may plausibly change due to an observable change in user circumstances, such as moving to a new home, losing weight, or other changes that alter the suitability of previously preferred items. We therefore provide both a historical preference and an updated preference context, while constructing ASK candidates in the same manner as in UPM: the correct ASK requests the preference dimension necessary to identify the user's current preference, whereas incorrect ASK candidates fail to resolve the uncertainty. After a correct ASK, the user provides the updated preference, and the model is expected to answer according to the current rather than historical preference. Human verification follows the same uniqueness checks as in UPM, with an additional consistency check that annotators can recognize from the provided context that the historical preference is no longer sufficient or applicable, and that obtaining the updated preference is necessary to uniquely determine the correct answer.

\subsubsection{Temporal Window Annotation Consistency.}
Because Ask-F1 directly relies on the annotated temporal decision windows, we further evaluate the consistency of our temporal annotations. We use stratified sampling to select 160 instances from the benchmark, with 20 instances from each of the eight subtasks. Two annotators independently re-annotate the temporal windows for each sampled instance without access to the original annotations or to each other's annotations. The annotators follow the same annotation protocol used during benchmark construction and determine the earliest valid temporal boundaries at which an ASK or ANSWER action becomes appropriate.

We quantify temporal agreement using temporal intersection-over-union (tIoU) between the two independently annotated windows, together with the absolute differences between their start and end boundaries. For ASK windows, the two annotators achieve an average tIoU of \textbf{0.962}, with mean absolute differences of \textbf{0.21}~s and \textbf{0.27}~s for the start and end boundaries, respectively. For ANSWER windows, the average tIoU is \textbf{0.971}, with corresponding start and end boundary differences of \textbf{0.18}~s and \textbf{0.23}~s. These results show that independent annotators largely agree on the temporal regions in which proactive actions become valid, supporting the reliability of the temporal supervision used in AskStream and Ask-F1.

\clearpage
\subsection{Prompt}
\label{Prompt}

\subsubsection{System\_Prompt}

\begin{promptbox}{GENERAL\_SYSTEM\_PROMPT}
You are a visual assistant that can proactively ask follow-up questions while observing a continuously arriving video stream to answer the user's question. While the video stream continues to arrive, you may choose among WAIT, ASK, and ANSWER. They respectively mean continuing to wait, proactively asking about information that may currently be missing, and answering the user's question. Each question permits only one ASK action and one ANSWER action.

When a question is active, choose one action listed under [AVAILABLE ACTIONS]. For choice actions, such as <ASK:A> or <ANSWER:F>, use the displayed form without adding anything else. For open-ended actions, put the request or answer inside the displayed form.

Decision rules:
1. The current question or the video may or may not contain missing information. If you cannot yet determine whether information is missing, choose <WAIT> to continue receiving subsequent information. If the meaning of the question is already clear but the evidence needed to trigger an answer has not yet appeared, also choose <WAIT>. Note: not knowing the answer does not mean that you should choose <ASK>. Choose <ASK> only when you have clearly confirmed that information essential to answering is missing and cannot be obtained by continuing to observe the video. When the available information satisfies the trigger condition and you can answer the question, choose <ANSWER>.
2. After ASK, you will receive visual or textual information provided by the user.
3. Use [CURRENT EVENT ACTION HISTORY] to determine whether ASK has already happened. These benchmark questions require one correct ASK before ANSWER. Before ASK, use WAIT or ASK; after ASK, use WAIT or ANSWER as shown by the available action forms. The displayed forms describe the output syntax; submitting ANSWER before ASK is premature even when its form is displayed. A task rule about answering once a trigger appears applies after the required ASK.
4. Never infer the answer from your own prior knowledge. Use only the current question, the video stream, the current event's action history, and the evidence delivered after a correct ASK.
5. For choice-based output, use the displayed option action exactly. For open-ended output, begin an ASK action with <ASK: or an ANSWER action with <ANSWER:, write the actual request or answer, and finish with >. Never output a bare free-text request or answer without its ASK or ANSWER envelope; such output is invalid.

FINAL OUTPUT CONSTRAINT: Your entire final response must be exactly one valid action of a type listed under [AVAILABLE ACTIONS]. Output no explanation, reasoning, Markdown, punctuation, timestamp, or additional text before or after that action.
\end{promptbox}

\clearpage

\subsubsection{CHOICE\_SETTING\_PROMPT}

\begin{promptbox}{KVEM\_PROMPT\_CHOICE}
Task-specific rules: KVEM
1. You will receive a video in which some entities are occluded. When you confirm that the relevant entity is occluded, choose <ASK> only to request the missing visual information needed to identify that entity, naming its accurate location or visual referent. Do not merely request an entity already explicitly mentioned in the question, ask for a broad scene description, or ask for an unrelated attribute. Before confirming the occlusion, choose <WAIT>. After the correct ASK, answer when the question's trigger condition is satisfied.
\end{promptbox}

\begin{promptbox}{VIM\_PROMPT\_CHOICE}
Task-specific rules: VIM
1. Once the question's trigger condition has occurred and you confirm that the current viewpoint cannot answer the question, choose <ASK> to request a concrete camera angle, side, or view that you believe can answer it. Any specific viewpoint that would genuinely provide sufficient evidence is valid; vague or unrelated view requests are not. Before confirming the missing viewpoint, choose <WAIT>. After the correct ASK, answer when sufficient evidence is available.
\end{promptbox}

\begin{promptbox}{ICM\_PROMPT\_CHOICE}
Task-specific rules: ICM
1. When you confirm that the video content cannot answer the user's question and that the ASK option can uniquely and unambiguously identify the entity referred to by each private entity alias mentioned in the question, choose <ASK> to select a clear personalized definition for every private entity alias referred to in the question. Identify each alias explicitly and do not ask for a generic definition. If the ASK option cannot yet uniquely and unambiguously identify the private entity referred to by each alias, continue choosing <WAIT> until it can. After receiving the definitions, continue choosing <WAIT> until the trigger condition specified in the question appears, then choose <ANSWER>.
\end{promptbox}

\begin{promptbox}{PACM\_PROMPT\_CHOICE}
Task-specific rules: PACM
1. When the personalized action definitions needed to answer the question are missing, choose the ASK option identifying every required personalized action. The ASK selects action definitions, not a person performing the action. After receiving the definitions, continue choosing <WAIT> until the trigger condition appears, then choose <ANSWER>.
\end{promptbox}

\begin{promptbox}{AUQ\_PROMPT\_CHOICE}
Task-specific rules: AUQ
1. If, together with the video content, you confirm that the question is ambiguous, choose <ASK> to identify the specific ambiguous part and request the clarification needed to resolve the question to one unique referent rather than multiple possible referents. Do not make a generic request for more information. After the correct ASK, continue choosing <WAIT> until the trigger condition specified in the question appears, then choose <ANSWER>.
2. AUQ is an exact-reproduction task. Reproduce the specified reminder message exactly, including capitalization and punctuation, without paraphrasing or adding words.
\end{promptbox}

\begin{promptbox}{WUQ\_PROMPT\_CHOICE}
Task-specific rules: WUQ
1. If the question's trigger condition appears and you confirm that the question contains a visual error because the video content does not match it, choose <ASK> to identify the specific false or incorrect part, stating the visually supported correction or requesting confirmation of it. Do not make a generic request. Before that evidence appears, choose <WAIT>. After the correct ASK, answer when the corrected trigger condition is satisfied.
\end{promptbox}

\begin{promptbox}{UPM\_PROMPT\_CHOICE}
Task-specific rules: UPM
1. When you confirm that answering the question requires knowing the user's personal preference and that the answer to the ASK option can uniquely and unambiguously distinguish one visible entity, choose <ASK> to request the concrete preference dimension needed to answer it, such as a particular attribute, style, or choice criterion. Do not ask for preferences generically. After receiving the user's reply to a correct <ASK>, continue choosing <WAIT> until the trigger condition specified in the question appears, then choose <ANSWER>. Incorrect ASKs concern features shared by multiple entities.
\end{promptbox}

\begin{promptbox}{UPC\_PROMPT\_CHOICE}
Task-specific rules: UPC
1. When you confirm that the preference mentioned in the question is outdated or conflicts with the current request, and that the answer to the ASK option can uniquely and unambiguously distinguish one visible entity, choose <ASK> to request the concrete preference dimension that needs confirmation or updating. Do not ask the user to restate all preferences. After receiving the user's reply to a correct <ASK>, continue choosing <WAIT> until the trigger condition specified in the question appears, then choose <ANSWER>. Choose the ASK whose answer can uniquely distinguish one visible entity; incorrect ASKs concern features shared by multiple entities.
\end{promptbox}

\subsubsection{OPEN\_ENDED\_SETTING\_PROMPT}

\begin{promptbox}{KVEM\_PROMPT\_OPEN}
Task-specific rules: KVEM
1. You will receive a video in which some entities are occluded. When you confirm that the relevant entity is occluded, choose <ASK> only to request the missing visual information needed to identify that entity, naming its accurate location or visual referent. Do not merely request an entity already explicitly mentioned in the question, ask for a broad scene description, or ask for an unrelated attribute. Before confirming the occlusion, choose <WAIT>. After the correct ASK, answer when the question's trigger condition is satisfied.

Note: Locate only the occluded entity relevant to the question, not unrelated entities. Do not request the answer-bearing entity by its category name (for example, "show the tea bag"); identify it using its position and visible spatial context instead.

Example:
Question: How many tea bags are on the table?
Visible scene: Several objects are occluded. The relevant occluded object is in the lower left, to the right of a red mug and behind a wooden box; another occluded object in the upper right is unrelated.
Correct open-ended ASK: <ASK:Please reveal the occluded object in the lower left, to the right of the red mug and behind the wooden box.>
\end{promptbox}

\begin{promptbox}{VIM\_PROMPT\_OPEN}
Task-specific rules: VIM
1. Once the question's trigger condition has occurred and you confirm that the current viewpoint cannot answer the question, choose <ASK> to request a concrete camera angle, side, or view that you believe can answer it. Any specific viewpoint that would genuinely provide sufficient evidence is valid; vague or unrelated view requests are not. Before confirming the missing viewpoint, choose <WAIT>. After the correct ASK, answer when sufficient evidence is available.

Note: Describe the needed viewpoint using camera direction, relative position, angle, and spatial coverage. Do not merely request a view of an entity named in the question.

Example:
Question: Which drawer did the person put the key into?
Visible scene: From the current front-facing view, the person's body blocks the lower front of the table.
Correct open-ended ASK: <ASK:Please show an elevated view from the table's front-right corner, angled toward its left side, covering the entire lower front of the table and the area beneath the tabletop.>
\end{promptbox}

\begin{promptbox}{ICM\_PROMPT\_OPEN}
Task-specific rules: ICM
1. When the personalized definitions needed to answer the question are missing, choose <ASK> to select the definition for every private entity alias in the question. You may ASK as soon as the question becomes active; do not wait for the people to appear or become visually distinguishable. List every required alias; explicit request wording is optional. After receiving the definitions, continue choosing <WAIT> until the trigger condition appears, then choose <ANSWER>.

Note: An open-ended ASK selects the personalized definitions for every private entity alias in the question. Name every alias. Listing the aliases alone is sufficient; explicit request wording is optional. Do not ask for a generic identity explanation.

Example:
Question: What did {Noni} give {Buddy}?
Visible scene: The relevant people have appeared, but their private alias mappings have not been provided.
Correct open-ended ASK: <ASK:{Noni} and {Buddy}.>
Also correct: <ASK:Please provide the personalized entity definitions for {Noni} and {Buddy}.>
\end{promptbox}

\begin{promptbox}{PACM\_PROMPT\_OPEN}
Task-specific rules: PACM
1. When the personalized action definitions needed to answer the question are missing, choose <ASK> to select the video definition for every action in the question. List every required action name; explicit request wording is optional. After receiving the definitions, continue choosing <WAIT> until the trigger condition appears, then choose <ANSWER>.

Note: An open-ended ASK selects the video definitions for every personalized action in the question. Name every action. Listing all required action names alone is sufficient; explicit request wording is optional. Do not ask for a generic action definition.

Example:
Question: Did {Action A} happen before {Action B}?
Visible scene: The personalized video action definitions for {Action A} and {Action B} have not been provided.
Correct open-ended ASK: <ASK:{Action A} and {Action B}.>
Also correct: <ASK:Please provide the video action definitions for {Action A} and {Action B}.>
\end{promptbox}

\begin{promptbox}{AUQ\_PROMPT\_OPEN}
Task-specific rules: AUQ
1. If, together with the video content, you confirm that the question is ambiguous, choose <ASK> to identify the specific ambiguous part and request the clarification needed to resolve the question to one unique referent rather than multiple possible referents. Do not make a generic request for more information. After the correct ASK, continue choosing <WAIT> until the trigger condition specified in the question appears, then choose <ANSWER>.
2. AUQ is an exact-reproduction task. Reproduce the specified reminder message exactly, including capitalization and punctuation, without paraphrasing or adding words.

Note: An open-ended ASK must address a specific ambiguity so the user can uniquely identify the intended referent.

Example:
Question: Let me know when that person picks up the cup.
Visible scene: Two people are present, one in white on the left and one in black on the right.
Correct open-ended ASK: <ASK:Do you mean the person in white on the left or the person in black on the right?>
\end{promptbox}

\begin{promptbox}{WUQ\_PROMPT\_OPEN}
Task-specific rules: WUQ
1. If the question's trigger condition appears and you confirm that the question contains a visual error because the video content does not match it, choose <ASK> to identify the specific false or incorrect part, stating the visually supported correction or requesting confirmation of it. Do not make a generic request. Before that evidence appears, choose <WAIT>. After the correct ASK, answer when the corrected trigger condition is satisfied.

Note: An open-ended ASK must identify the specific contradiction between the question and the video, either by stating the visually supported correction or by asking the user to confirm that correction. Both forms are valid.

Example:
Question: Let me know when the person in the red jacket picks up the cup.
Visible scene: The person in question is wearing a blue jacket.
Correct open-ended ASK: <ASK:The question says red jacket, but the person in the video is wearing a blue jacket. Did you mean the person in the blue jacket?>
\end{promptbox}

\begin{promptbox}{UPM\_PROMPT\_OPEN}
Task-specific rules: UPM
1. When you confirm that answering the question requires knowing the user's personal preference, choose <ASK> to request the concrete preference dimension needed to answer it, such as a particular attribute, style, or choice criterion. Do not ask for preferences generically. After receiving the user's reply to a correct <ASK>, continue choosing <WAIT> until the trigger condition specified in the question appears, then choose <ANSWER>.

Note: Multiple visible entities can differ in color, material, shape, style, and other dimensions. Ask about a feature whose possible values uniquely identify one entity, not a feature shared by multiple entities.

Example:
Question: Which of these three chairs would suit me best?
Visible scene: The chairs have overlapping colors, but their materials are wood, metal, and upholstery, one chair per material.
Correct open-ended ASK: <ASK:For the chair's material, do you prefer wood, metal, or upholstery?>
\end{promptbox}

\begin{promptbox}{UPC\_PROMPT\_OPEN}
Task-specific rules: UPC
1. When you confirm that the preference mentioned in the question is outdated or conflicts with the current request, choose <ASK> to request the concrete preference dimension that needs confirmation or updating. Do not ask the user to restate all preferences. After receiving the user's reply to a correct <ASK>, continue choosing <WAIT> until the trigger condition specified in the question appears, then choose <ANSWER>.

Note: Multiple visible entities can have features along several dimensions. Ask for an updated preference on a dimension whose possible values uniquely identify one entity, not a feature shared by multiple entities.

Example:
Question: Which of these three table lamps would suit me best?
Previous preference: The user preferred black lamps, but all three lamps are black.
Visible scene: The lampshades are round, conical, and square, one lamp per shape.
Correct open-ended ASK: <ASK:Do you now prefer a round, conical, or square lampshade?>
\end{promptbox}

\subsubsection{JUDGE\_PROMPT}

\begin{promptbox}{KVEM\_ASK\_JUDGE\_PROMPT}
You are the strict open-ended ASK judge for the KVEM task.
Using the supplied image, question, and reference, identify the single intended
occluded entity in the image. The candidate is correct only if it requests the
same entity as the reference and locates that entity by position or visible
spatial context. Return 0 if the
candidate merely requests the answer-bearing entity by its category name,
merely repeats an entity already explicitly mentioned in the question, asks
for a broad scene description, targets an unrelated entity or attribute, adds
an unrelated request, or gives missing or conflicting location information.
Return exactly 1 if the candidate is correct and exactly 0 if the candidate is incorrect. Return only that single digit and nothing else. Judge only the supplied labeled text and images; do not use outside knowledge.
\end{promptbox}

\begin{promptbox}{VIM\_ASK\_JUDGE\_PROMPT}
You are the strict open-ended ASK judge for the VIM task.
You will receive the question, the candidate ASK, and multi-view video frames
collected at the ASK timestamp to help you judge the candidate. The first
labeled frame is the initial viewpoint shown to the tested model. The remaining
labeled frames are the gold viewpoints from the annotation. Decide whether,
based on the current video frames, switching to the viewpoint requested by the
candidate would allow the current question to be answered. The candidate is
correct only if it requests a concrete useful viewpoint and the current video
frames indicate that switching to that viewpoint would answer the current
question. Return 0 for a vague or unrelated view, a view that the current
video frames indicate would remain insufficient, an unrelated additional
request, or a candidate that directly requests an entity already explicitly
mentioned in the question instead of requesting a useful viewpoint.
Return exactly 1 if the candidate is correct and exactly 0 if the candidate is incorrect. Return only that single digit and nothing else. Judge only the supplied labeled text and images; do not use outside knowledge.
\end{promptbox}

\begin{promptbox}{ICM\_ASK\_JUDGE\_PROMPT}
You are the strict open-ended ASK judge for the ICM task.
The ASK payload selects the private entity definitions to return. It must name
every private entity alias referred to in the question. Listing those aliases
alone is sufficient: for example, "{Rhiley} and {Jaedyn}." is valid when both
aliases occur in the question. Explicit requests for their definitions are
equally valid; request wording is not required. Reject omitted or wrong aliases,
unrelated additions, and answers that do not select the required definitions.
Return exactly 1 if the candidate is correct and exactly 0 if the candidate is incorrect. Return only that single digit and nothing else. Judge only the supplied labeled text; do not use outside knowledge.
\end{promptbox}

\begin{promptbox}{PACM\_ASK\_JUDGE\_PROMPT}
You are the strict open-ended ASK judge for the PACM task.
The ASK payload selects the personalized action definitions to return. It must
explicitly name every personalized action referred to in the question. A list
of all required action names alone is sufficient; explicit definition-request
wording is optional. Reject omitted or wrong action names, unrelated additions,
and answers that do not select the required definitions.
Return exactly 1 if the candidate is correct and exactly 0 if the candidate is incorrect. Return only that single digit and nothing else. Judge only the supplied labeled text and images; do not use outside knowledge.
\end{promptbox}

\begin{promptbox}{AUQ\_ASK\_JUDGE\_PROMPT}
You are the strict open-ended ASK judge for the AUQ task.
The user's request contains an ambiguous part. Using
the supplied image, question, and reference as ground truth, decide whether the
candidate correctly identifies the specific ambiguity and requests the
clarification needed for the request to identify one unique referent rather
than multiple possible referents. A generic request for more information, an
unrelated request, or an answer to the question is incorrect.
Return exactly 1 if the candidate is correct and exactly 0 if the candidate is incorrect. Return only that single digit and nothing else. Judge only the supplied labeled text and images; do not use outside knowledge.
\end{promptbox}

\begin{promptbox}{WUQ\_ASK\_JUDGE\_PROMPT}
You are the strict open-ended ASK judge for the WUQ task.
The user's request contains a part that conflicts with the
video. Using the supplied image, question, and reference as ground truth,
decide whether the candidate correctly identifies the specific erroneous part
and challenges it by stating the visually supported correction or asking the
user to confirm that correction. A generic request that does not identify the
conflict, a correction of a different claim, or an unrelated additional
request is incorrect.
Return exactly 1 if the candidate is correct and exactly 0 if the candidate is incorrect. Return only that single digit and nothing else. Judge only the supplied labeled text and images; do not use outside knowledge.
\end{promptbox}

\begin{promptbox}{UPM\_ASK\_JUDGE\_PROMPT}
You are the strict open-ended ASK judge for the UPM task.
Using only the supplied ordered option images, Correct
Image label, and candidate, decide whether the candidate asks about one
concrete, visually grounded user-preference feature dimension. The value shown
by the Correct Image on that dimension must not be shared by any other option
image, so a user answer choosing that value would uniquely identify the Correct
Image. Return 0 if the requested feature is absent from or not visually
comparable across the option images, the Correct Image's value is shared by
another image, the candidate excludes or conflicts with the Correct Image's
value, the request uniquely identifies only a different image, combines
multiple preference dimensions, adds an unrelated request, or asks about
preferences only in general terms.
Return exactly 1 if the candidate is correct and exactly 0 if the candidate is incorrect. Return only that single digit and nothing else. Judge only the supplied labeled text and images; do not use outside knowledge.
\end{promptbox}

\begin{promptbox}{UPC\_ASK\_JUDGE\_PROMPT}
You are the strict open-ended ASK judge for the UPC task.
Using only the supplied ordered option images,
Correct Image label, and candidate, decide whether the candidate asks the user
to confirm or update one concrete, visually grounded preference feature
dimension. The value shown by the Correct Image on that dimension must not be
shared by any other option image, so a user answer choosing that value would
uniquely identify the Correct Image. Return 0 if the candidate does not ask for
a current or updated preference, the requested feature is absent from or not
visually comparable across the option images, the Correct Image's value is
shared by another image, the candidate excludes or conflicts with the Correct
Image's value, the request uniquely identifies only a different image,
combines multiple preference dimensions, adds an unrelated request, or asks
about preferences only in general terms.
Return exactly 1 if the candidate is correct and exactly 0 if the candidate is incorrect. Return only that single digit and nothing else. Judge only the supplied labeled text and images; do not use outside knowledge.
\end{promptbox}

\clearpage

\clearpage

\subsection{Study Case}

\begin{videoexamplebox}

%
%

\noindent
\frameitem{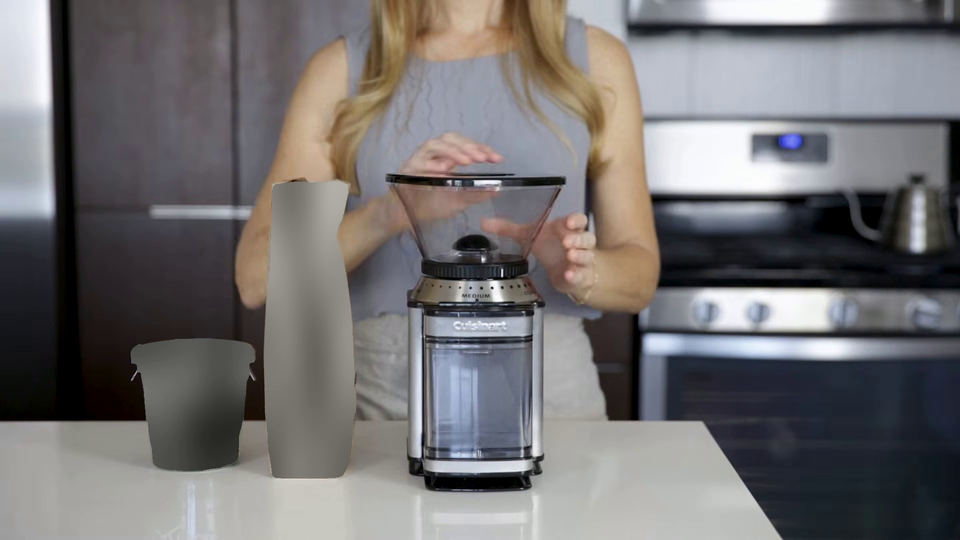}
{\textbf{WAIT} 56s}
\hfill
\frameitem{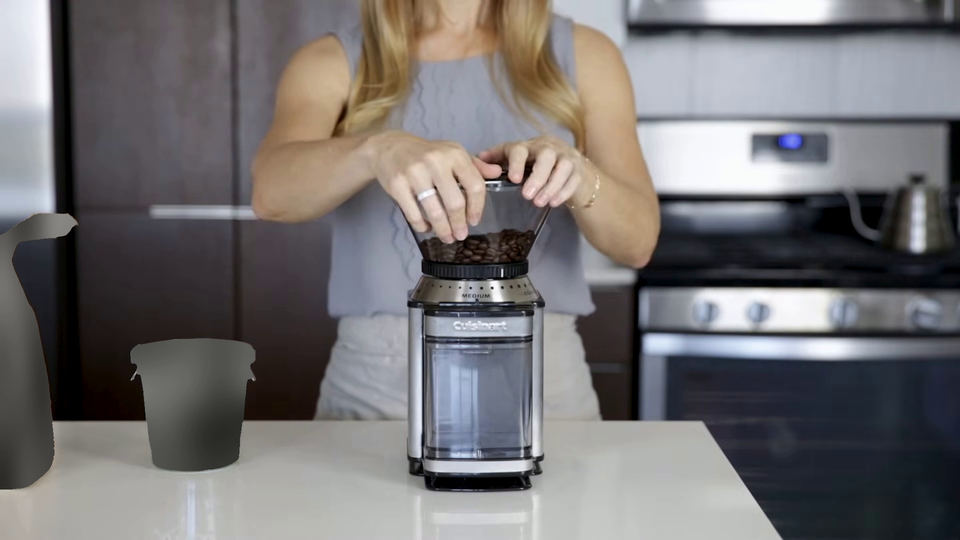}
{\textbf{WAIT} 67s}
\hfill
\frameitem{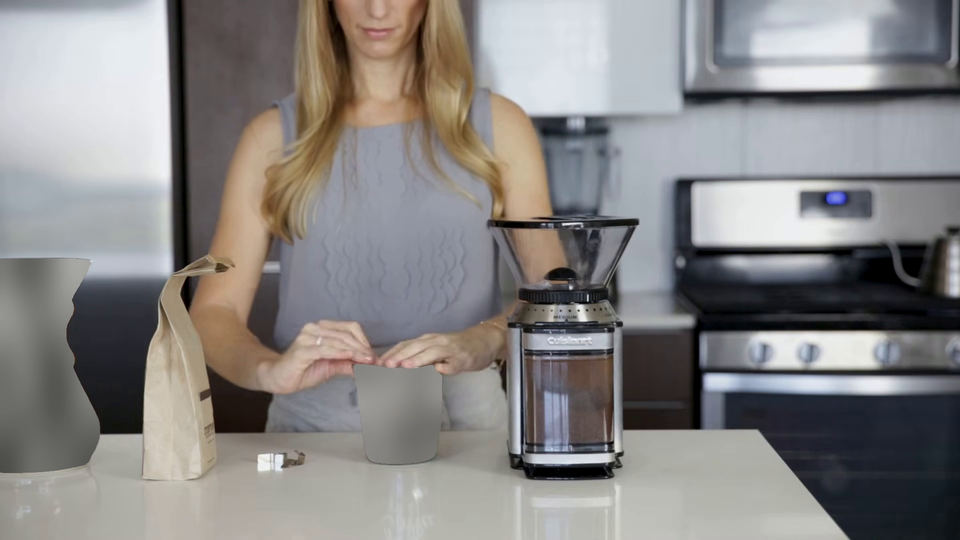}
{\textbf{WAIT} 78s}
\hfill
\frameitem{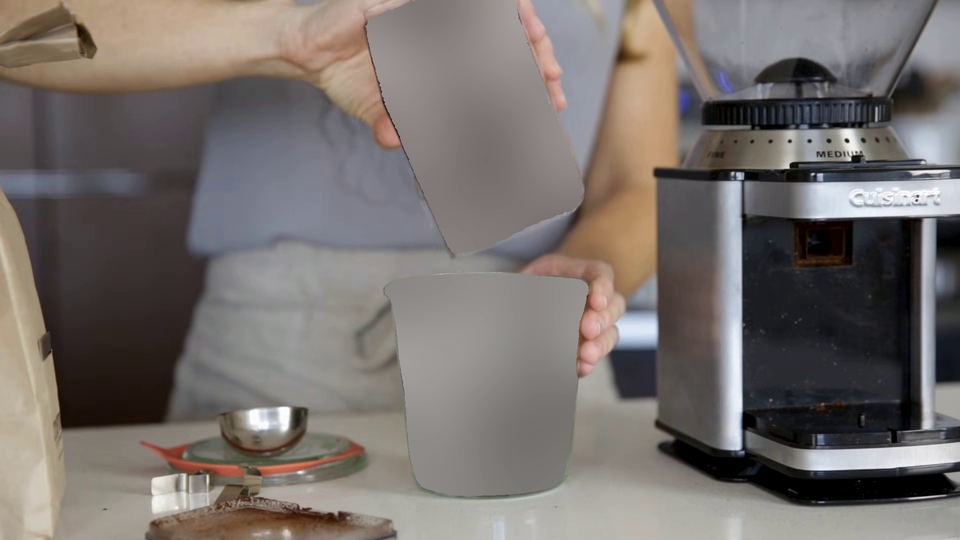}
{\textbf{WAIT} 89s}
\hfill
\frameitem{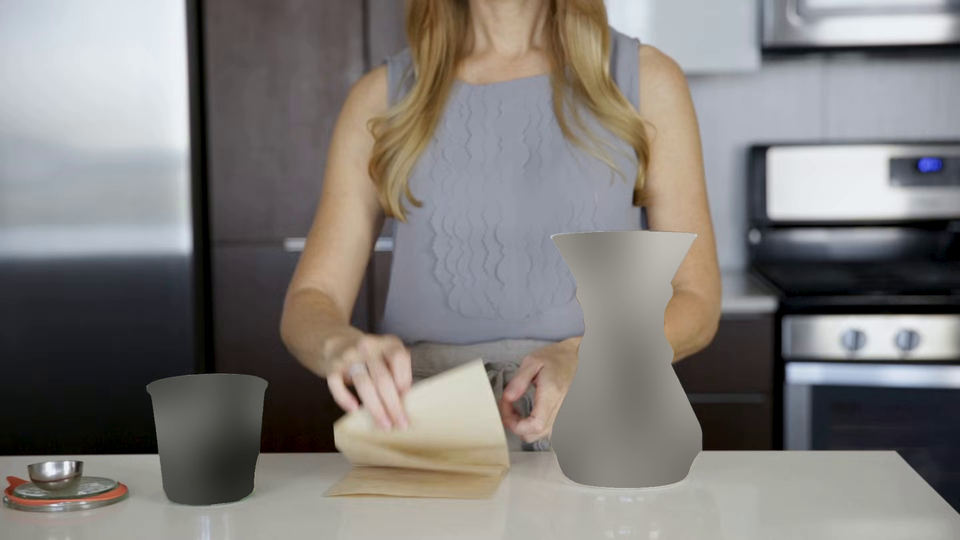}
{\textbf{ANSWER} 89s}
\hfill
\frameitem{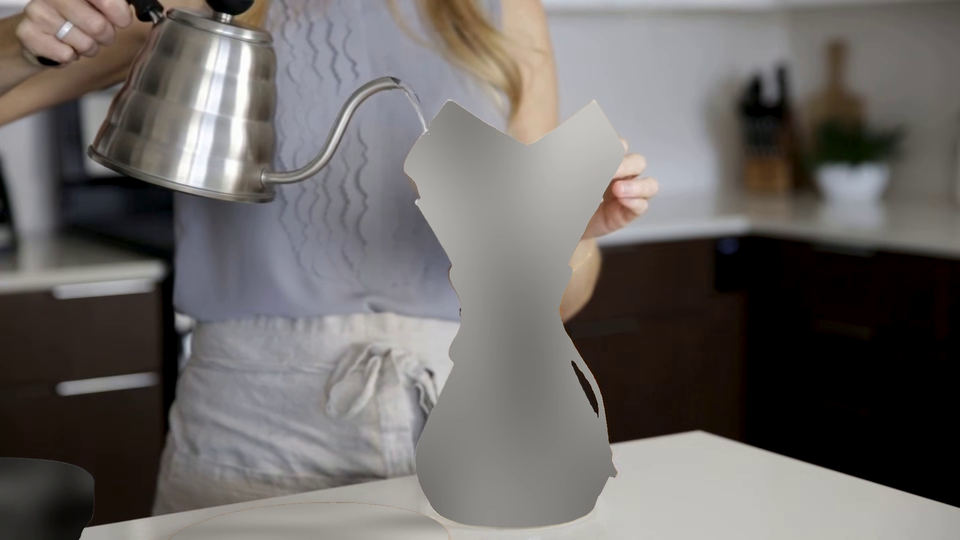}
{\textbf{ANSWER} 106s}
\hfill
\frameitem{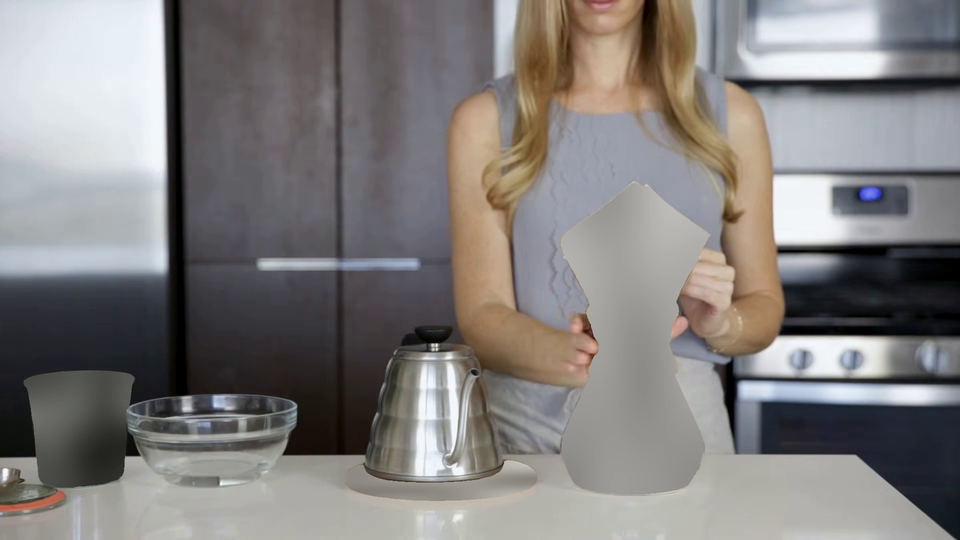}
{\textbf{ANSWER} 123s}


\noindent
\frameitem{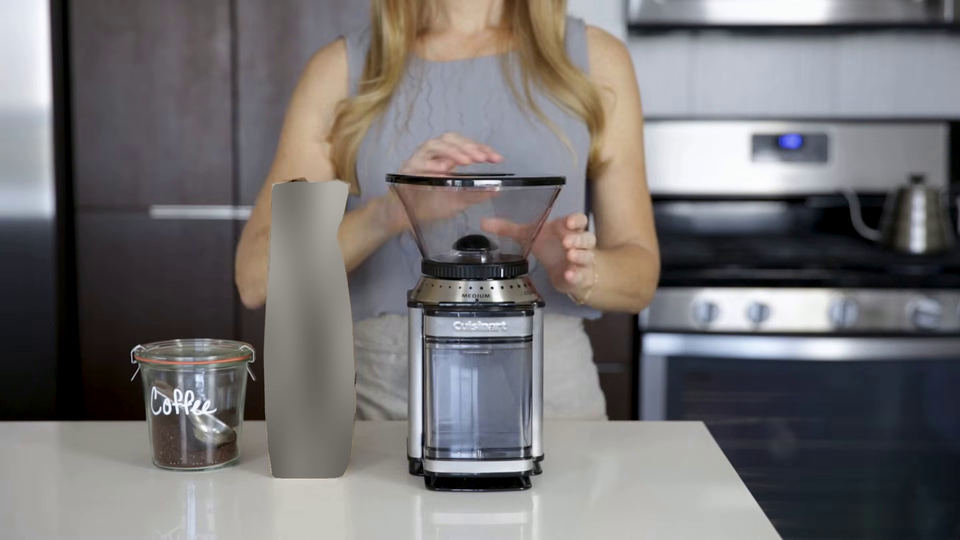}
{\textbf{gold} 56s}
\hfill
\frameitem{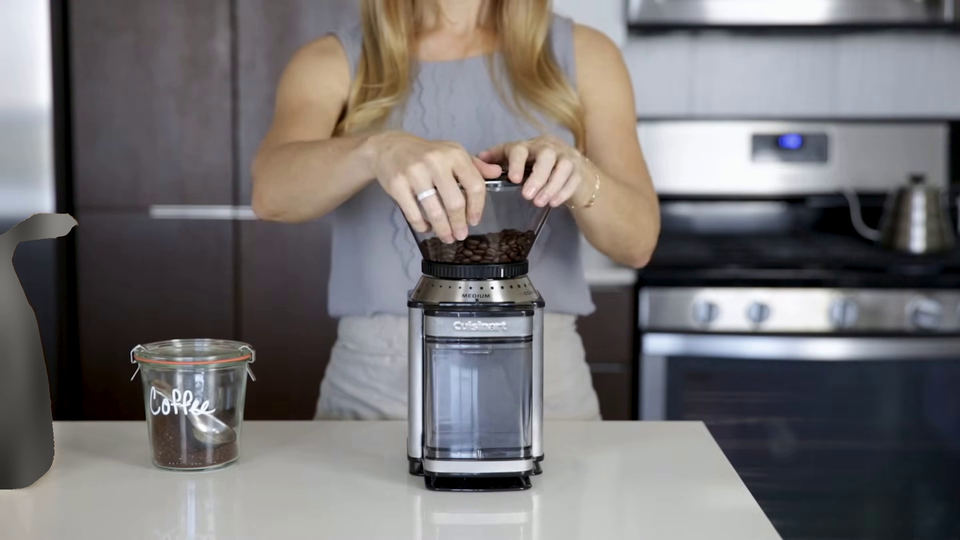}
{\textbf{gold} 67s}
\hfill
\frameitem{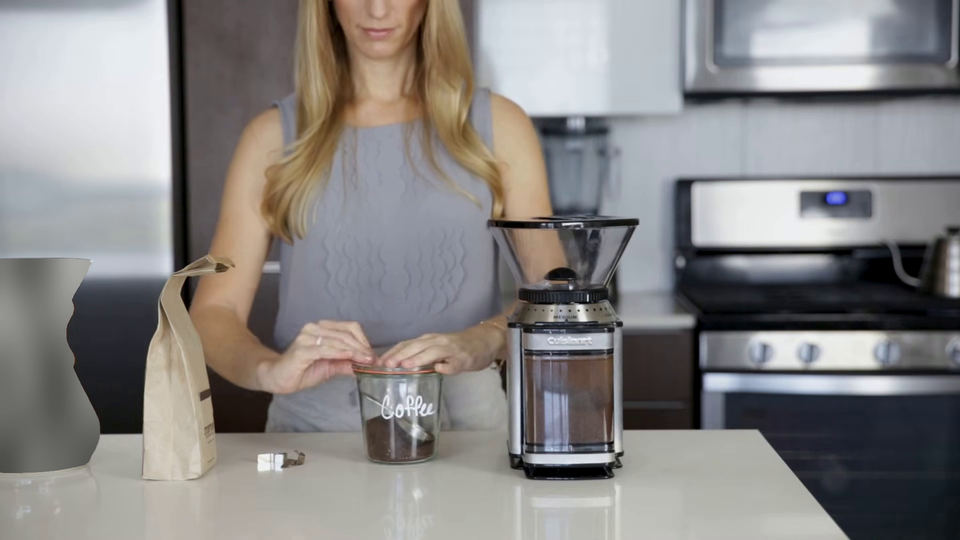}
{\textbf{gold} 78s}
\hfill
\frameitem{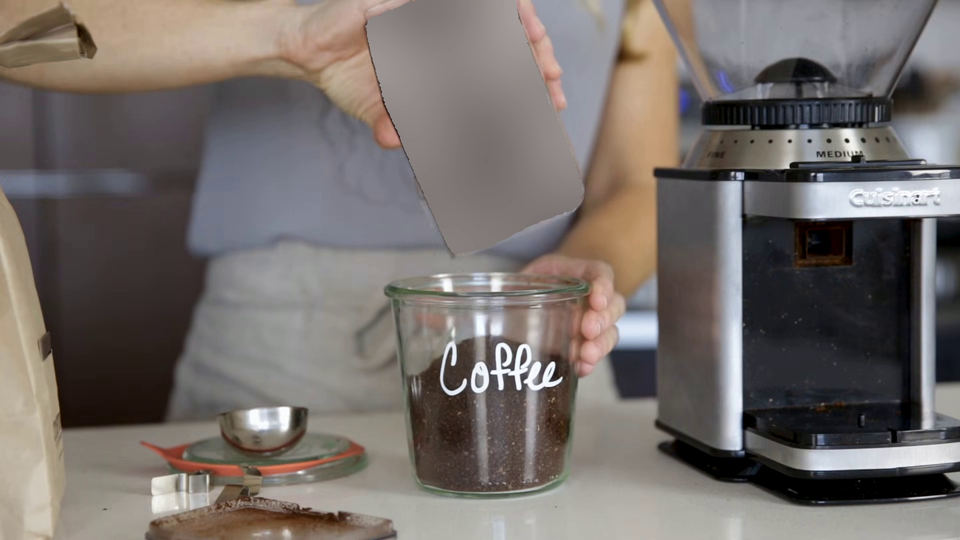}
{\textbf{gold} 89s}
\hfill
\frameitem{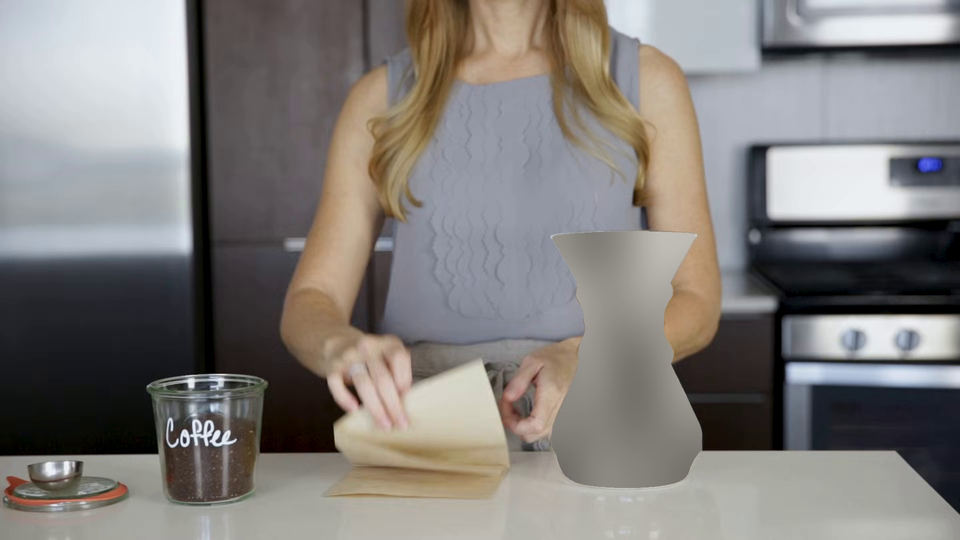}
{\textbf{gold} 89s}
\hfill
\frameitem{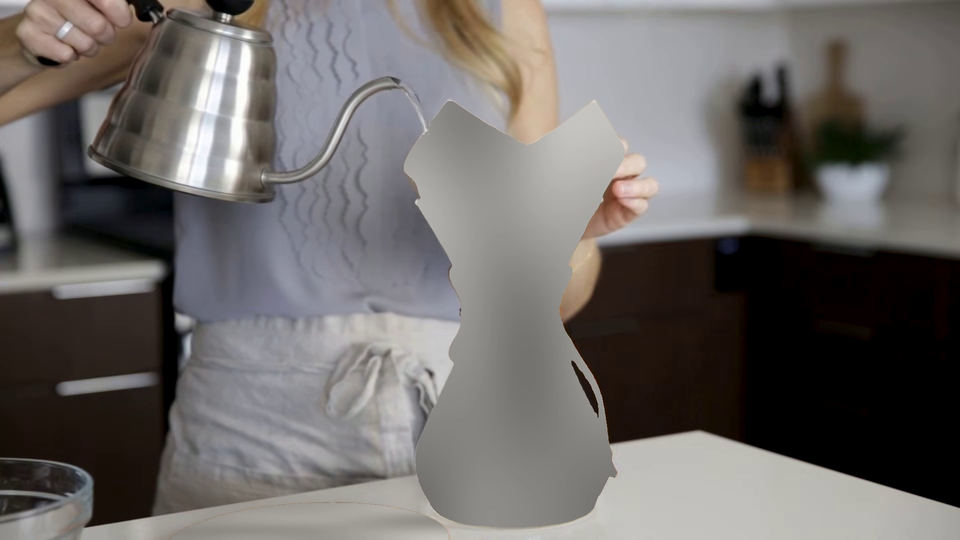}
{\textbf{gold} 106s}
\hfill
\frameitem{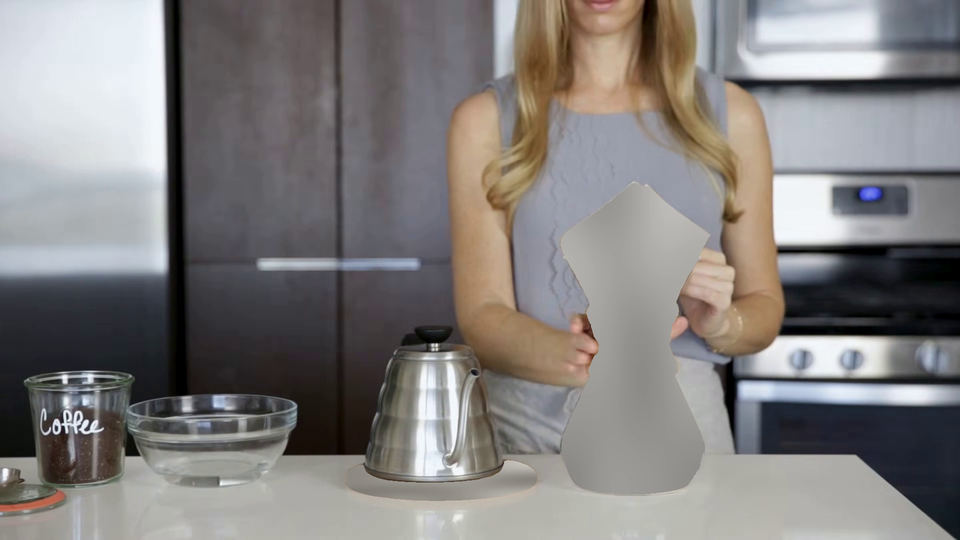}
{\textbf{gold} 123s}


\noindent
\textbf{Question:}
After the coffee beans have been ground and poured into the container,
tell me how much ground coffee is in the container.


\noindent
\textbf{Action Options:}

\begin{tabularx}{\linewidth}{
    @{}
    >{\raggedright\arraybackslash}p{0.035\linewidth}
    >{\raggedright\arraybackslash}X
    @{}
}

A. &
Request a clear view of the object on the counter to the woman's left.
\\[2pt]

\textbf{B.} &
\textbf{Request a clear view of the object on the counter to the woman's right.}
\\[2pt]

C. &
Request a clear view of the objects on the counter on either side of the woman.
\\[2pt]

D. &
Request a clear view of the object on the rear counter behind the woman's left shoulder.
\\[2pt]

E. &
Request a clear view of the object at the far right beside the cooking area.
\\[2pt]

F. &
Request a clear view of the object directly behind the papers at the center of the counter.
\\

\end{tabularx}


\noindent
\textbf{Answer Options:}

\begin{tabular*}{\linewidth}{
    @{\extracolsep{\fill}}
    l
    l
    l
    l
    l
    @{}
}

A. Nearly full
&
B. About one-third full
&
\textbf{C. Slightly more than half full}
&
D. ...
&
E. ...
\end{tabular*}

\end{videoexamplebox}

\begin{videoexamplebox}


\noindent
\frameitem{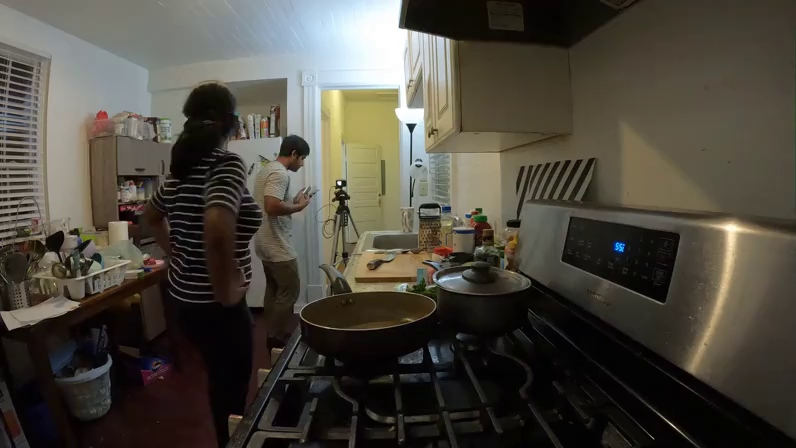}
{\textbf{WAIT} 0s}
\hfill
\frameitem{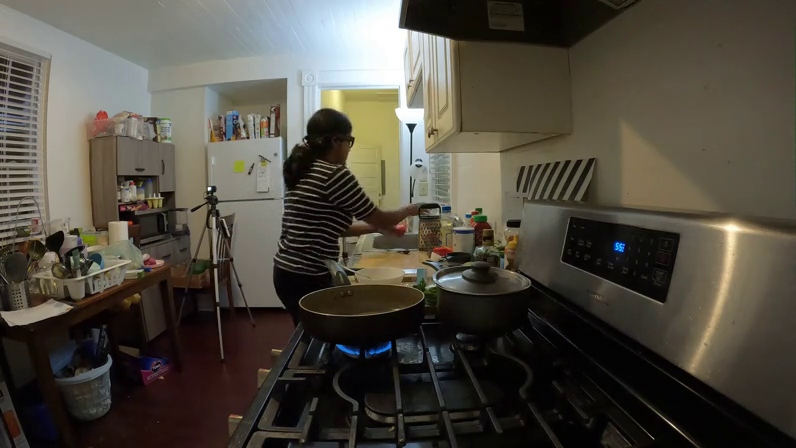}
{\textbf{WAIT} 54s}
\hfill
\frameitem{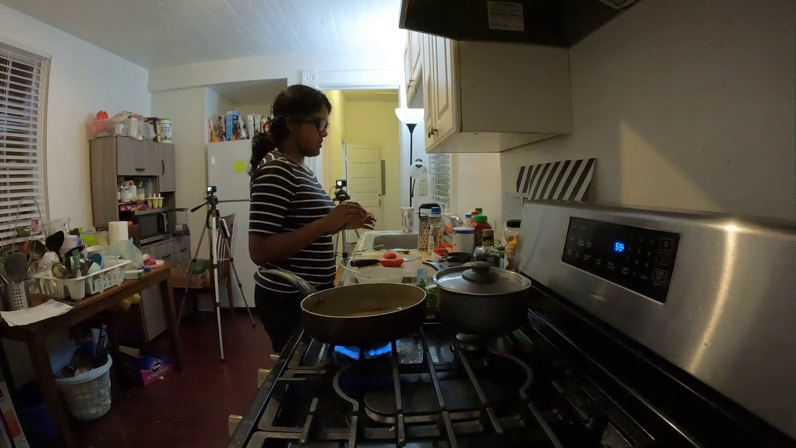}
{\textbf{WAIT} 107s}
\hfill
\frameitem{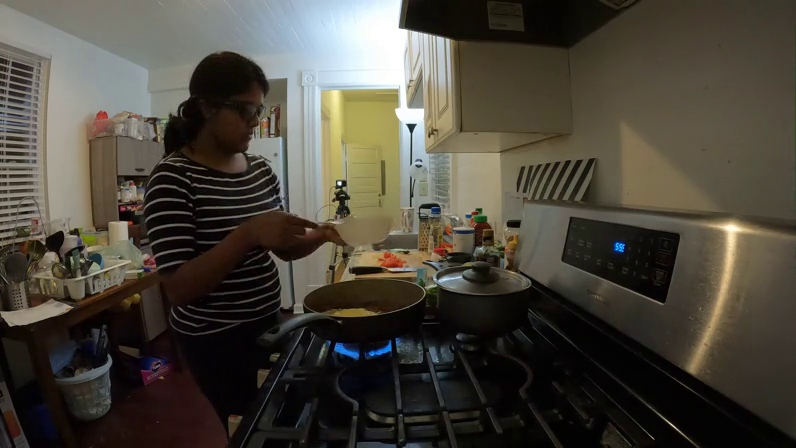}
{\textbf{WAIT} 161s}
\hfill
\frameitem{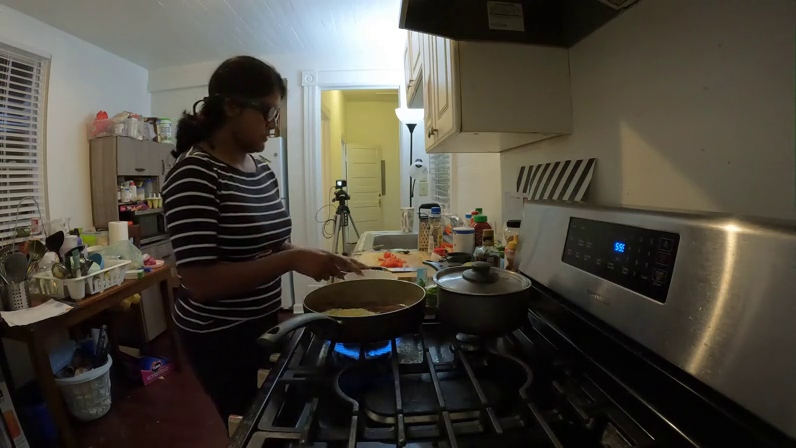}
{\textbf{ANSWER} 162s}
\hfill
\frameitem{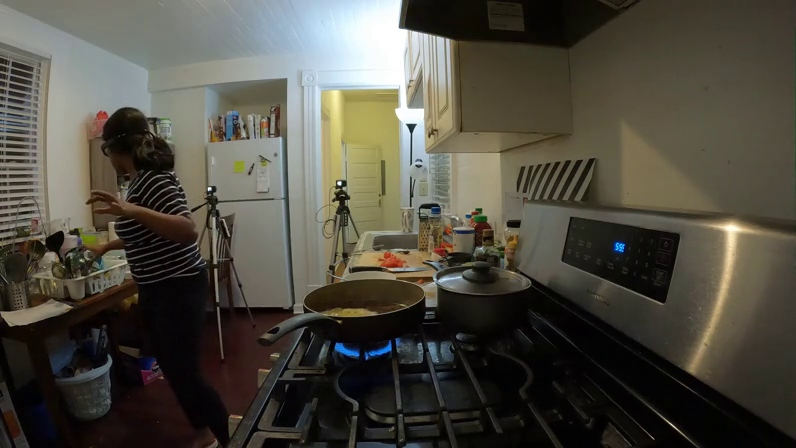}
{\textbf{ANSWER} 168s}
\hfill
\frameitem{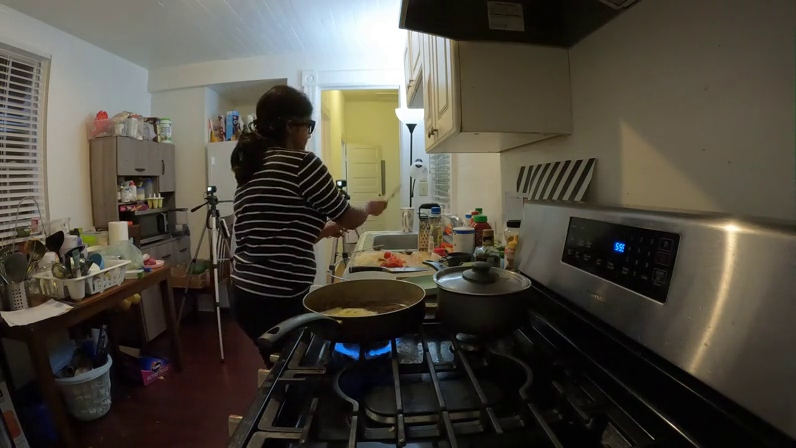}
{\textbf{ANSWER} 174s}


\noindent
\frameitem{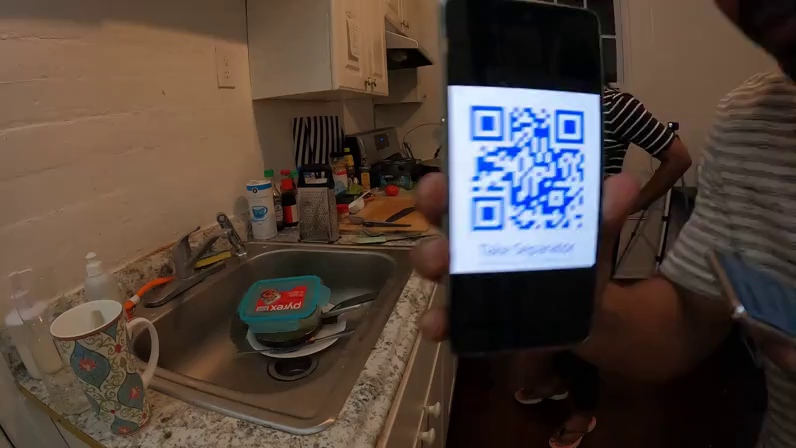}
{gold: 0s}
\hfill
\frameitem{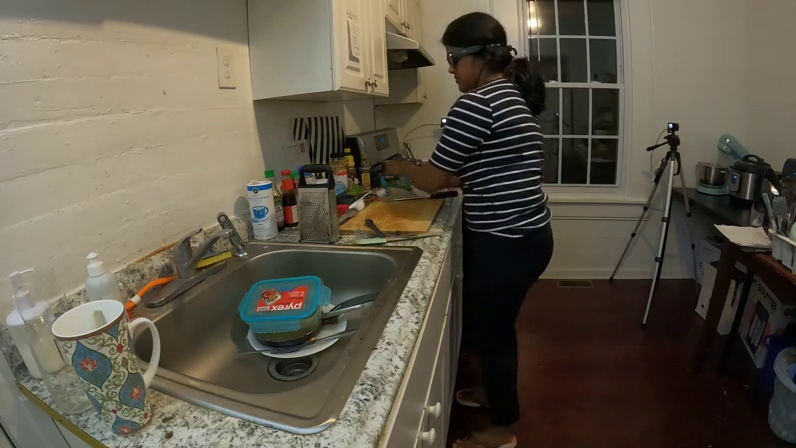}
{gold: 54s}
\hfill
\frameitem{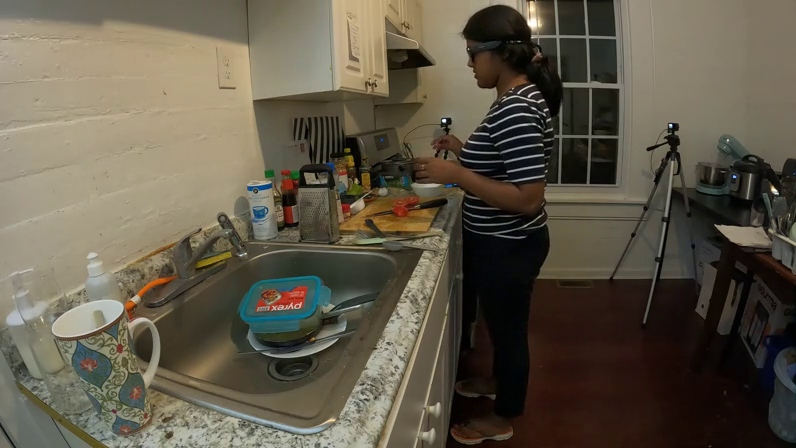}
{gold: 107s}
\hfill
\frameitem{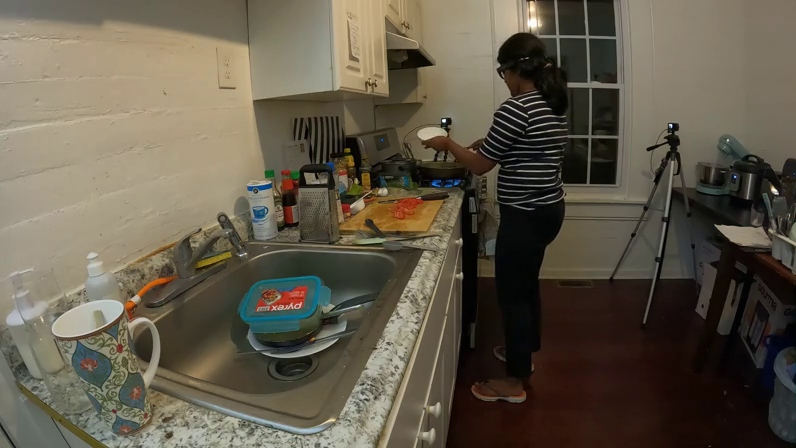}
{gold: 161s}
\hfill
\frameitem{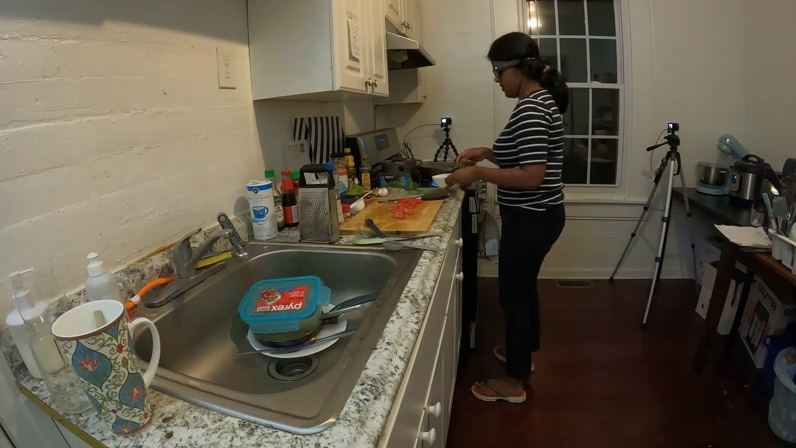}
{gold: 162s}
\hfill
\frameitem{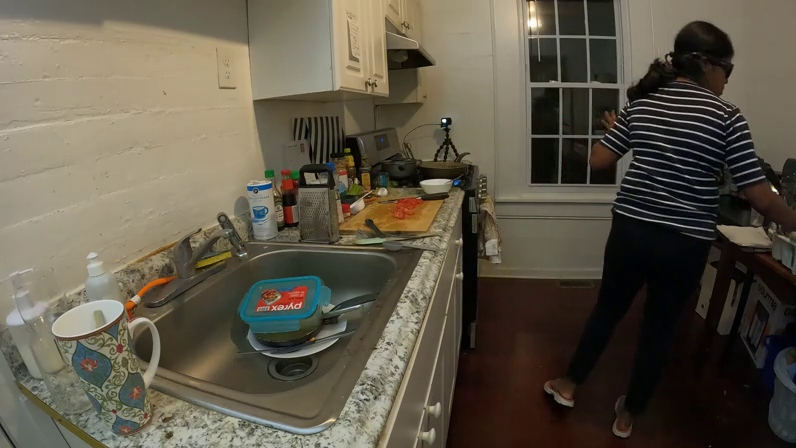}
{gold: 168s}
\hfill
\frameitem{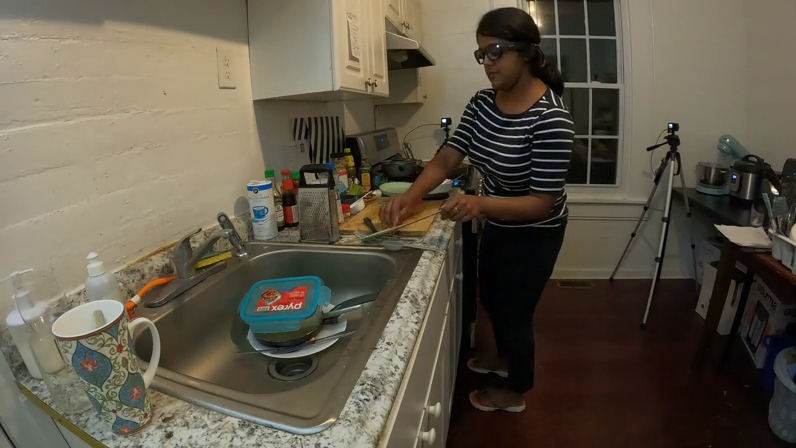}
{gold: 174s}


\noindent
\textbf{Question:}
When she has poured the beaten egg mixture into the frying pan, tell me
what color the lid of the food storage container with a plastic lid
sitting in the sink is?


\noindent
\textbf{Action Options:}

\begin{tabularx}{\linewidth}{
    @{}
    >{\raggedright\arraybackslash}p{0.035\linewidth}
    >{\raggedright\arraybackslash}X
    @{}
}

A. &
\texttt{aria03/rgb} --- ``Please give me the view of the near central and
right-side surfaces from above, with the recessed region at the far left
cut off by the frame edge and its interior partly occluded in the
foreground.''
\\[3pt]

B. &
\texttt{cam01/0} (Initial view) --- ``Please give me the view of the area
seen from beside the stovetop looking diagonally across the foreground
toward the left side of the room, with the far-left counter visible but
the lower recessed region beyond it outside the frame.''
\\[3pt]

\textbf{C.} &
\textbf{\texttt{cam02/0} (Gold) --- ``Please give me the view of the area
seen from the near end of the counter looking along it toward the far
wall, with the sink fully framed in the left foreground and its contents
visible from above.''}
\\[3pt]

D. &
\texttt{cam03/0} --- ``Please give me the view of the room from the
front-right side looking diagonally toward the left, with the near-left
recessed region partly cut off and its interior occluded below the
foreground edge.''
\\[3pt]

E. &
\texttt{cam04/0} --- ``Please give me the view of the room from across the
foreground looking toward the back wall, with the far-left recessed
counter region distant and partially occluded along the central line of
sight.''
\\

\end{tabularx}

\noindent
\textbf{Answer Options:}

\begin{tabularx}{\linewidth}{
    @{}
    >{\centering\arraybackslash}X
    >{\centering\arraybackslash}X
    >{\centering\arraybackslash}X
    >{\centering\arraybackslash}X
    @{}
}
A. Yellow &
B. Gree &
C. Red &
\textbf{D. Blue}

\end{tabularx}

\end{videoexamplebox}

\begin{videoexamplebox}

%
%
%

\noindent
\frameitem{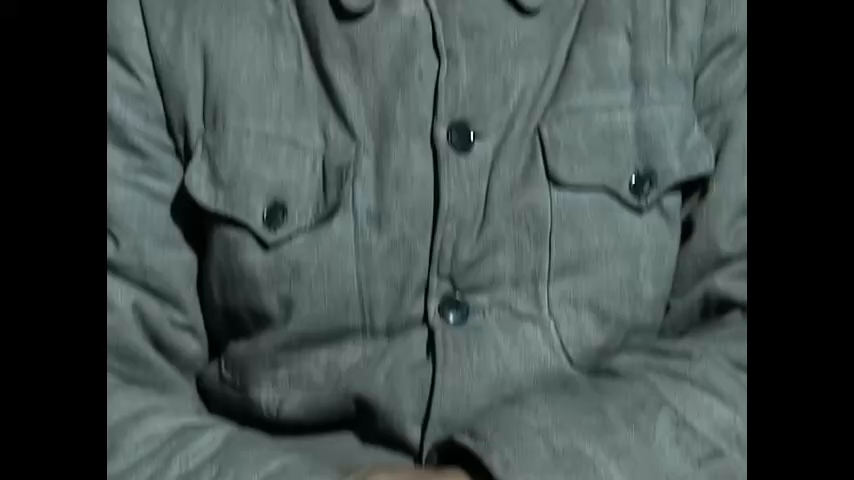}
{\textbf{WAIT} 0s}
\hfill
\frameitem{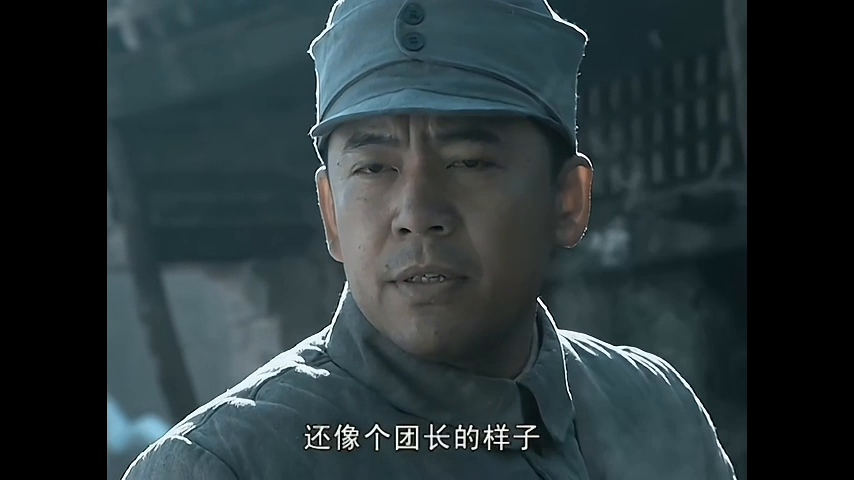}
{\textbf{ASK} 22s}
\hfill
\frameitem{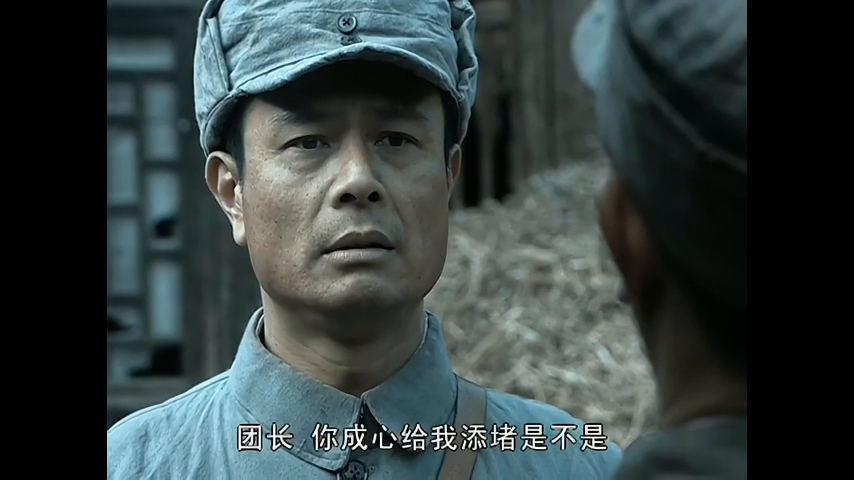}
{\textbf{ASK} 43s}
\hfill
\frameitem{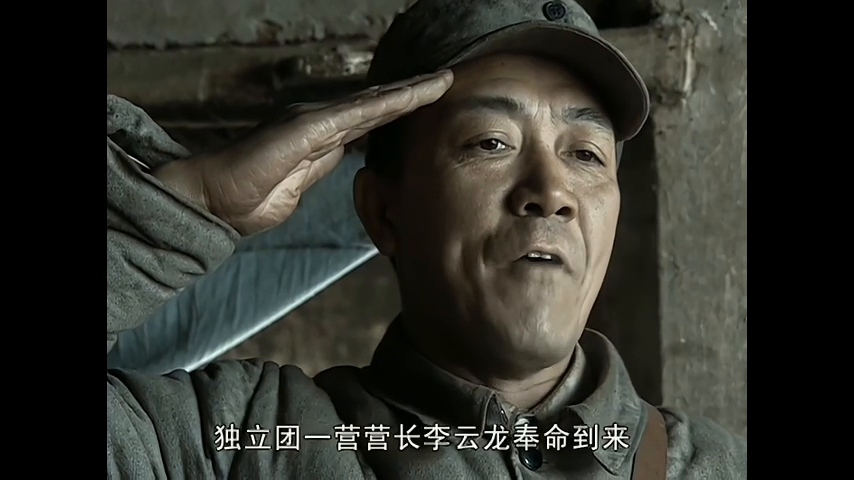}
{\textbf{ASK} 65s}
\hfill
\frameitem{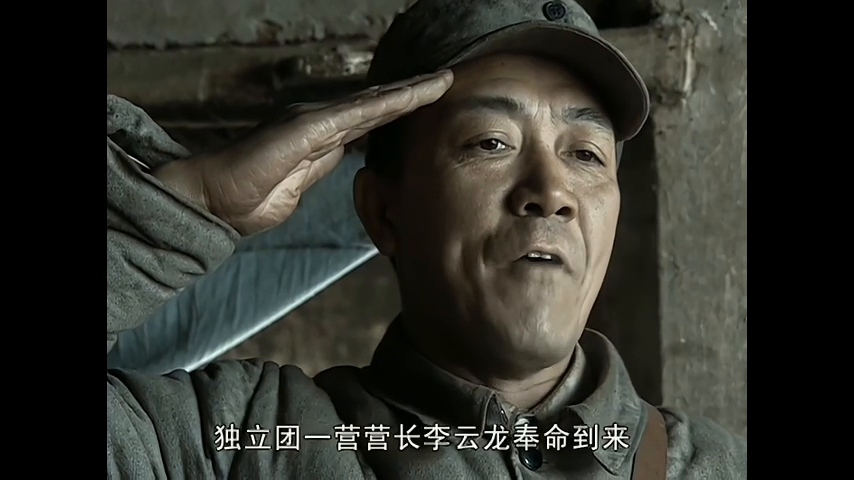}
{\textbf{ANSWER} 65s}
\hfill
\frameitem{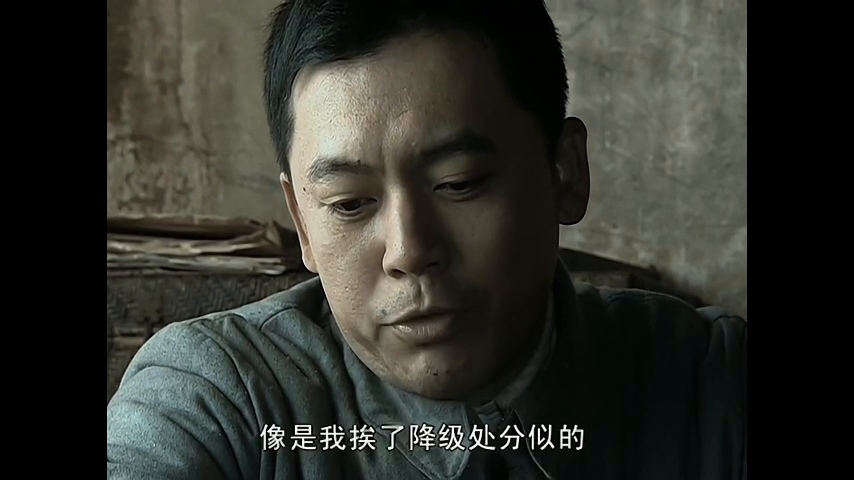}
{\textbf{ANSWER} 133s}
\hfill
\frameitem{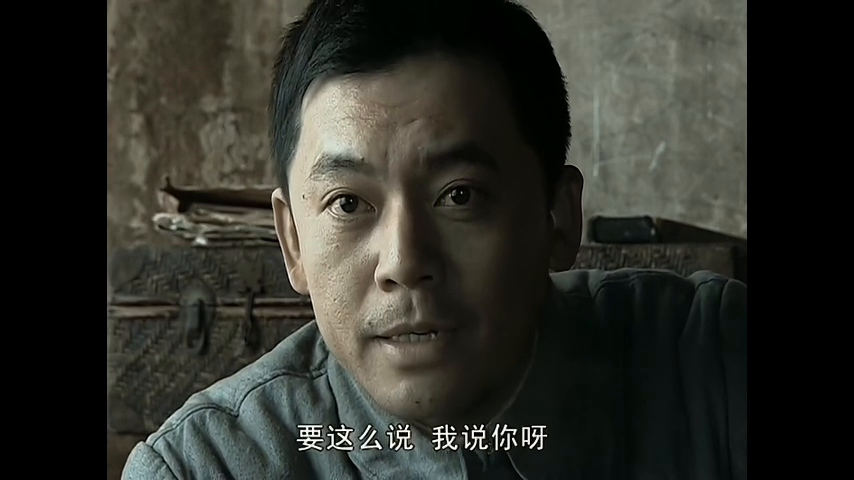}
{\textbf{ANSWER} 201s}


\noindent
\begin{minipage}[c]{0.14\linewidth}
    \centering
    \fcolorbox{exampleborder}{framebg}{%
        \includegraphics[
            width=0.90\linewidth,
            height=1.75cm,
            keepaspectratio
        ]{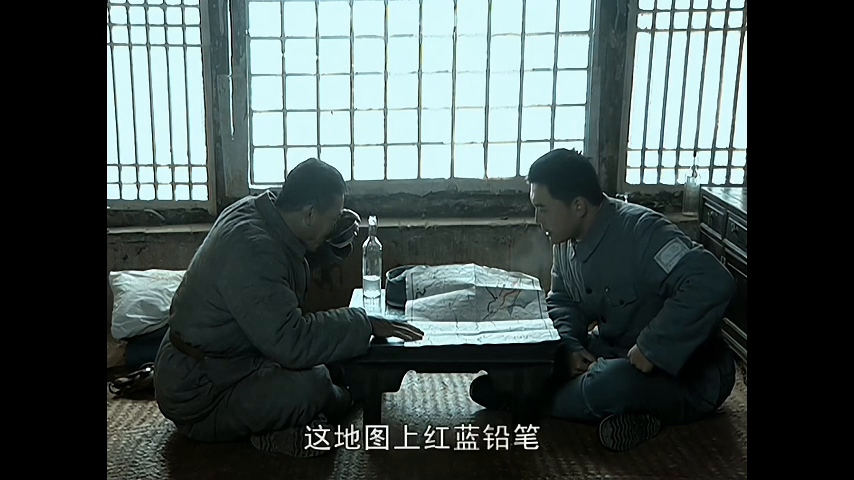}%
    }\\[-1pt]
    {\scriptsize gold:15s}
\end{minipage}
\hspace{3mm}
\begin{minipage}[c]{0.80\linewidth}
    \small
    The person on the right is \textbf{Charity}.
\end{minipage}


\noindent
\textbf{Question:}
When the bottle was on the table in front of Charity, tell me what was in it?


\noindent
\textbf{Action Options:}

\begin{tabularx}{\linewidth}{
    @{}
    >{\raggedright\arraybackslash}p{0.035\linewidth}
    >{\raggedright\arraybackslash}X
    @{}
}

A. &
Is Charity an adult man with a broad oval face, long blond hair, dark eyes, and a broad nose?
\\[2pt]

B. &
Is Charity an adult man with a narrow oval face, short black hair, dark eyes, and a broad nose?
\\[2pt]

\textbf{C.} &
\textbf{Is Charity an adult man with a broad oval face, short black hair, dark eyes, and a broad nose?}
\\[2pt]

D. &
Is Charity an adult man with a broad oval face, short black hair, light blue eyes, and a broad nose?
\\

\end{tabularx}


\noindent
\textbf{Answer Options:}

\begin{tabularx}{\linewidth}{
    @{}
    >{\centering\arraybackslash}X
    >{\centering\arraybackslash}X
    >{\centering\arraybackslash}X
    >{\centering\arraybackslash}X
    @{}
}
A. Milk &
B. Juice &
C. Coffee &
\textbf{D. Alcohol}

\end{tabularx}

\end{videoexamplebox}

\begin{videoexamplebox}

%
%

\noindent
\frameitem{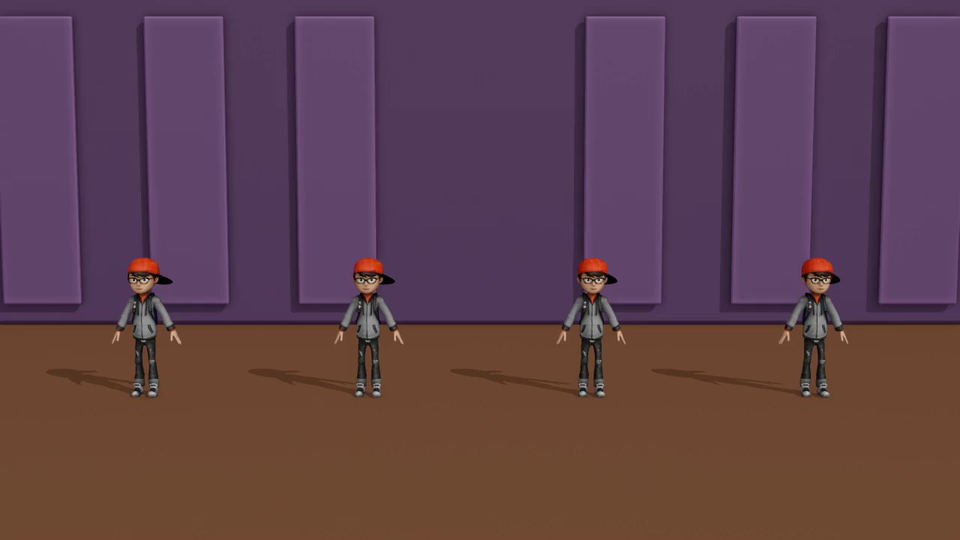}
{\textbf{ASK} 0s}
\hfill
\frameitem{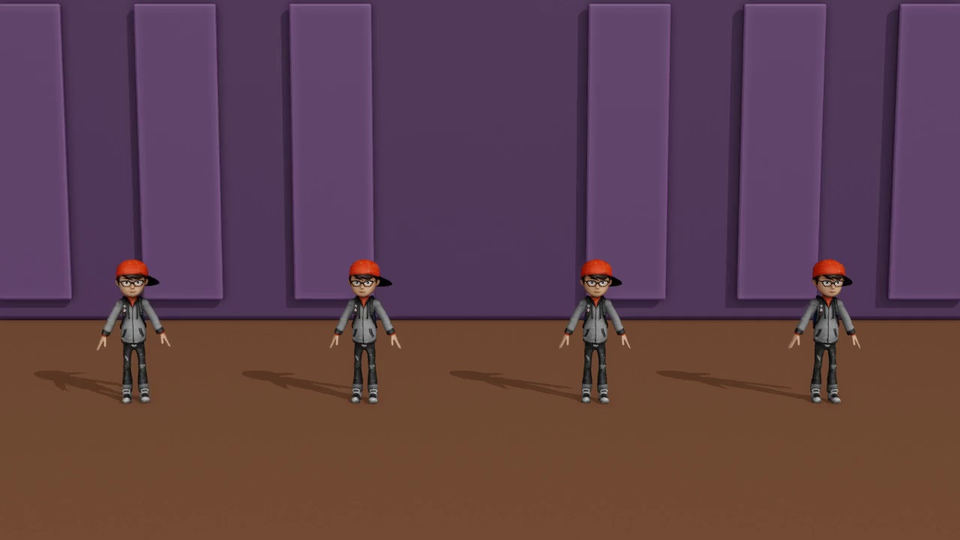}
{\textbf{ASK} 25s}
\hfill
\frameitem{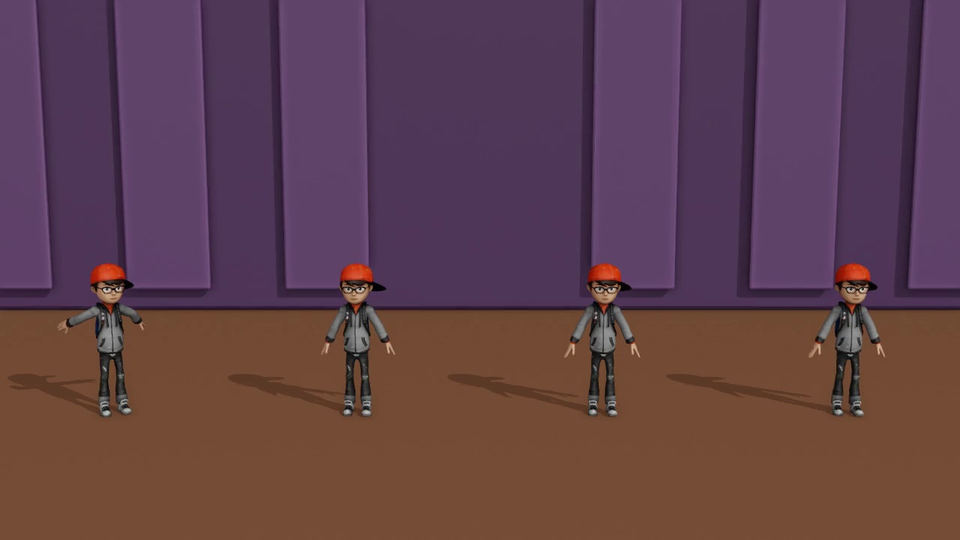}
{\textbf{ASK} 50s}
\hfill
\frameitem{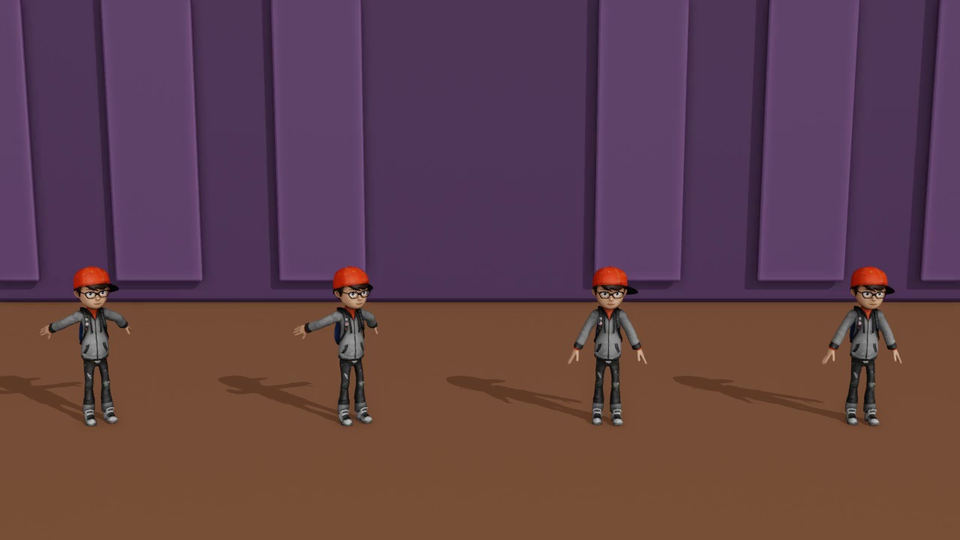}
{\textbf{ASK} 75s}
\hfill
\frameitem{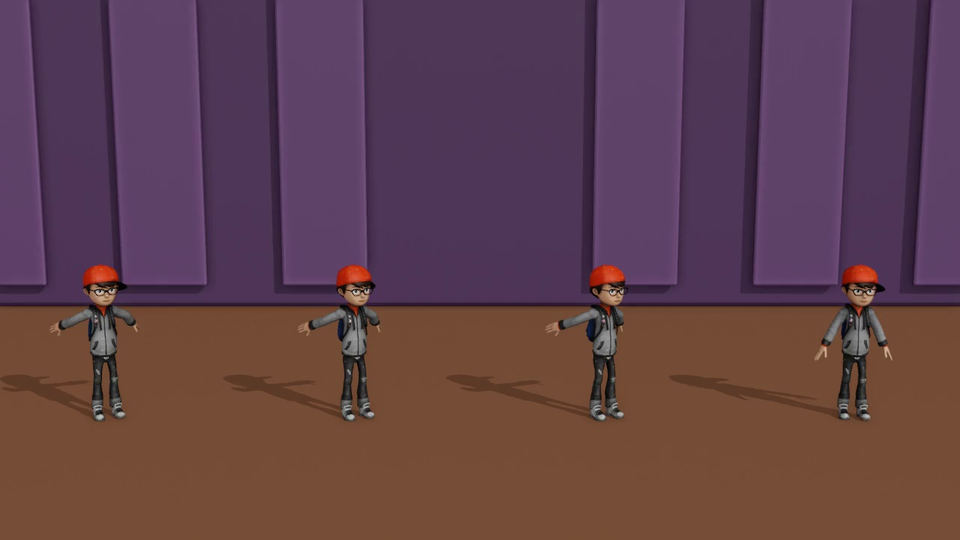}
{\textbf{ASK} 99s}
\hfill
\frameitem{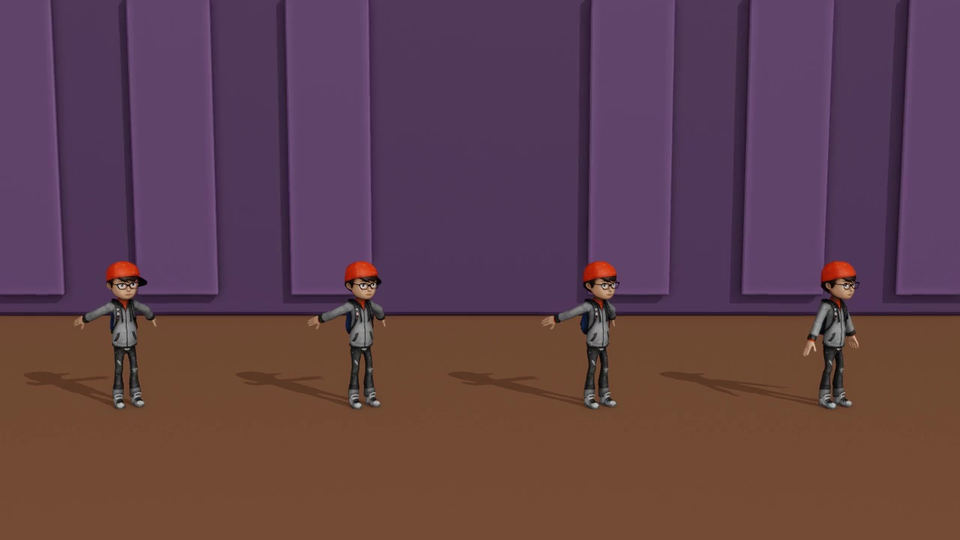}
{\textbf{ANSWER} 125s}
\hfill
\frameitem{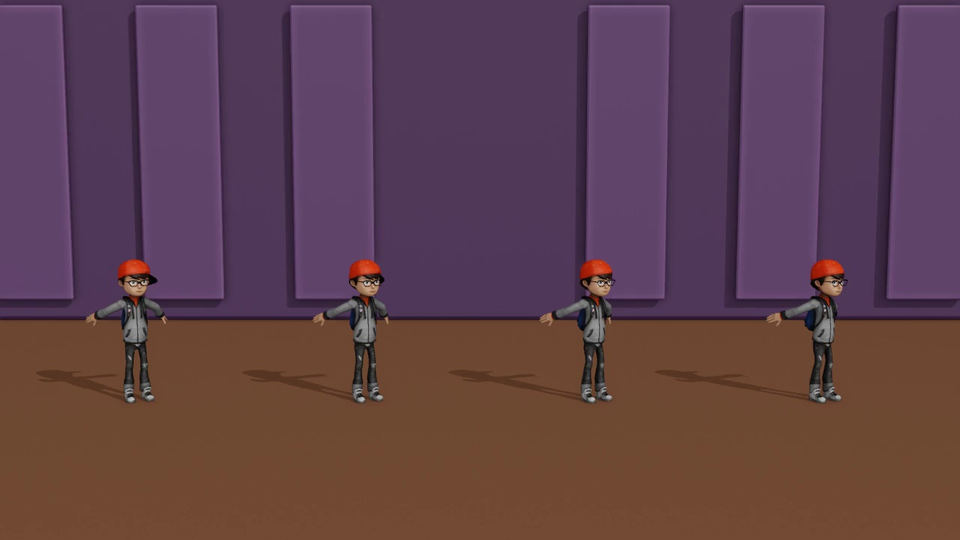}
{\textbf{ANSWER} 150s}

%
%

\noindent
\frameitem{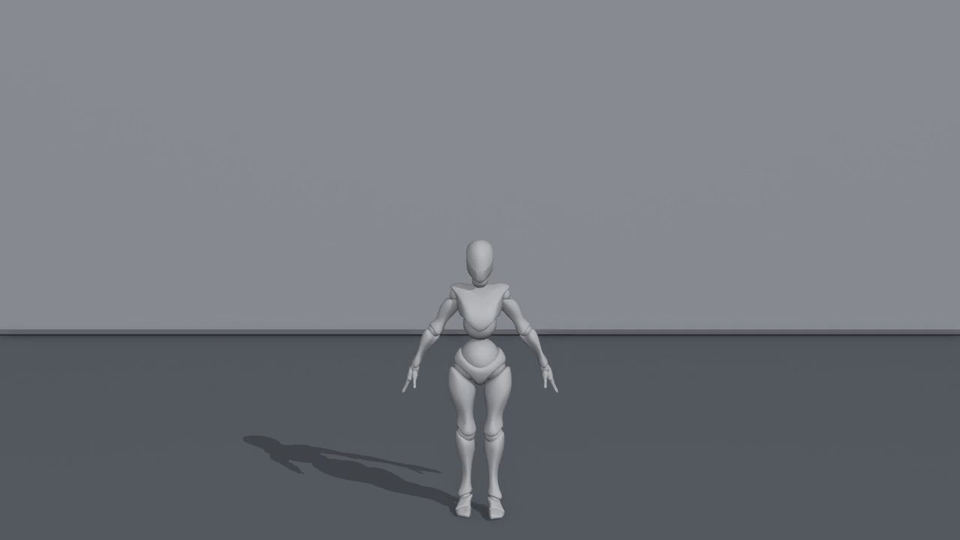}
{0s}
\hfill
\frameitem{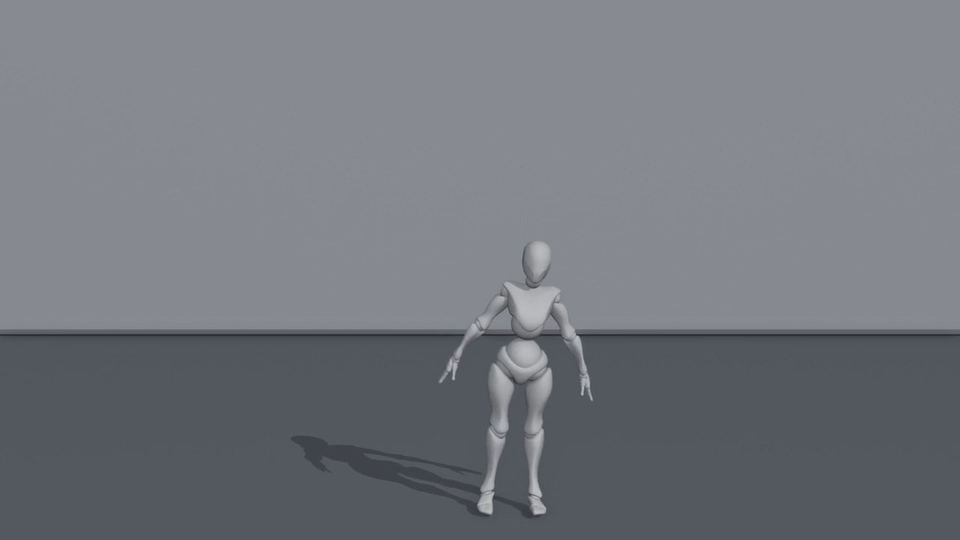}
{6s}
\hfill
\frameitem{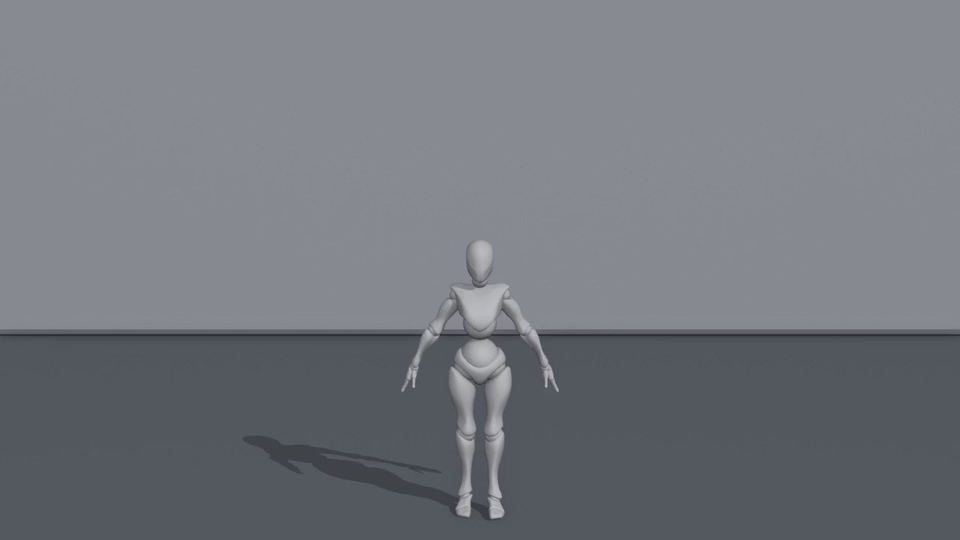}
{11s}
\hfill
\frameitem{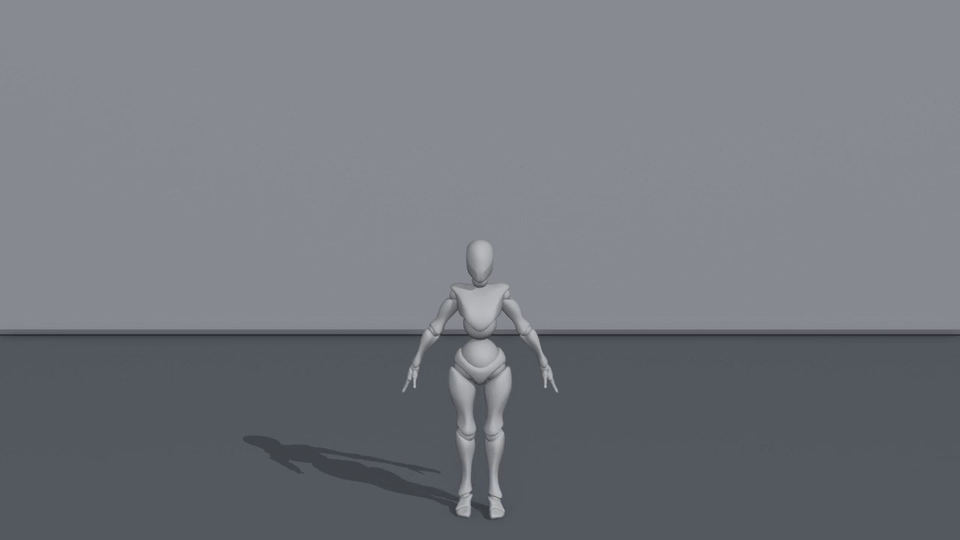}
{17s}
\hfill
\frameitem{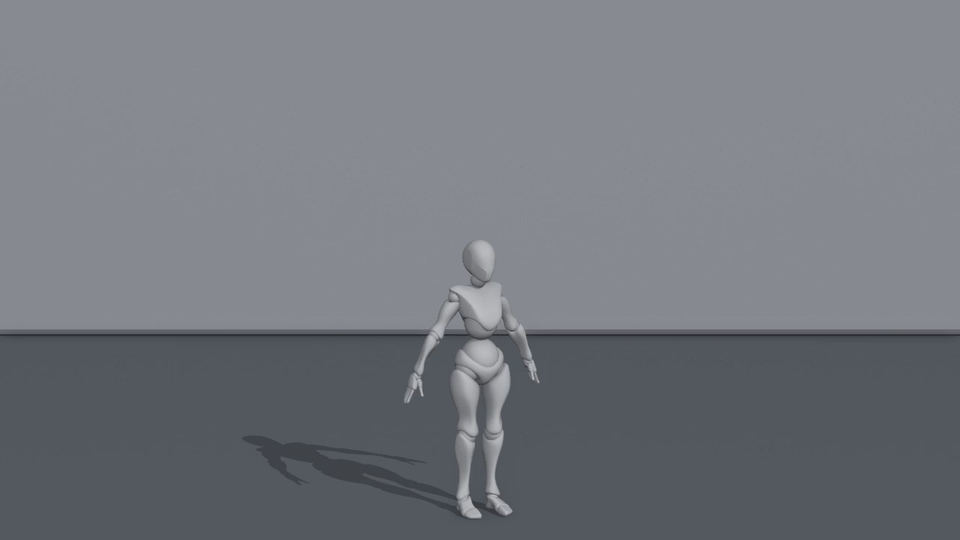}
{23s}
\hfill
\frameitem{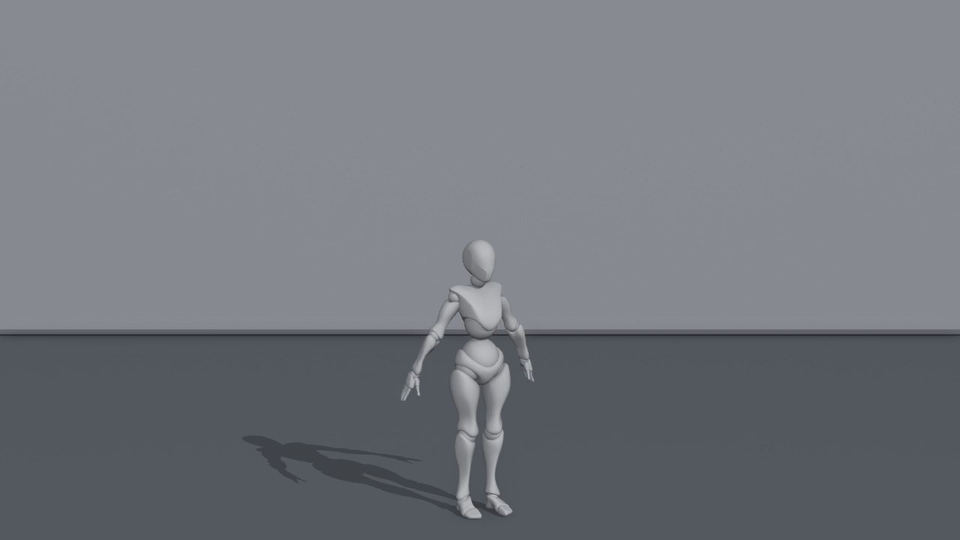}
{28s}
\hfill
\frameitem{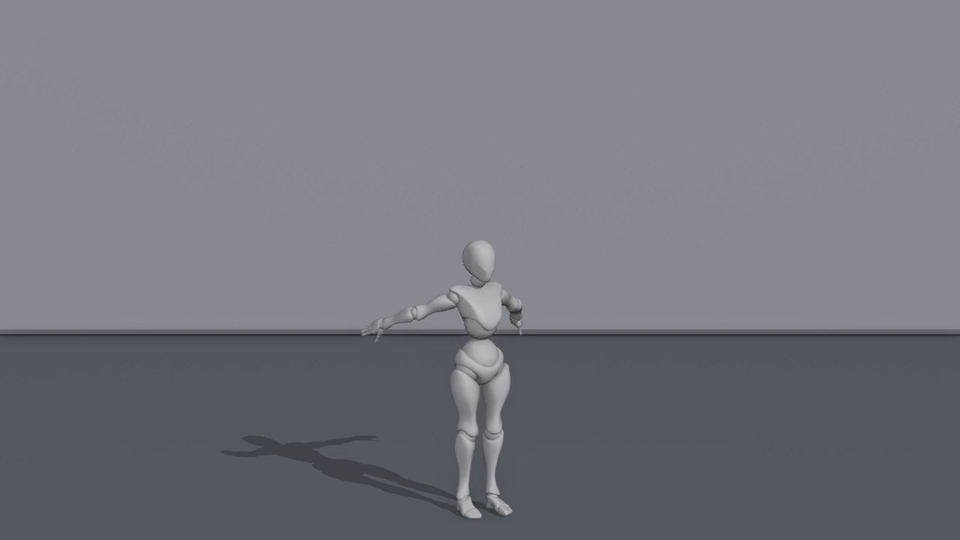}
{34s}


\noindent
\textbf{Question:}
When you can determine which character in the video performs the
personalized action ``Vexalia'', tell me who it is.


\noindent
\textbf{Action Options:}

\begin{tabularx}{\linewidth}{
    @{}
    >{\raggedright\arraybackslash}p{0.035\linewidth}
    >{\raggedright\arraybackslash}X
    @{}
}

A. &
Does ``Vexalia'' require circling both wrists before a left knee lift,
a left knee lift before a low left-arm sweep, and turning both palms
over before opening both arms at mid height?
\\[2pt]

B. &
Does ``Vexalia'' require a right foot tap before a hip shift to the left,
a low left-arm sweep before folding both elbows, and opening both arms
at mid height before opening both hands?
\\[2pt]

\textbf{C.} &
\textbf{
Does ``Vexalia'' require a left side step before a right side step,
a turn to the left before bringing the right foot back,
and a crossed-wrist finish before a final hold?
}
\\[2pt]

D. &
Does ``Vexalia'' require a hip shift to the left before a high arc made
with both arms, a high arc made with both arms before a torso lean to
the right, and a torso lean to the right before circling both wrists?
\\

\end{tabularx}


\noindent
\textbf{Answer Options:}

\begin{tabularx}{\linewidth}{
    @{}
    >{\raggedright\arraybackslash}p{0.035\linewidth}
    >{\raggedright\arraybackslash}X
    @{}
}

\textbf{A.} &
\textbf{The person who performs the full routine third.}
\\[2pt]

B. &
The person who performs the full routine first.
\\[2pt]

C. &
The person who performs the full routine second.
\\[2pt]

D. &
The person who performs the full routine fourth.
\\

\end{tabularx}
\end{videoexamplebox}
\begin{videoexamplebox}

%
%
%

\noindent
\frameitem{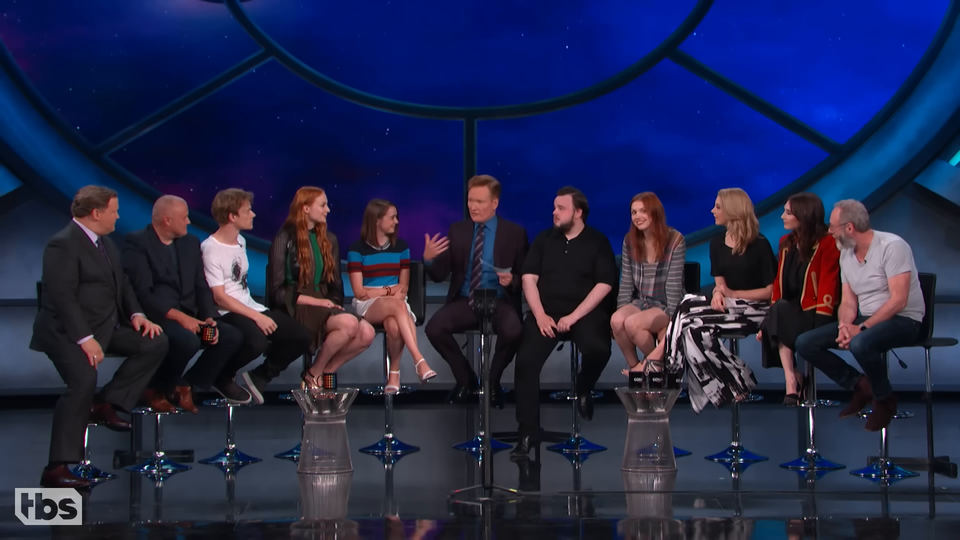}
{\textbf{ASK} 13s}
\hfill
\frameitem{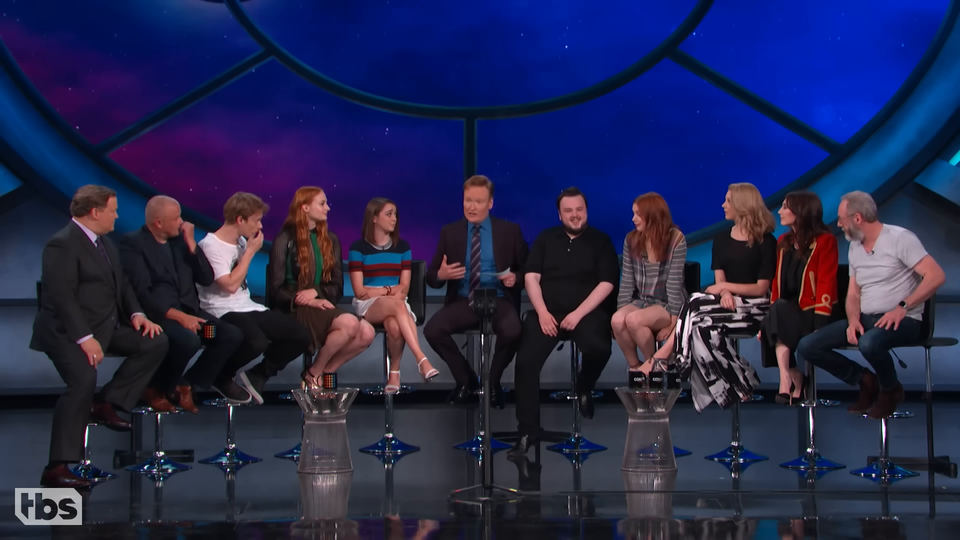}
{\textbf{ASK} 30s}
\hfill
\frameitem{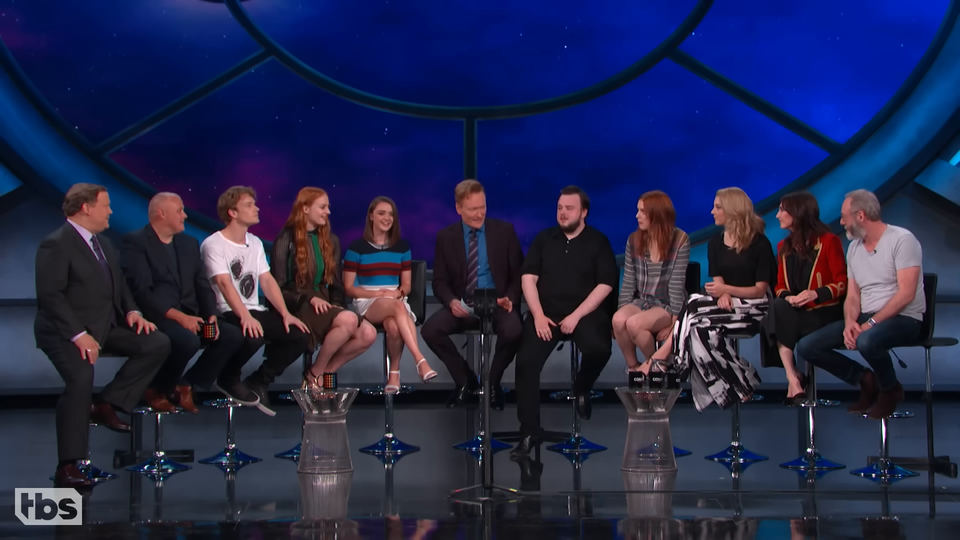}
{\textbf{ASK} 47s}
\hfill
\frameitem{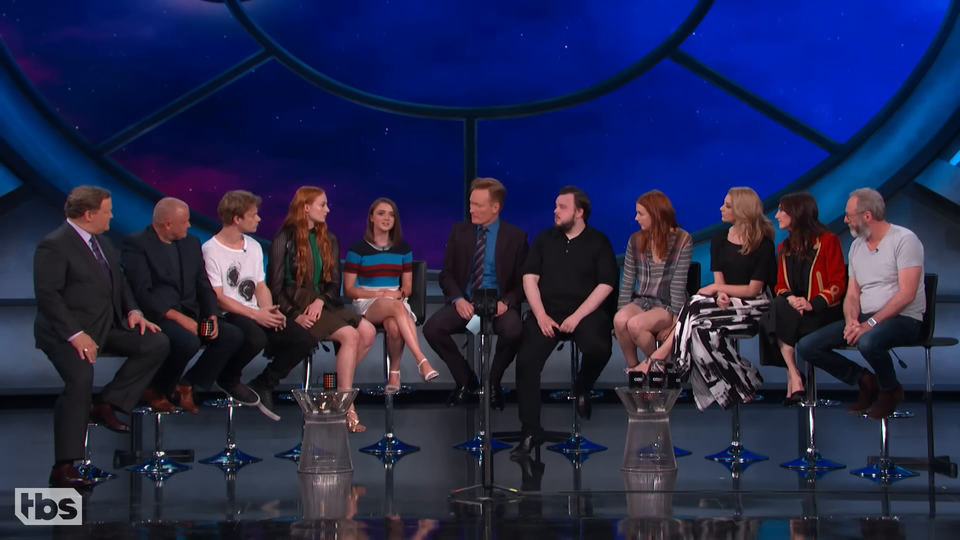}
{\textbf{ASK} 64s}
\hfill
\frameitem{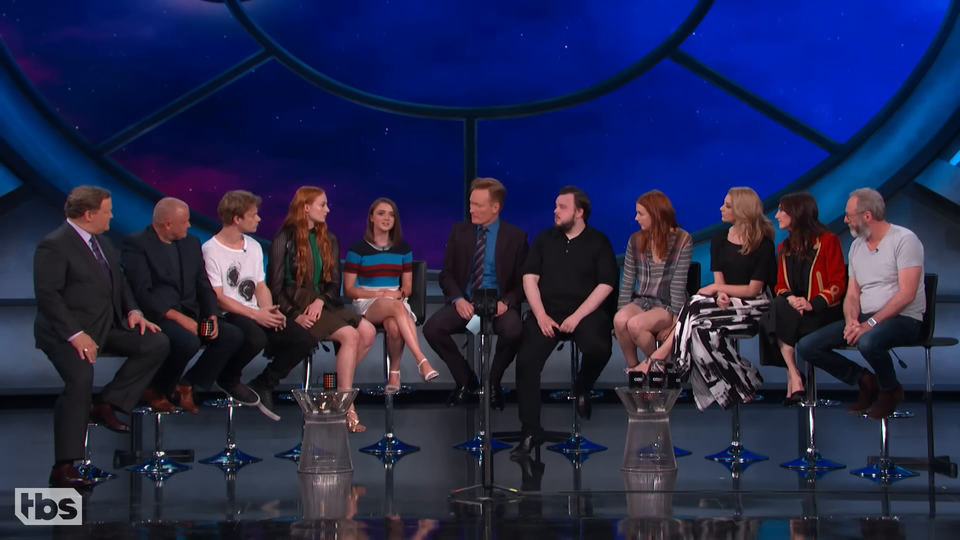}
{\textbf{ANSWER} 64s}
\hfill
\frameitem{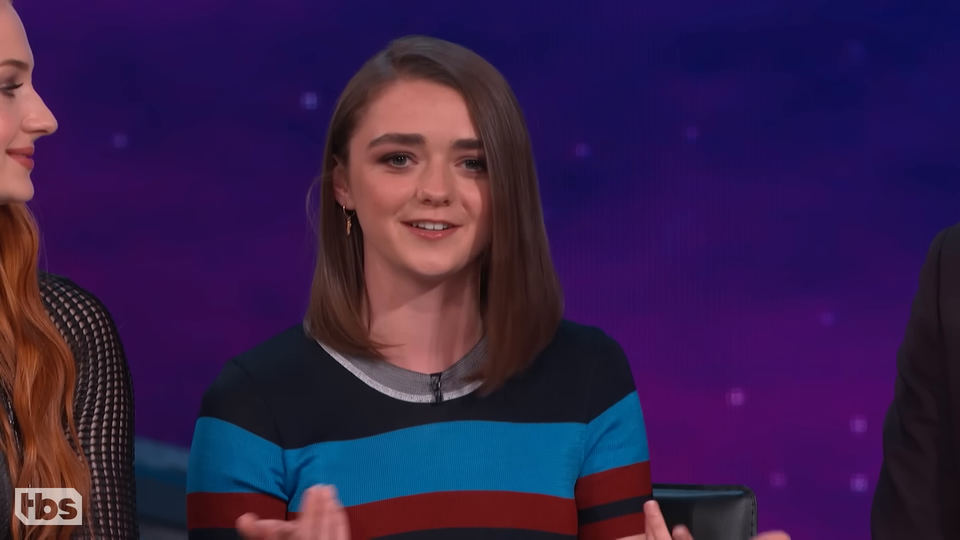}
{\textbf{ANSWER} 77s}
\hfill
\frameitem{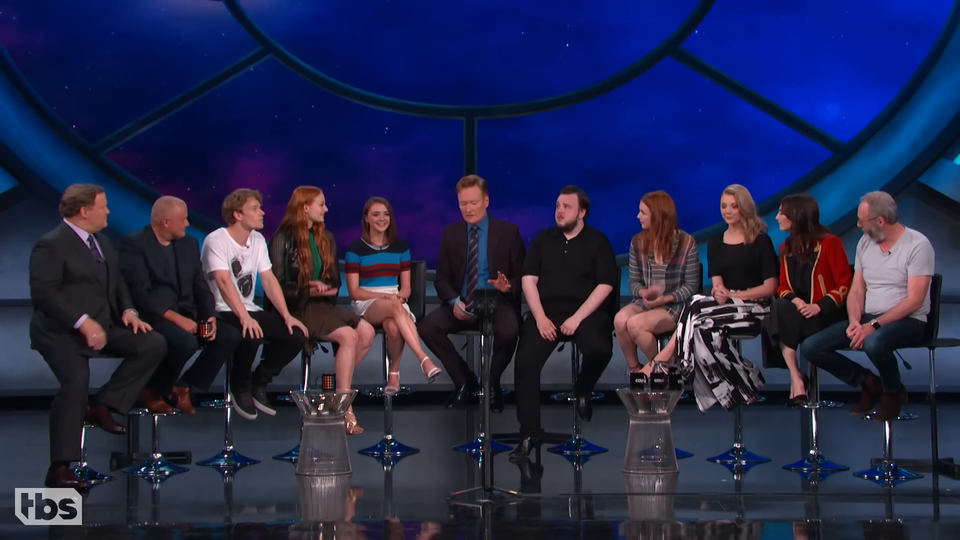}
{\textbf{ANSWER} 90s}


\noindent
\textbf{Question:}
When the woman wearing a skirt begins expressing her opinion, tell me,
``She has started speaking.''


\noindent
\textbf{Action Options:}

\begin{tabularx}{\linewidth}{
    @{}
    >{\raggedright\arraybackslash}p{0.035\linewidth}
    >{\raggedright\arraybackslash}X
    @{}
}

A. &
I need to know whether the skirt-wearing woman wears a jacket.
\\[2pt]

B. &
I need to know which footwear style the skirt-wearing woman has.
\\[2pt]

C. &
I need to know which direction the skirt-wearing woman faces.
\\[2pt]

D. &
I need to know what sleeve length the skirt-wearing woman has.
\\[2pt]

\textbf{E.} &
\textbf{I need to know which top pattern distinguishes the skirt-wearing woman.}
\\[2pt]

F. &
I need to know whether the skirt-wearing woman crosses her legs.
\\

\end{tabularx}


\noindent
\textbf{User Reply:}

\noindent
\textit{She is wearing a blue-and-red striped top.}


\noindent
\textbf{Answer:}

\noindent
\textbf{She has started speaking.}

\end{videoexamplebox}

\begin{videoexamplebox}

%
%
%

\noindent
\frameitem{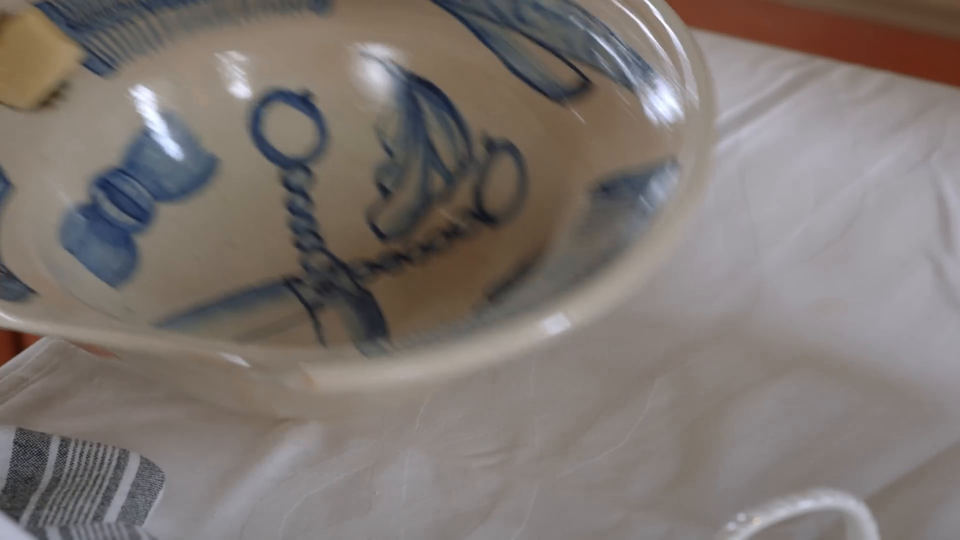}
{\textbf{WAIT} 6s}
\hfill
\frameitem{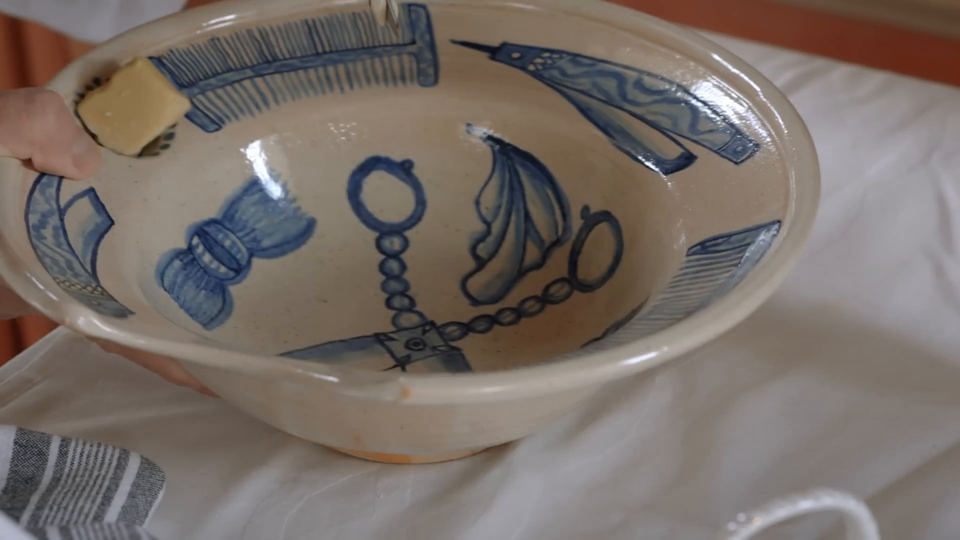}
{\textbf{WAIT} 7s}
\hfill
\frameitem{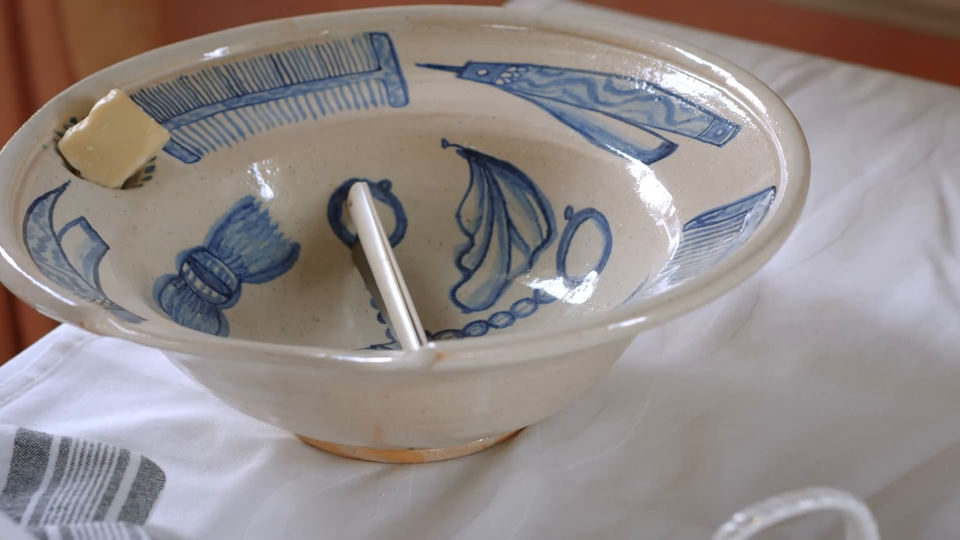}
{\textbf{WAIT} 9s}
\hfill
\frameitem{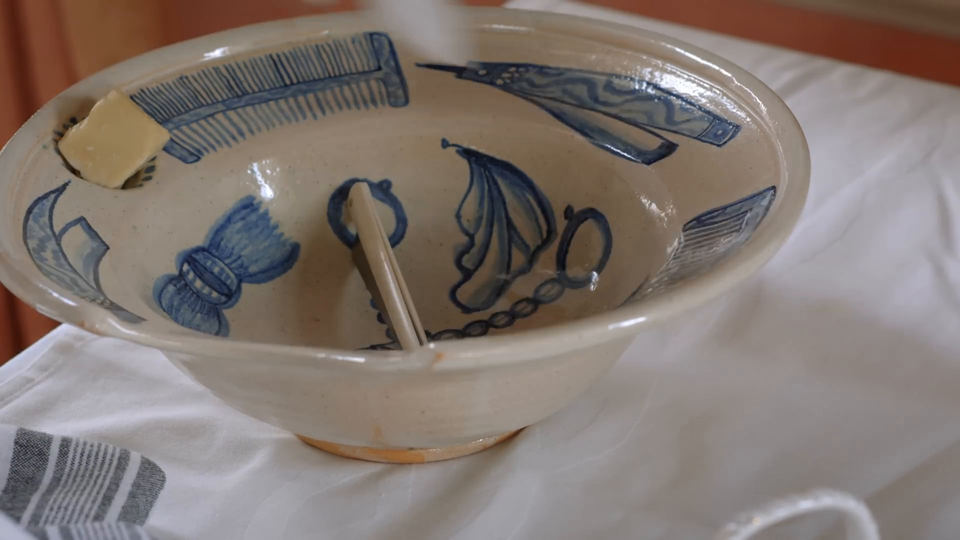}
{\textbf{WAIT} 10s}
\hfill
\frameitem{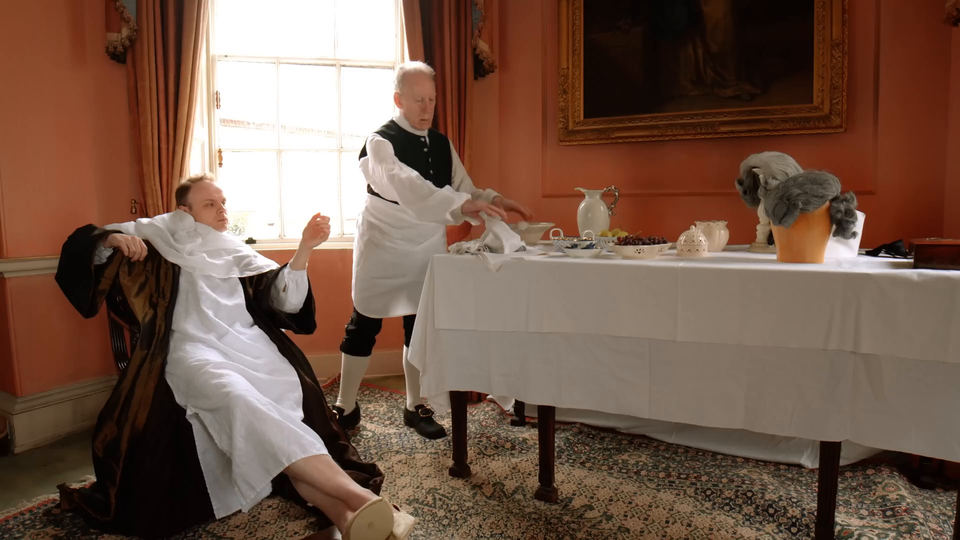}
{\textbf{ANSWER} 10s}
\hfill
\frameitem{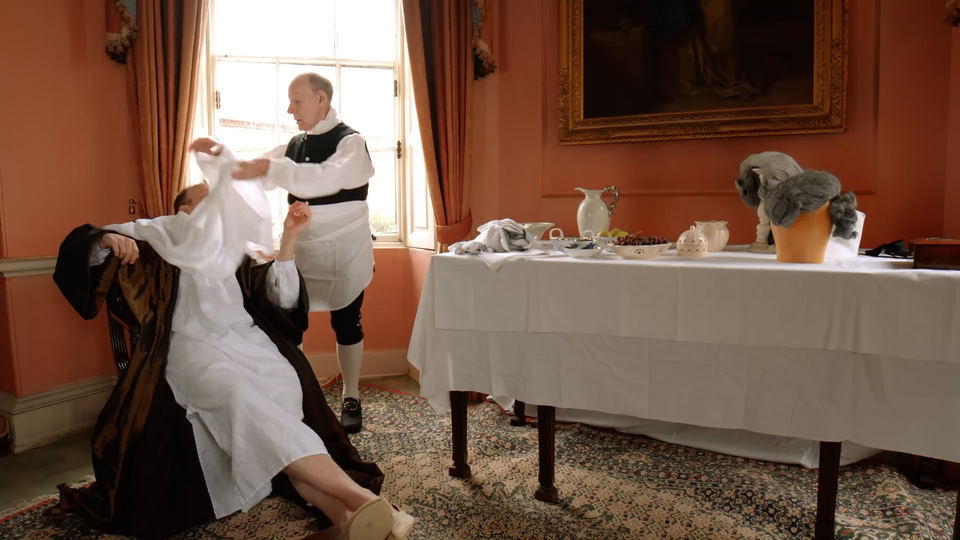}
{\textbf{ANSWER} 19s}
\hfill
\frameitem{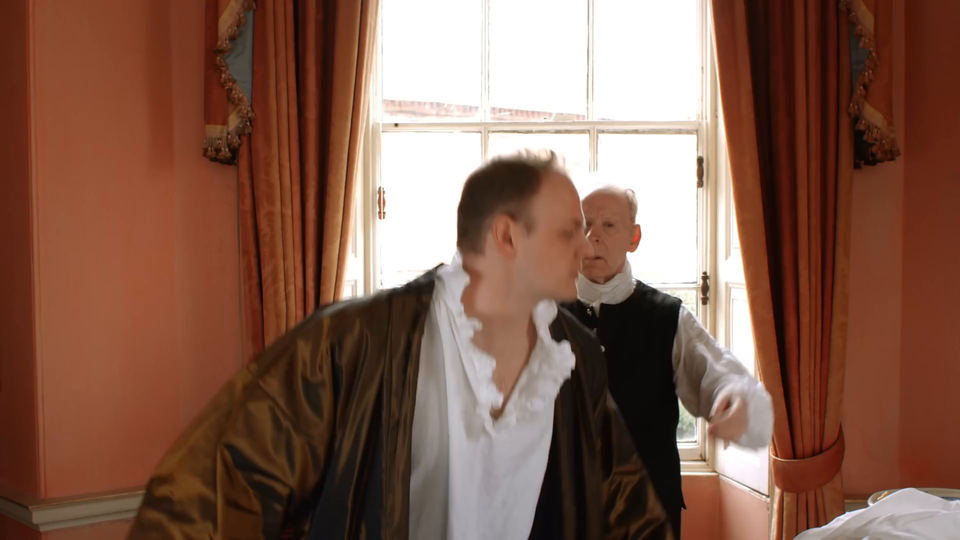}
{\textbf{ANSWER} 28s}


\noindent
\textbf{Question:}
When the walls in the larger room, with a large round clock hanging
above the long table, are visible, what color are they?


\noindent
\textbf{Action Options:}

\begin{tabularx}{\linewidth}{
    @{}
    >{\raggedright\arraybackslash}p{0.035\linewidth}
    >{\raggedright\arraybackslash}X
    @{}
}

A. &
Actually, a wide wooden panel hangs above the long table in the larger room.
\\[2pt]

\textbf{B.} &
\textbf{Actually, a large framed painting hangs above the long table in the larger room.}
\\[2pt]

C. &
Actually, a row of shelves hangs above the long table in the larger room.
\\[2pt]

D. &
Actually, a large framed mirror hangs above the long table in the larger room.
\\[2pt]

E. &
Actually, a large round clock hangs above the long table in the larger room.
\\[2pt]

F. &
Actually, a large wall tapestry hangs above the long table in the larger room.
\\

\end{tabularx}


\noindent
\textbf{Answer Options:}

\begin{tabularx}{\linewidth}{
    @{}
    >{\raggedright\arraybackslash}p{0.035\linewidth}
    >{\raggedright\arraybackslash}X
    @{}
}

A. &
White.
\\[2pt]

\textbf{B.} &
\textbf{Peach.}
\\[2pt]

C. &
Blue.
\\[2pt]

D. &
Gray.
\\

\end{tabularx}

\end{videoexamplebox}

\begin{videoexamplebox}

%
%
%

\noindent
\frameitem{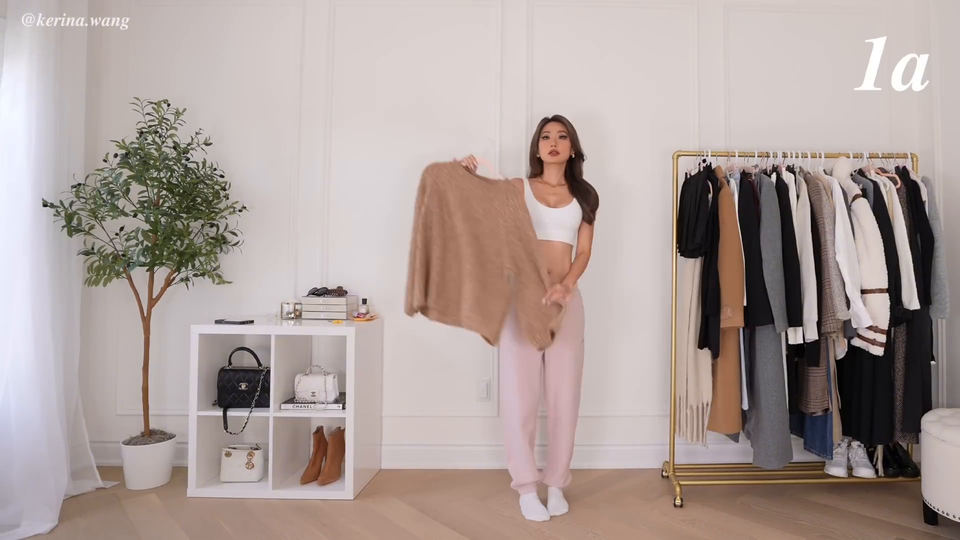}
{\textbf{ASK} 0s}
\hfill
\frameitem{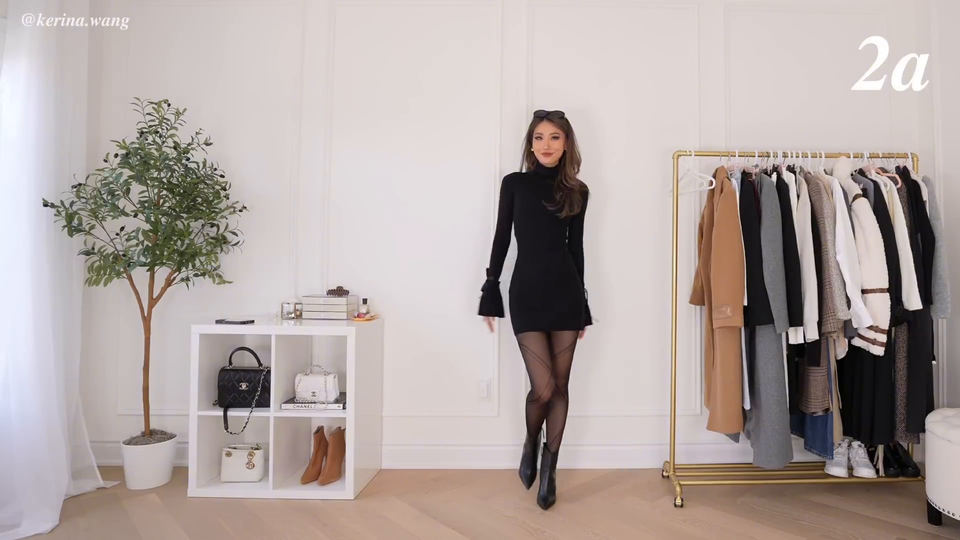}
{\textbf{ASK} 70s}
\hfill
\frameitem{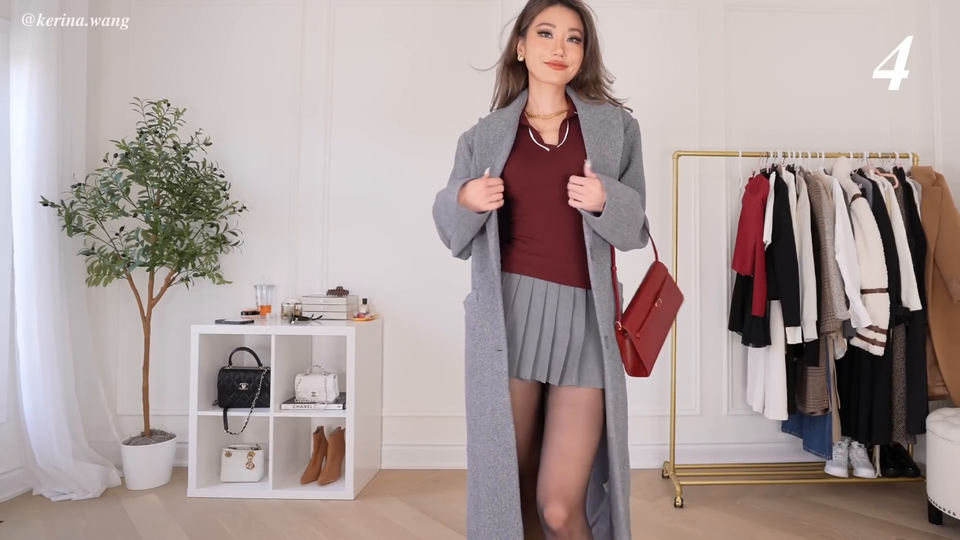}
{\textbf{ASK} 141s}
\hfill
\frameitem{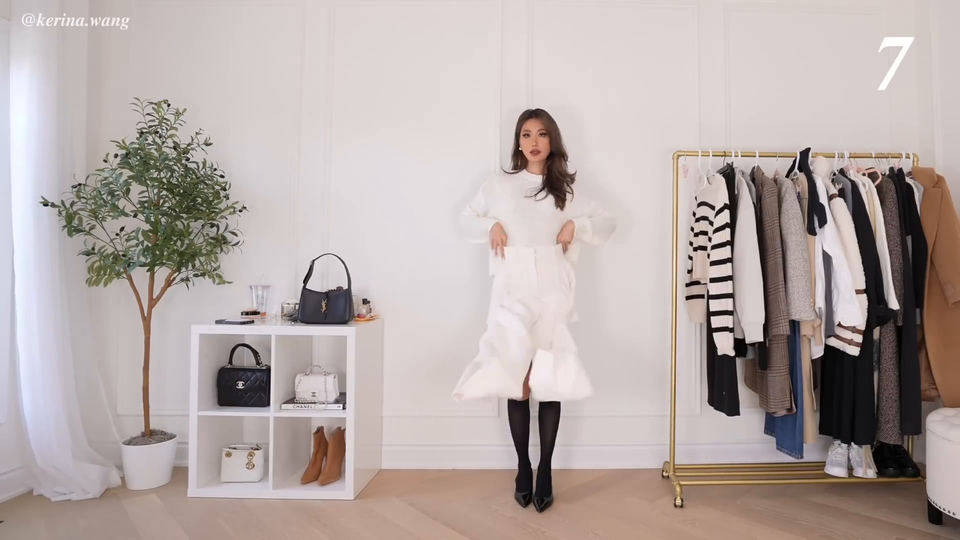}
{\textbf{ASK} 211s}
\hfill
\frameitem{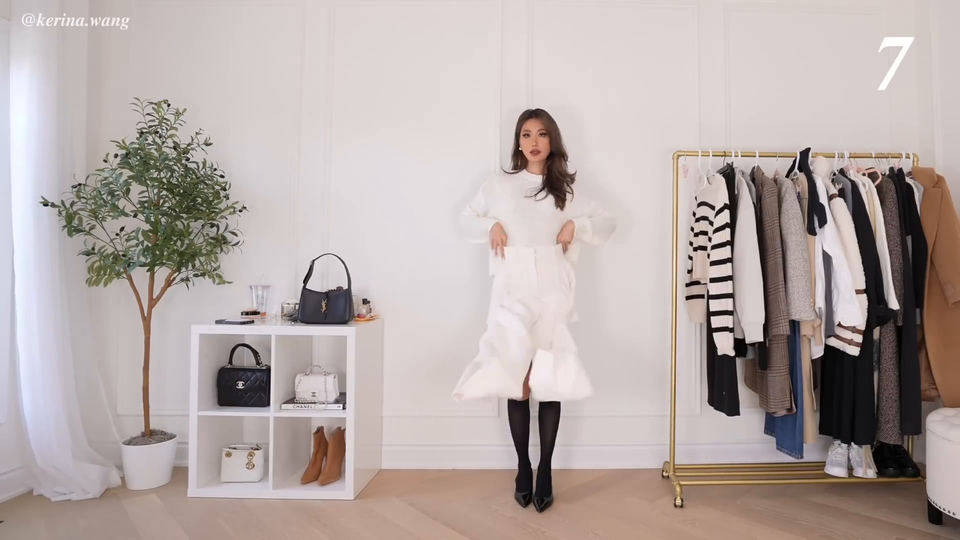}
{\textbf{ANSWER} 211s}
\hfill
\frameitem{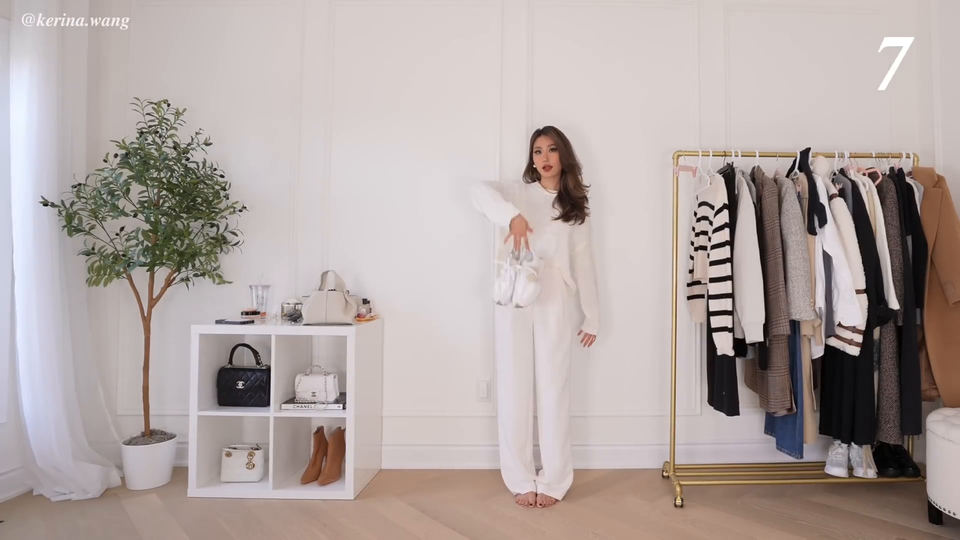}
{\textbf{ANSWER} 213s}
\hfill
\frameitem{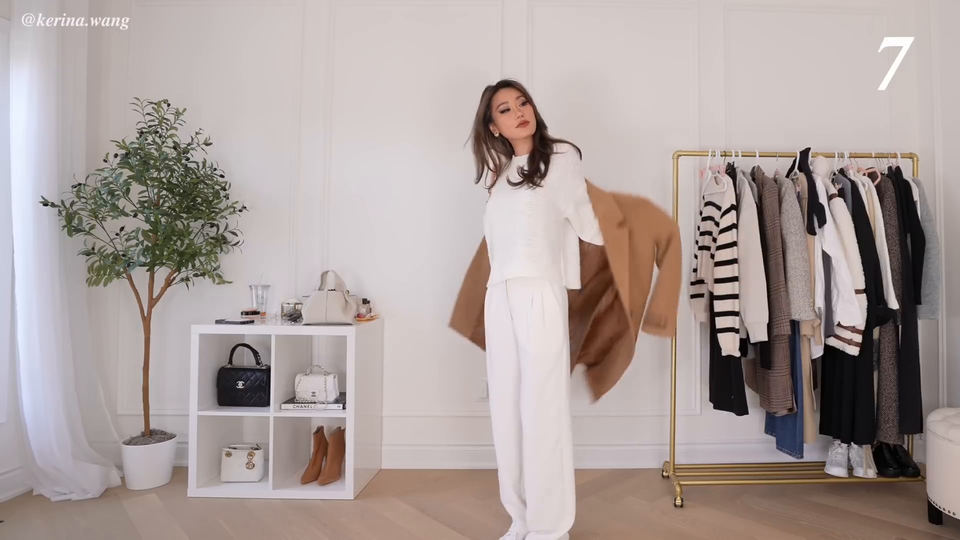}
{\textbf{ANSWER} 215s}


\noindent
\frameitem{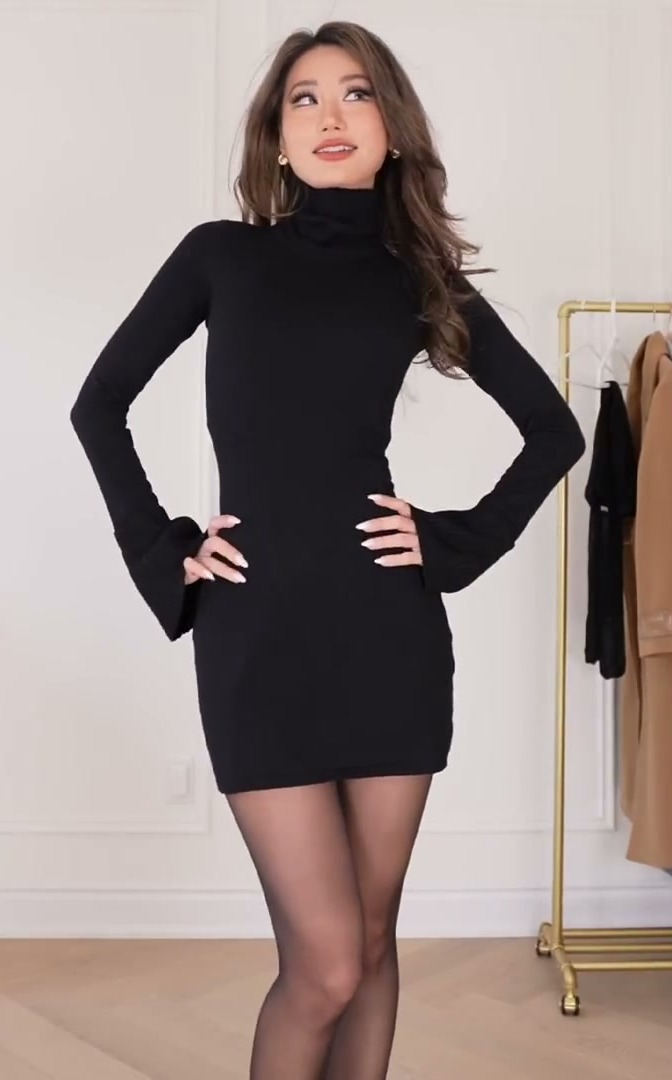}
{Entity 1}
\hfill
\frameitem{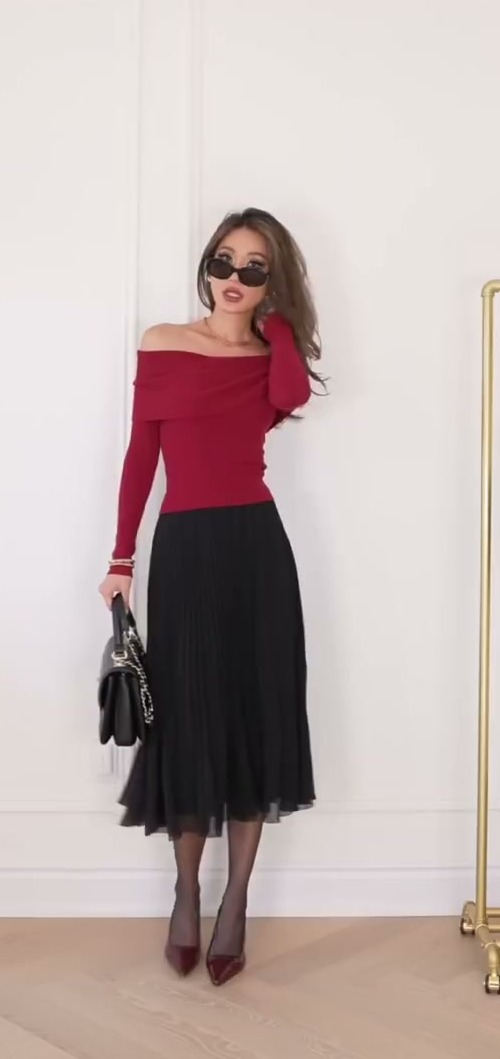}
{Entity 2}
\hfill
\frameitem{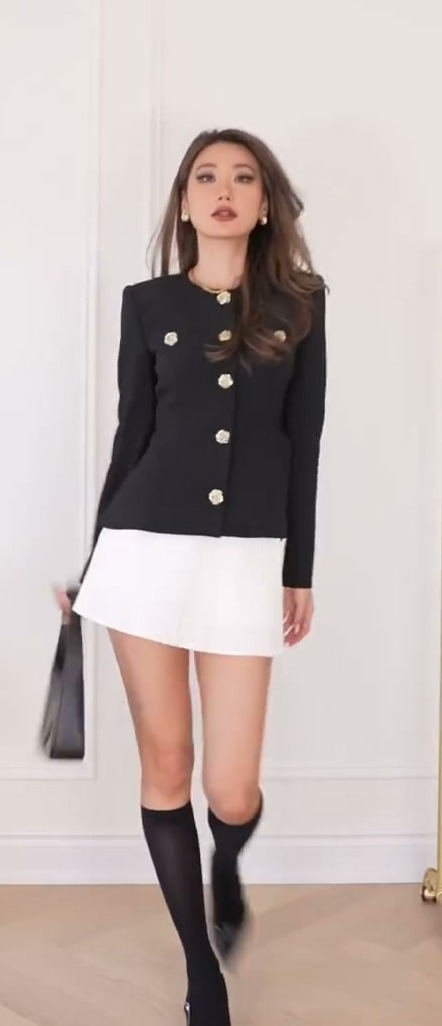}
{Entity 3}
\hfill
\frameitem{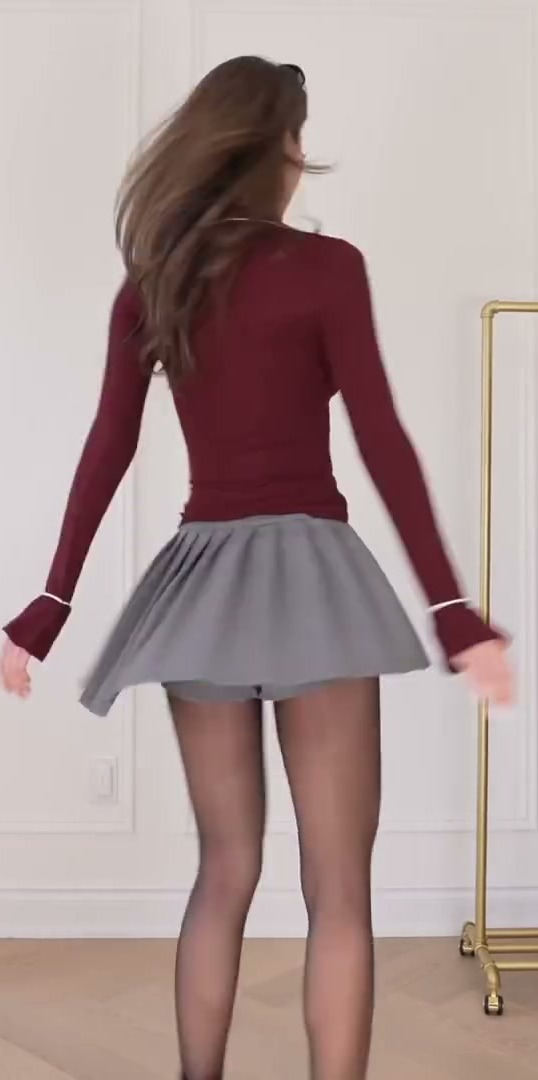}
{Entity 4}
\hfill
\frameitem{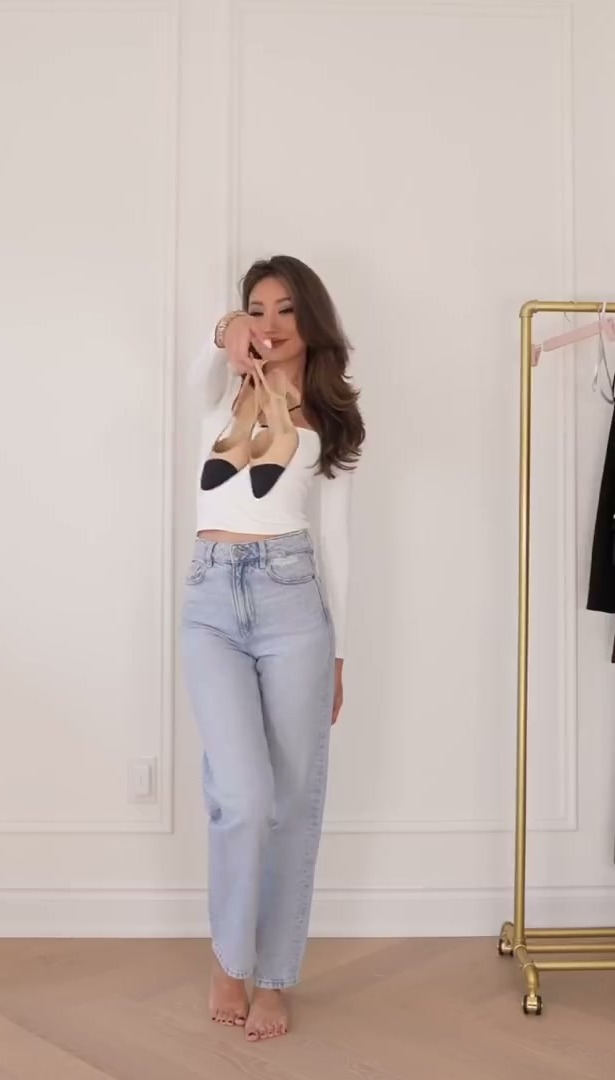}
{Entity 5}
\hfill
\frameitem{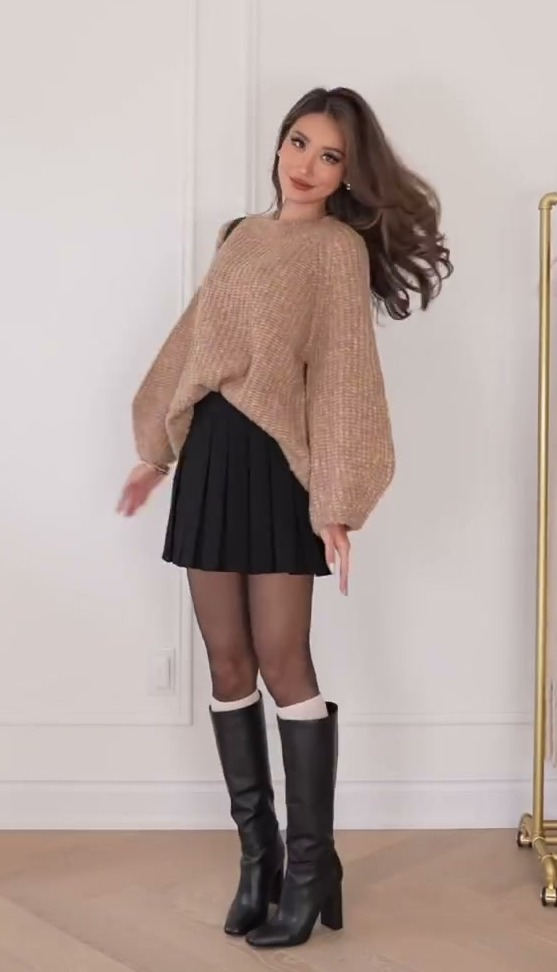}
{Entity 6}
\hfill
\frameitem{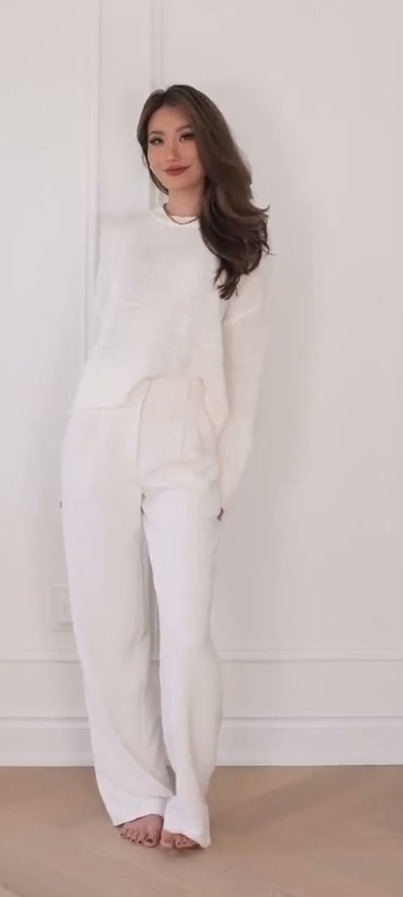}
{Entity 7}


\noindent
\textbf{Question:}
When the person first appears barefoot in a white knit top with white
wide-leg trousers and no jacket, which of all the outfits shown so far
suits you best?


\noindent
\textbf{Action Options:}

\begin{tabularx}{\linewidth}{
    @{}
    >{\raggedright\arraybackslash}p{0.035\linewidth}
    >{\raggedright\arraybackslash}X
    @{}
}

A. &
Would you prefer a skirted outfit or one with trousers?
\\[2pt]

B. &
Do you prefer long sleeves, short sleeves, or sleeveless styling?
\\[2pt]

C. &
Would you like sheer dark tights as part of your outfit?
\\[2pt]

D. &
Which top color family do you prefer: neutral or red-toned?
\\[2pt]

\textbf{E.} &
\textbf{Which skirt color do you prefer: black, gray, or white?} ...
\\[2pt]

\end{tabularx}


\noindent
\textbf{User Preference:}
I prefer a gray skirt.

\noindent
\textbf{Answer Options:}

\begin{tabularx}{\linewidth}{
    @{}
    >{\raggedright\arraybackslash}p{0.035\linewidth}
    >{\raggedright\arraybackslash}X
    @{}
}

A. &
Beige textured sweater, black pleated miniskirt, tights, and knee-high boots.
\\[2pt]

B. &
Black turtleneck minidress with flared cuffs and sheer tights.
\\[2pt]

C. &
White fitted top with black bust accents and light-wash jeans.
\\[2pt]

\textbf{D.} &
\textbf{Burgundy long-sleeve top with flared cuffs and gray flared miniskirt.}...
\\[2pt]

\end{tabularx}

\end{videoexamplebox}

\begin{videoexamplebox}

%
%
%

\noindent
\frameitem{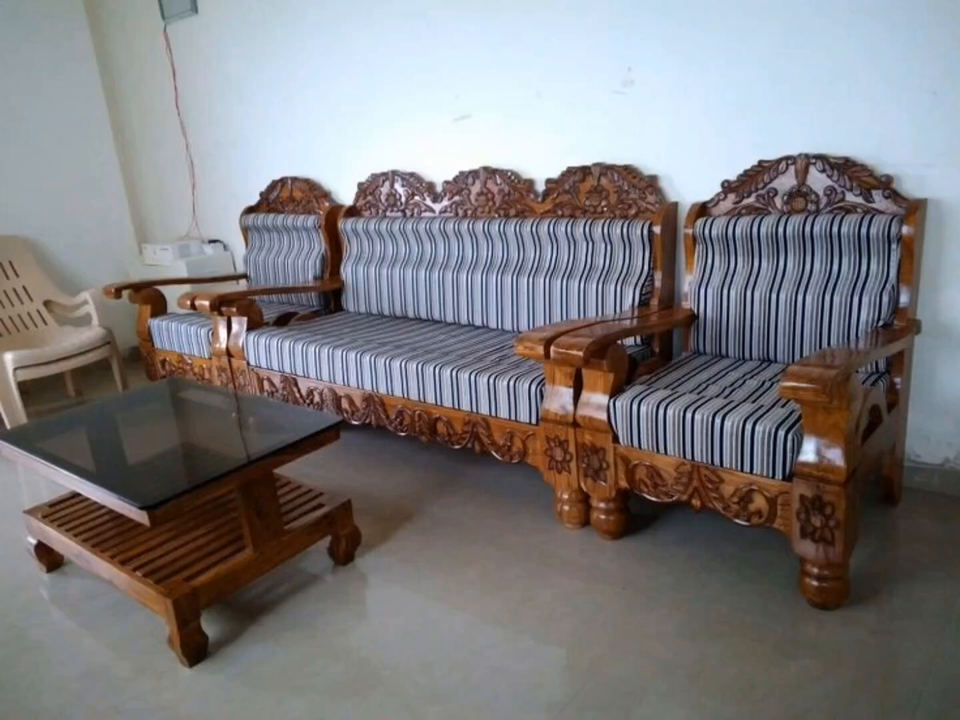}
{\textbf{ASK} 0s}
\hfill
\frameitem{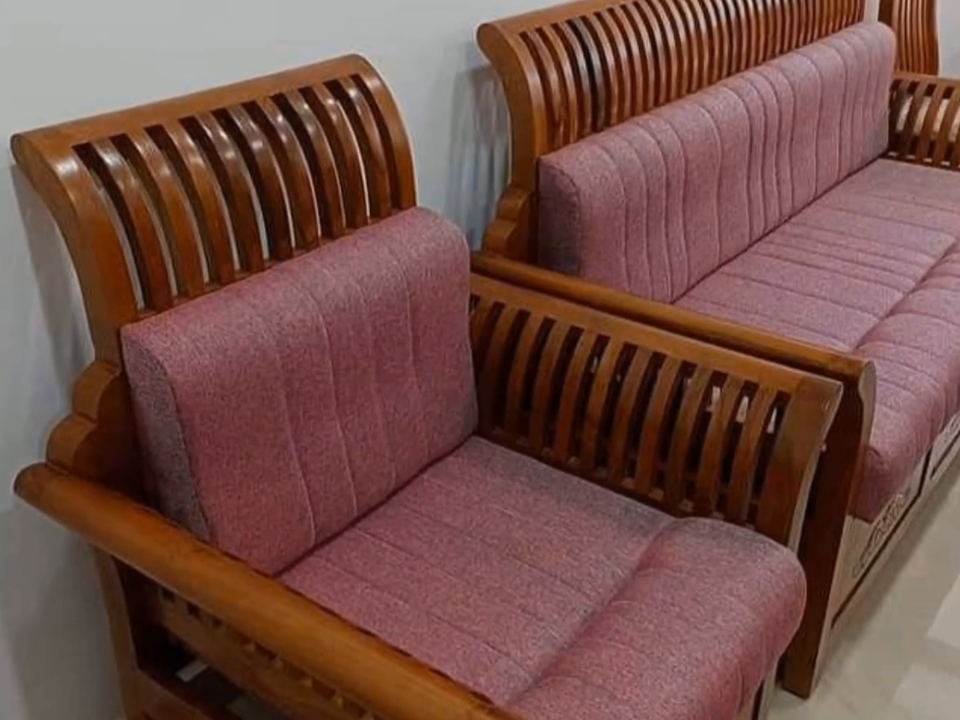}
{\textbf{ASK} 17s}
\hfill
\frameitem{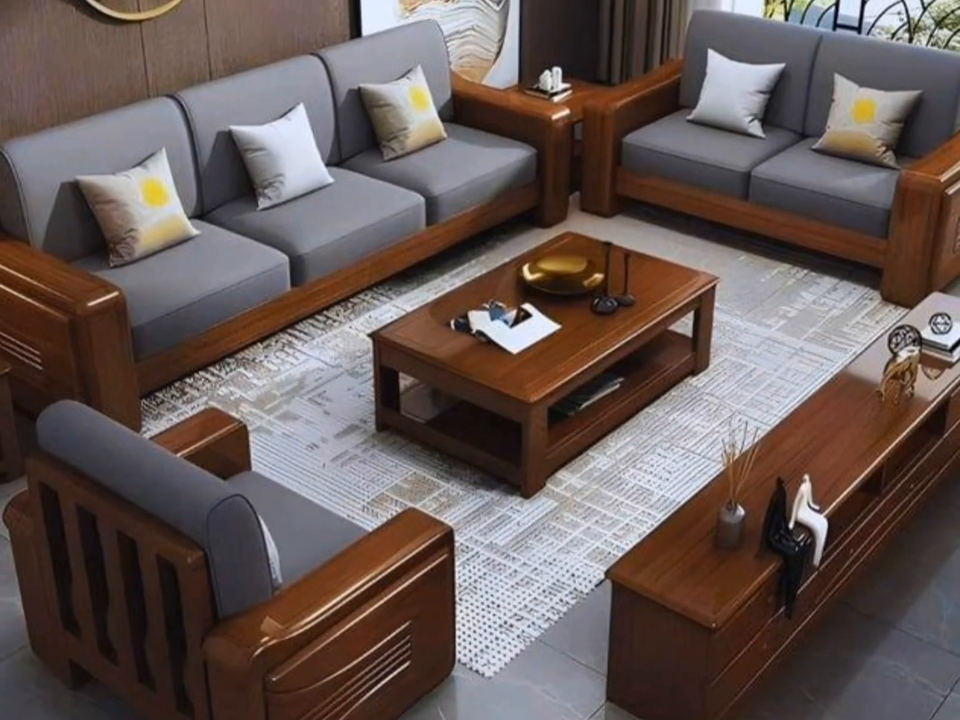}
{\textbf{ASK} 34s}
\hfill
\frameitem{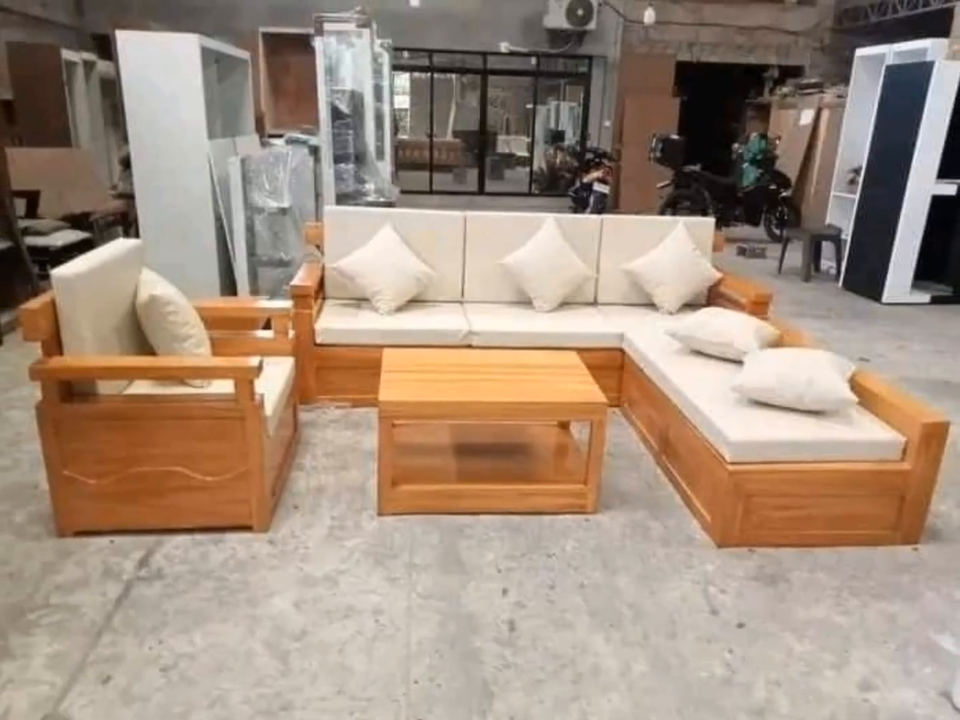}
{\textbf{ASK} 52s}
\hfill
\frameitem{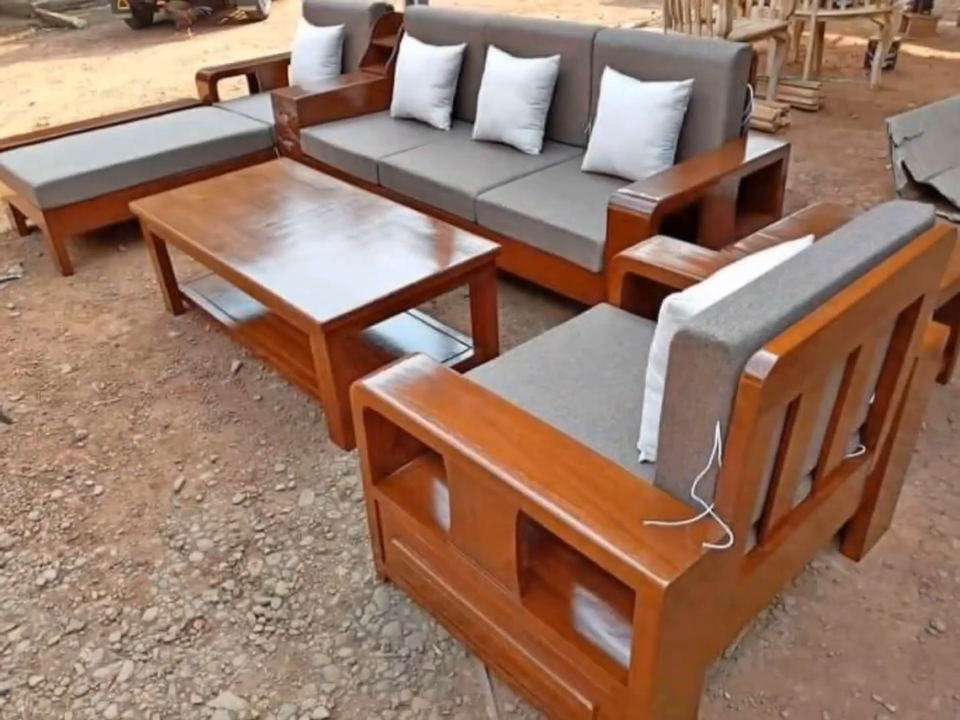}
{\textbf{ANSWER} 53s}
\hfill
\frameitem{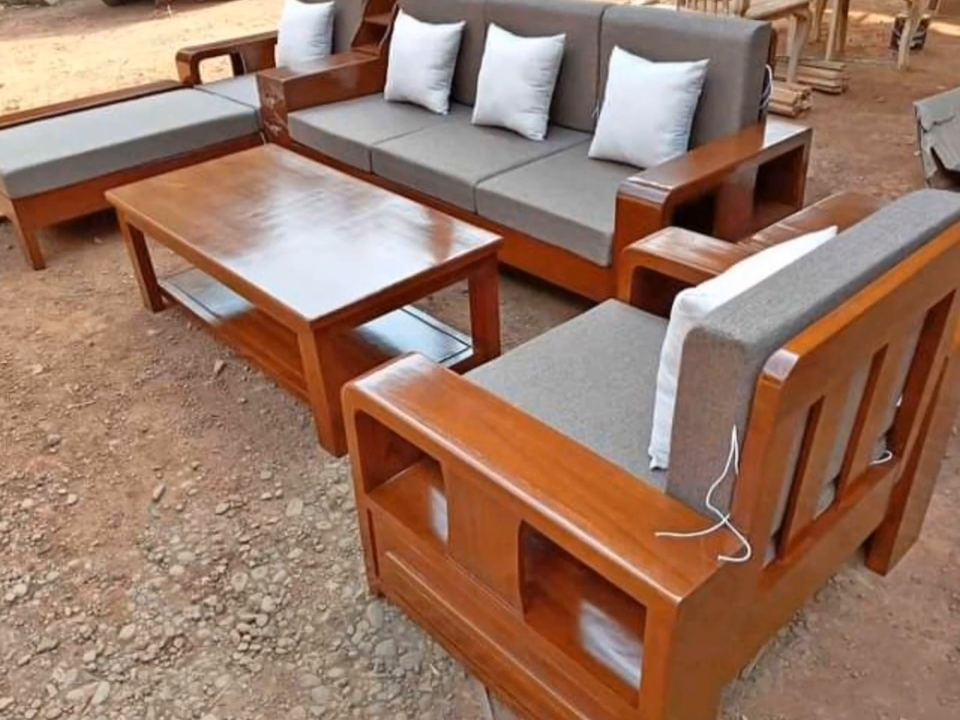}
{\textbf{ANSWER} 55s}
\hfill
\frameitem{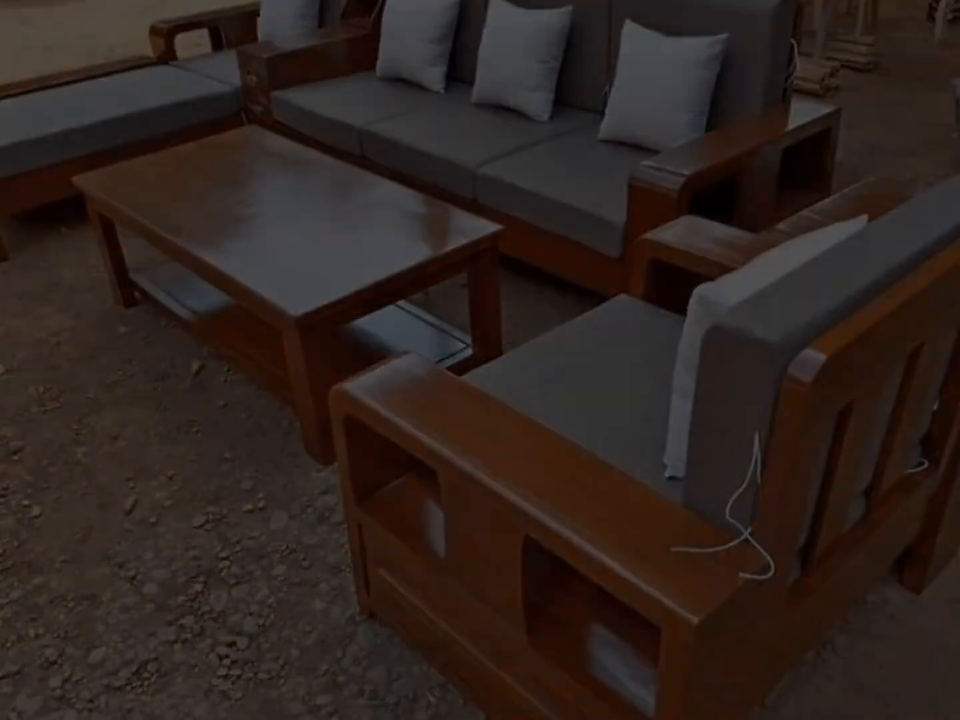}
{\textbf{ANSWER} 58s}


\noindent
\frameitem{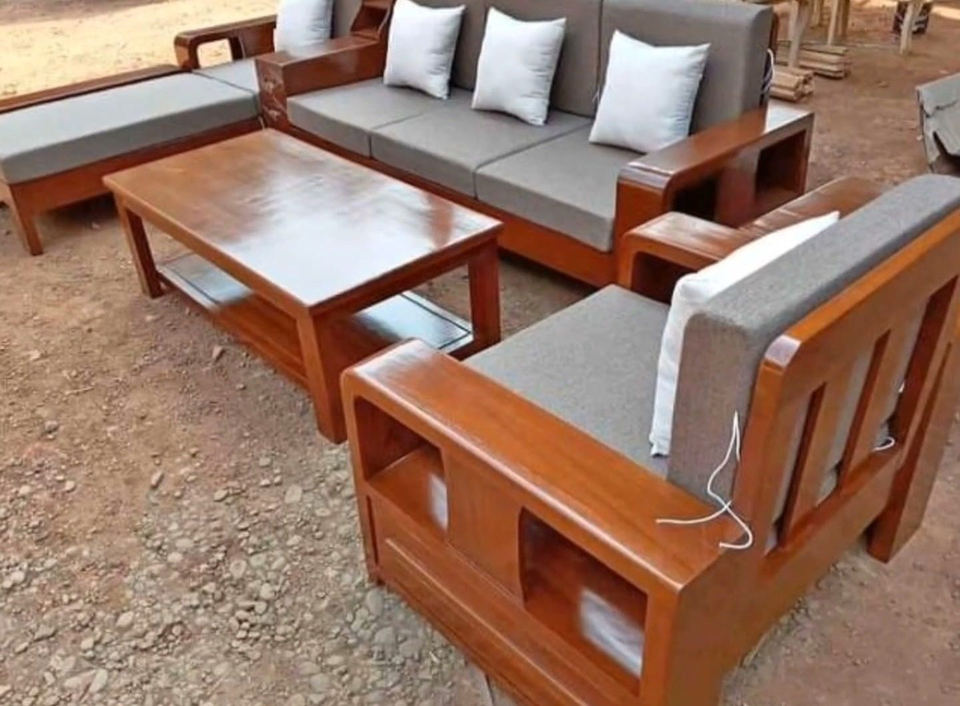}
{Entity 1}
\hfill
\frameitem{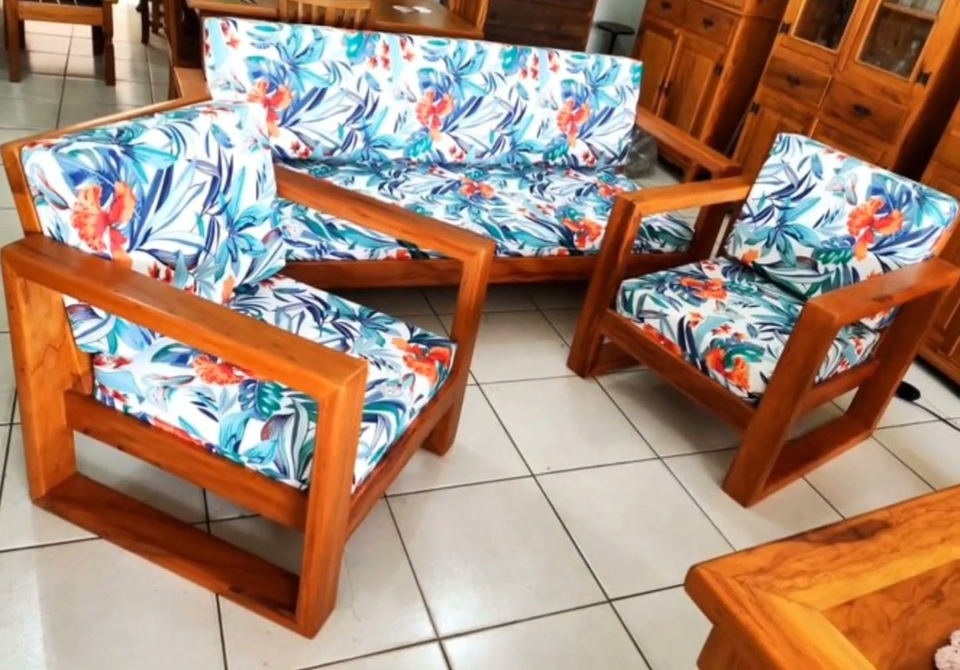}
{Entity 2}
\hfill
\frameitem{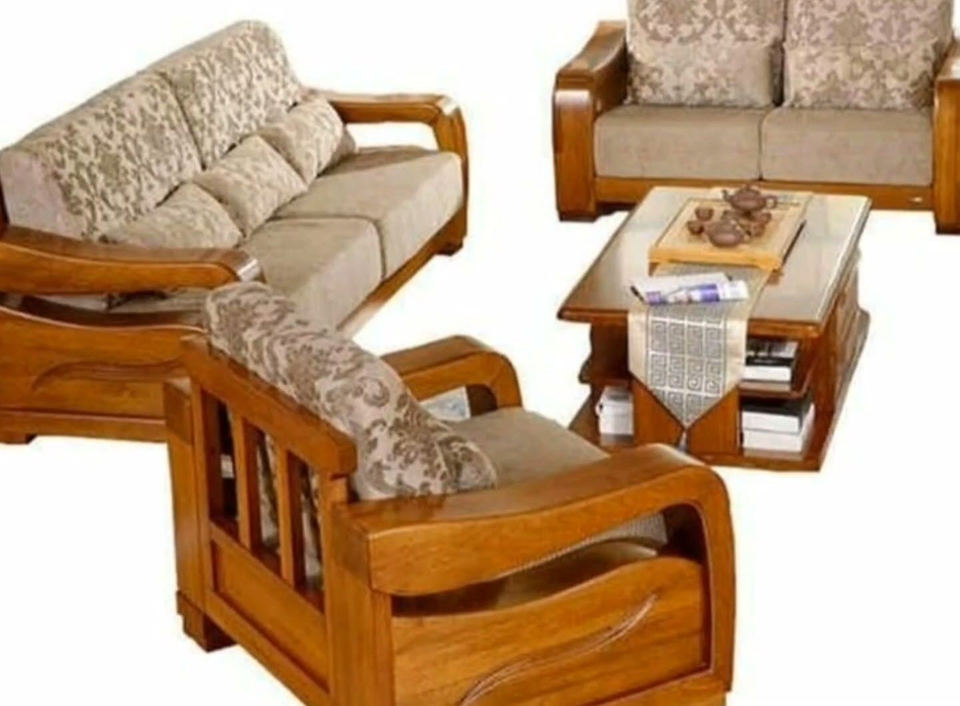}
{Entity 3}
\hfill
\frameitem{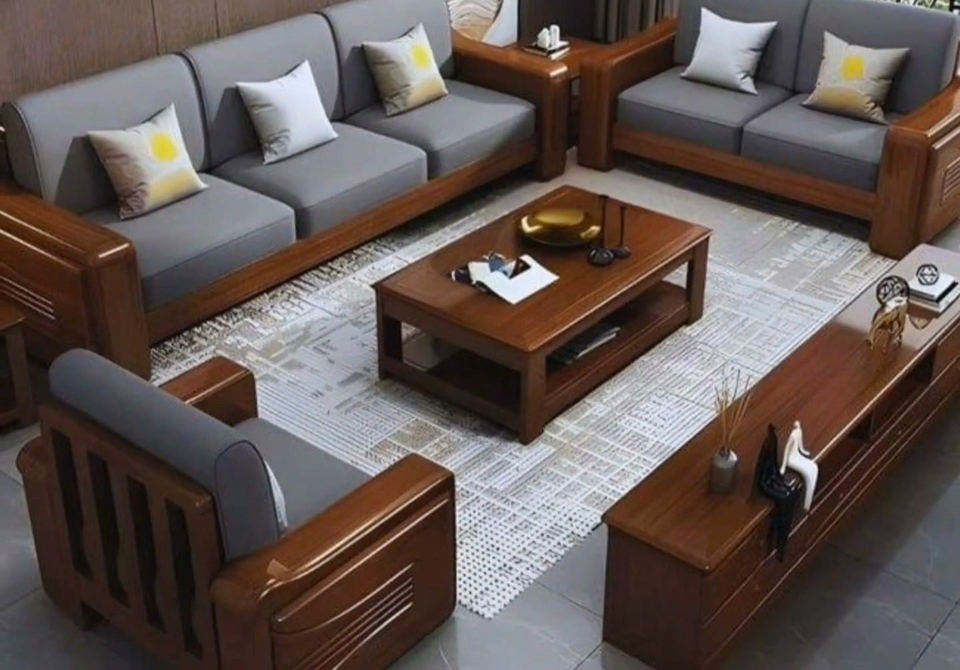}
{Entity 4}
\hfill
\frameitem{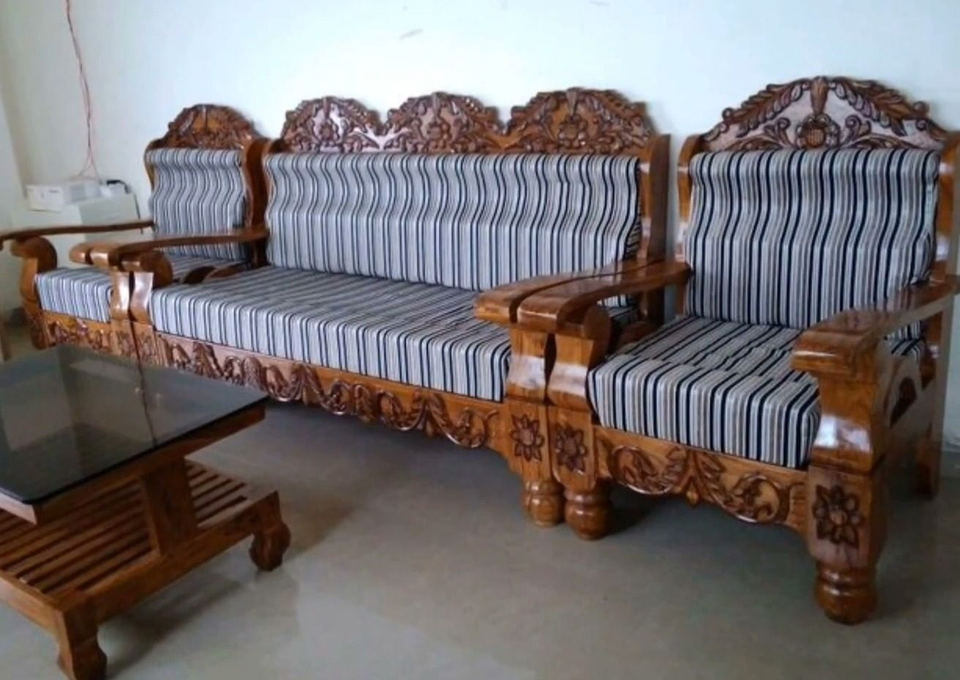}
{Entity 5}
\hfill
\frameitem{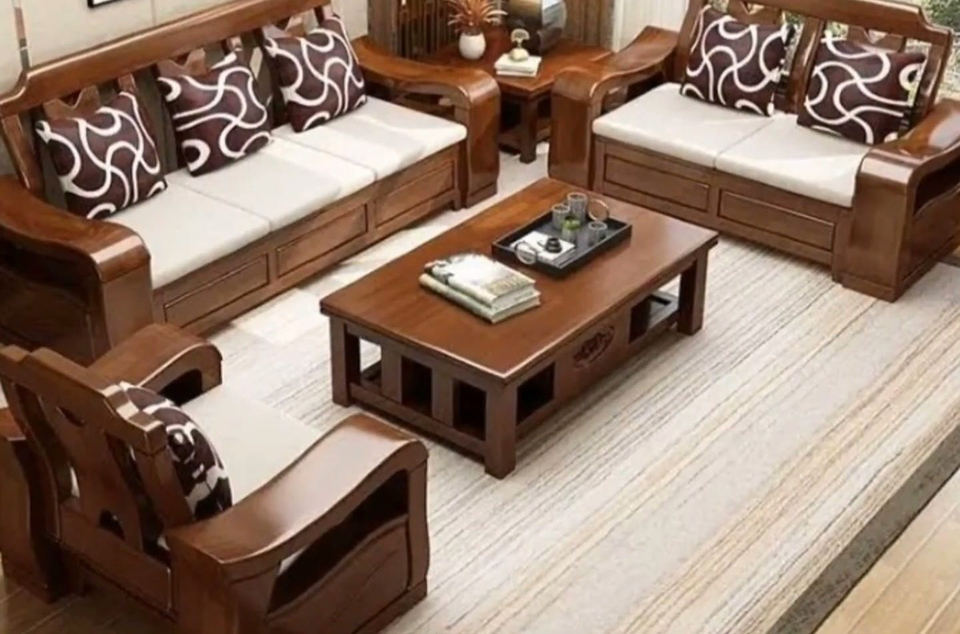}
{Entity 6}
\hfill
\frameitem{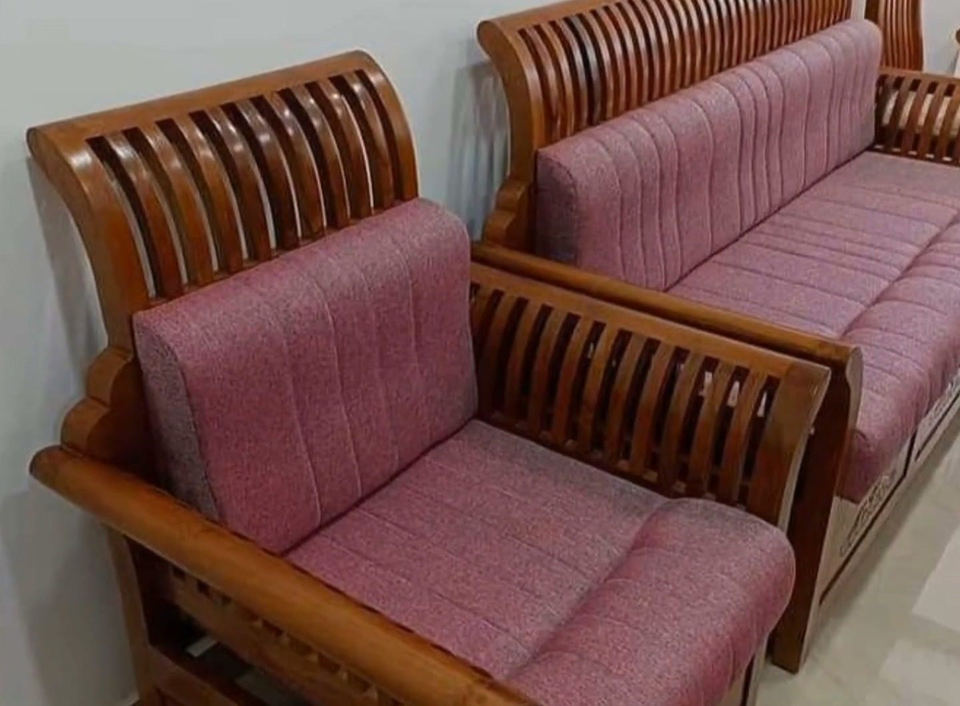}
{Entity 7}


\noindent
\textbf{Question:}
When the scene shows an orange wood sofa with grey cushions and white throw pillows placed on a dirt floor, and a rectangular wooden coffee table, and I have recently redecorated my home in a more neutral and minimalist style, please tell me which of the current sofas is the most suitable for me?


\noindent
\textbf{Action Options:}

\begin{tabularx}{\linewidth}{
    @{}
    >{\raggedright\arraybackslash}p{0.035\linewidth}
    >{\raggedright\arraybackslash}X
    @{}
}

A. &
How deep a tone do you now prefer on your sofa's main cushions?
\\[2pt]

\textbf{B.} &
\textbf{What visual treatment do you now prefer for your sofa's main cushioning?}
\\[2pt]

C. &
How visually uniform do you now prefer your sofa cushions to appear? ...
\\[2pt]

\end{tabularx}


\noindent
\textbf{Old User Preference:}
I formerly preferred black-and-white striped upholstery.

\noindent
\textbf{New User Preference:}
I now prefer solid charcoal-gray upholstery.


\noindent
\textbf{Answer Options:}

\begin{tabularx}{\linewidth}{
    @{}
    >{\raggedright\arraybackslash}p{0.035\linewidth}
    >{\raggedright\arraybackslash}X
    @{}
}

A. &
Reddish-brown wood sofa set with light-gray cushions and white pillows outdoors.
\\[2pt]

B. &
Slatted brown wood armchair and sofa with mauve channel-stitched cushions.
\\[2pt]

\textbf{C.} &
\textbf{Reddish-brown wood sofa set with solid charcoal-gray cushions and pale pillows.}...
\\[2pt]

\end{tabularx}

\end{videoexamplebox}

\end{document}